\documentclass{article}

\PassOptionsToPackage{numbers, compress}{natbib}

\usepackage[preprint]{neurips_2025}

\usepackage{wrapfig}
\usepackage[utf8]{inputenc} 
\usepackage[T1]{fontenc}    
\usepackage{amsmath}
\usepackage{amsfonts}
\usepackage{amsthm}
\usepackage{amssymb}

\usepackage{array}
\usepackage{multirow}
\usepackage{makecell}
\usepackage{tabularx}
\usepackage{booktabs}
\usepackage{graphicx}
\usepackage{subcaption}  

\usepackage{hyperref}       
\usepackage{url}            
\usepackage{booktabs}       
\usepackage{amsfonts}       
\usepackage{nicefrac}       
\usepackage{microtype}      
\usepackage{xcolor}         
\usepackage[normalem]{ulem}
\useunder{\uline}{\ul}{}

\usepackage{algorithmic}
\usepackage{algorithm}

\usepackage{graphicx}
\usepackage{booktabs}
\usepackage{multirow}
\usepackage{array}
\usepackage{xcolor}
\usepackage{colortbl}

\title{MolEditRL: Structure-Preserving Molecular Editing via Discrete Diffusion and Reinforcement Learning}

\author{
Yuanxin Zhuang$^{1}$,  Dazhong Shen$^{2}$, Ying Sun$^{1}$\\
$^{1}$Artificial Intelligence Thrust, Hong Kong University of Science and Technology (Guangzhou) \\
$^{2}$College of Computer Science and Technology, Nanjing University of Aeronautics and Astronautics \\
\texttt{yzhuang436@connect.hkust-gz.edu.cn, shendazhong@nuaa.edu.cn, yings@hkust-gz.edu.cn}
}

\begin{document}

\maketitle

\begin{abstract}
    Molecular editing aims to modify a given molecule to optimize desired chemical properties while preserving structural similarity. However, current approaches typically rely on string-based or continuous representations, which fail to adequately capture the discrete, graph-structured nature of molecules, resulting in limited structural fidelity and poor controllability.
    In this paper, we propose MolEditRL, a molecular editing framework that explicitly integrates structural constraints with precise property optimization. Specifically, MolEditRL consists of two stages: (1) a discrete graph diffusion model pretrained to reconstruct target molecules conditioned on source structures and natural language instructions; (2) an editing-aware reinforcement learning fine-tuning stage that further enhances property alignment and structural preservation by explicitly optimizing editing decisions under graph constraints.
    For comprehensive evaluation, we construct MolEdit-Instruct, the largest and most property-rich molecular editing dataset, comprising 3 million diverse examples spanning single- and multi-property tasks across 10 chemical attributes. Experimental results demonstrate that MolEditRL significantly outperforms state-of-the-art methods in both property optimization accuracy and structural fidelity, achieving a 74\% improvement in editing success rate while using 98\% fewer parameters.

  \end{abstract}

\section{Introduction}~\label{Introduction}

\begin{wrapfigure}{r}{0.5\textwidth}
  \vspace{-5pt}  
  \centering
  \includegraphics[width=0.45\textwidth]{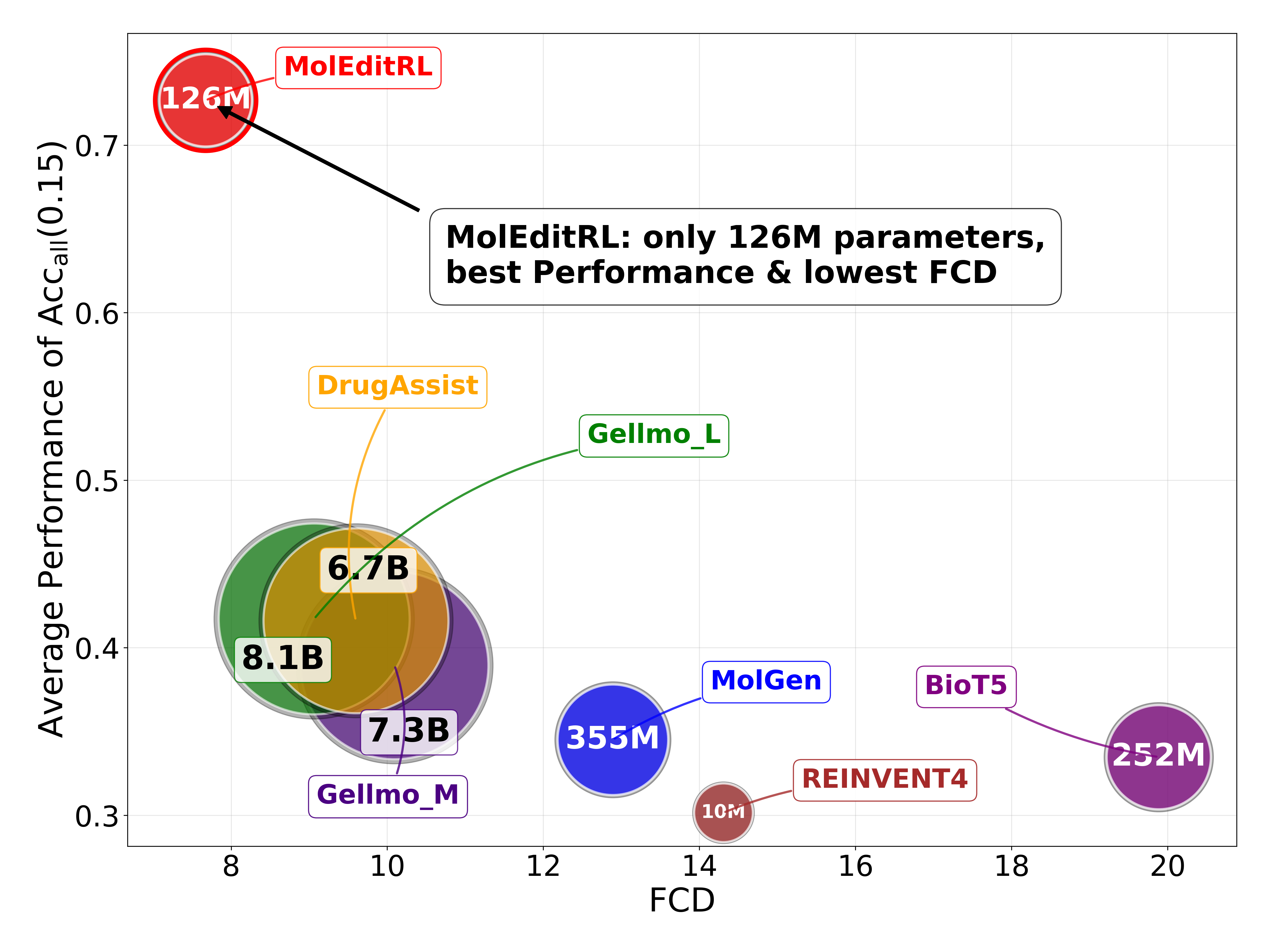}
  \vspace{-5pt}
  \caption{Performance, FCD, and parameter size comparison.}
  \label{fig:parameter}
  \vspace{-5pt}  
\end{wrapfigure}

Designing molecules with tailored properties is essential for drug discovery~\cite{ma2024rational}. Unlike \textit{de novo} molecular generation that creates molecules from scratch~\cite{wang2022deep}, molecular editing~\cite{hui2022molecular} focuses on precisely modifying existing molecules to optimize targeted properties while preserving known structure-activity relationships~\cite{hansch1969quantitative}.

Traditional molecular editing approaches fall into three main paradigms:
(1) Rule-based methods apply predefined fragment transformations~\cite{chen2021deep, fu2021mimosa}, but their generalization is limited by manually designed rules.
(2) Latent generative models\cite{jin2018junction, shi2020graphaf} optimize molecular properties in continuous spaces, yet often struggle with fine-grained control due to latent compression.
(3) Sequence-to-sequence methods\cite{he2021molecular, loeffler2024reinvent, wu2024leveraging} frame editing as SMILES translation, enabling scalable learning but lacking structural precision, as small token changes can produce unpredictable or invalid edits~\cite{kusner2017grammar, krenn2020self}.

Recently, language models have expanded the landscape of molecular editing by integrating natural language understanding with chemical representations, enabling models to leverage semantic instructions for molecular modifications:
(1) Graph-embedding approach~\cite{liu2023multi} encodes both molecules and textual instructions into a shared latent space via contrastive learning.
(2) SMILES-based models~\cite{liu2024conversational, ye2025drugassist, le2024utilizing, dey2025mathtt} use retrieval-augmented generation or adopt instruction tuning to enhance editing relevance.
(3) SELFIES-based method~\cite{fang2024domain} incorporates chemical feedback to reduce syntax errors and improve property alignment during generation.
Despite their promise, these language-based methods often struggle to preserve scaffolds or perform precise, localized modifications, due to the non-uniqueness of textual molecular representations. Structurally similar molecules may appear textually distant, leading to inconsistent and unreliable edits~\cite{noutahi2024gotta, arus2019randomized}.

There are several critical challenges in molecular editing for ensuring structural preservation and editing precision:
First, molecular editing must explicitly align with the discrete, graph-structured nature of molecules. String-based representations fail to explicitly encode topological constraints, often limiting the model’s ability to preserve scaffolds and perform localized modifications.
Second, existing methods trained solely on fixed datasets lack mechanisms for actively exploring novel editing strategies, restricting generalization and adaptability to complex or underexplored regions of chemical space.
Third, performing discrete edits directly on molecular graphs while preserving structural fidelity and aligning with natural language instructions is technically challenging due to the non-differentiable, high-dimensional nature of graph representations.

In this paper, we propose MolEditRL, a structure-aware molecular editing framework that combines discrete graph diffusion with reinforcement learning. MolEditRL first employs discrete diffusion to reconstruct target molecules conditioned simultaneously on source molecular structures and natural language instructions, effectively capturing both structural and semantic relationships. To further enhance the precision of property optimization and alignment with instructions, we introduce editing-aware reinforcement learning guided by explicit property rewards, while incorporating constraints to preserve structural integrity.
To enable comprehensive evaluation, we introduce MolEdit-Instruct, a large-scale molecular editing dataset containing 3 million editing examples spanning 10 diverse chemical properties, including biological activities, physicochemical attributes, and synthetic accessibility. Compared to existing datasets~\cite{ye2025drugassist, dey2025mathtt}, MolEdit-Instruct provides broader property coverage and more realistic single- and multi-property editing scenarios. We release MolEdit-Instruct publicly on Hugging Face to facilitate future research.

Experimental results demonstrate that MolEditRL significantly outperforms state-of-the-art methods in both editing accuracy and distributional fidelity (measured by Fréchet ChemNet Distance, FCD). Remarkably, MolEditRL achieves a 74\% improvement in editing success rate over leading baselines while requiring 98\% fewer parameters (Figure~\ref{fig:parameter}).
Our contributions are summarized as follows:
(1) We propose MolEditRL, a molecular editing framework explicitly designed to maintain structural integrity during editing.
(2) We introduce a two-stage training strategy that combines discrete diffusion pretraining with reinforcement learning fine-tuning, achieving precise property optimization with structural constraints.
(3) MolEditRL achieves SOTA editing performance with substantially fewer parameters and the lowest distributional distance (FCD) compared to existing methods.

\section{Related Works}

\textbf{Molecular Editing. }
Molecular editing aims to modify a given molecule to enhance specific chemical properties while preserving its structural similarity—a critical task in drug discovery. Formally, given a source molecule $G_{\text{src}}$ and a textual instruction $S$ describing desired modifications, the goal is to generate a target molecule $G_{\text{tgt}}$ that satisfies the instructed property changes while maintaining structural fidelity.
Existing molecular editing approaches typically fall into three main paradigms:
(1) Rule-based Graph Editing.
These methods directly manipulate molecular graphs using predefined or data-driven transformation rules, such as fragment replacements or bond editing templates, inspired by Matched Molecular Pairs (MMP)~\cite{dalke2018mmpdb, chen2021deep, fu2021mimosa}. While offering high chemical interpretability and precise local modifications, their generalizability is limited by the coverage and flexibility of manually or heuristically derived rules.
(2) Latent Generative Graph Editing.
Approaches such as JT-VAE~\cite{jin2018junction} and GraphAF~\cite{shi2020graphaf} encode molecules into continuous latent spaces and decode edited structures by sampling. Hierarchical decoding techniques like HierG2G~\cite{jin2020hierarchical} enhance structural preservation by generating molecules in a coarse-to-fine manner. However, these methods frequently face issues such as information loss due to latent compression, resulting in limited accuracy and insufficient control over edits.
(3) Sequence-based Generation.
These approaches treat molecular editing as a sequence translation task, converting source SMILES strings into target SMILES using Transformer-based architectures~\cite{he2021molecular, loeffler2024reinvent, wu2024leveraging}. Despite their scalability and ease of implementation, these models suffer from syntactic instability and representation ambiguity: structurally similar molecules can have significantly different SMILES representations, and small token-level edits may lead to unpredictable or chemically invalid outputs, limiting their precision and structural controllability.
(4) Language-based models.
MOLGEN~\cite{fang2024domain} addresses SMILES fragility by adopting the SELFIES representation. Methods such as ChatDrug~\cite{liu2024conversational}, DrugAssist~\cite{ye2025drugassist}, and Re2DF~\cite{le2024utilizing} utilize retrieval-augmented generation or instruction tuning to enhance editing relevance. Additionally, embedding-based methods leveraging diffusion~\cite{xiong2024text} or contrastive learning~\cite{liu2023multi} have been proposed. Nonetheless, these approaches continue to rely on textual or continuous representations that lack explicit alignment with discrete molecular graph structures, compromising structural fidelity and editing accuracy.

\textbf{Reinforcement Learning in Molecular Generation. }
Reinforcement learning (RL) provides a flexible framework for molecular optimization by formulating molecule generation as a Markov Decision Process (MDP), in which agents sequentially modify molecular structures to maximize rewards associated with desired chemical properties~\cite{sridharan2024deep}. SMILES-based RL methods such as ReLeaSE~\cite{popova2018deep} and REINVENT~\cite{olivecrona2017molecular} guide generative models using property predictors and prior policies. Graph-based RL methods, including MolGAN~\cite{de2018molgan}, GCPN~\cite{you2018graph}, and MolDQN~\cite{zhou2019optimization}, facilitate goal-directed graph construction through adversarial training, policy gradients, or Q-learning. Although effective in exploring chemical space, these RL-based frameworks typically focus on \textit{de novo} molecule generation and lack explicit mechanisms to enforce structural constraints derived from source molecules, limiting their applicability to structurally constrained molecular editing tasks.

\section{Method}

We present MolEditRL, a structure-preserving molecular editing framework trained in two stages. First, molecules and instructions are encoded into unified graph-text representations. Then, a structure-aware editing network is trained via (1) discrete diffusion pretraining to reconstruct target molecules from noisy graphs and instructions, and (2) reinforcement learning fine-tuning to optimize property alignment while preserving structural fidelity.

\subsection{Molecular Tokenizing}
\begin{figure}[htbp]
    \centering
    \includegraphics[width=1\textwidth]{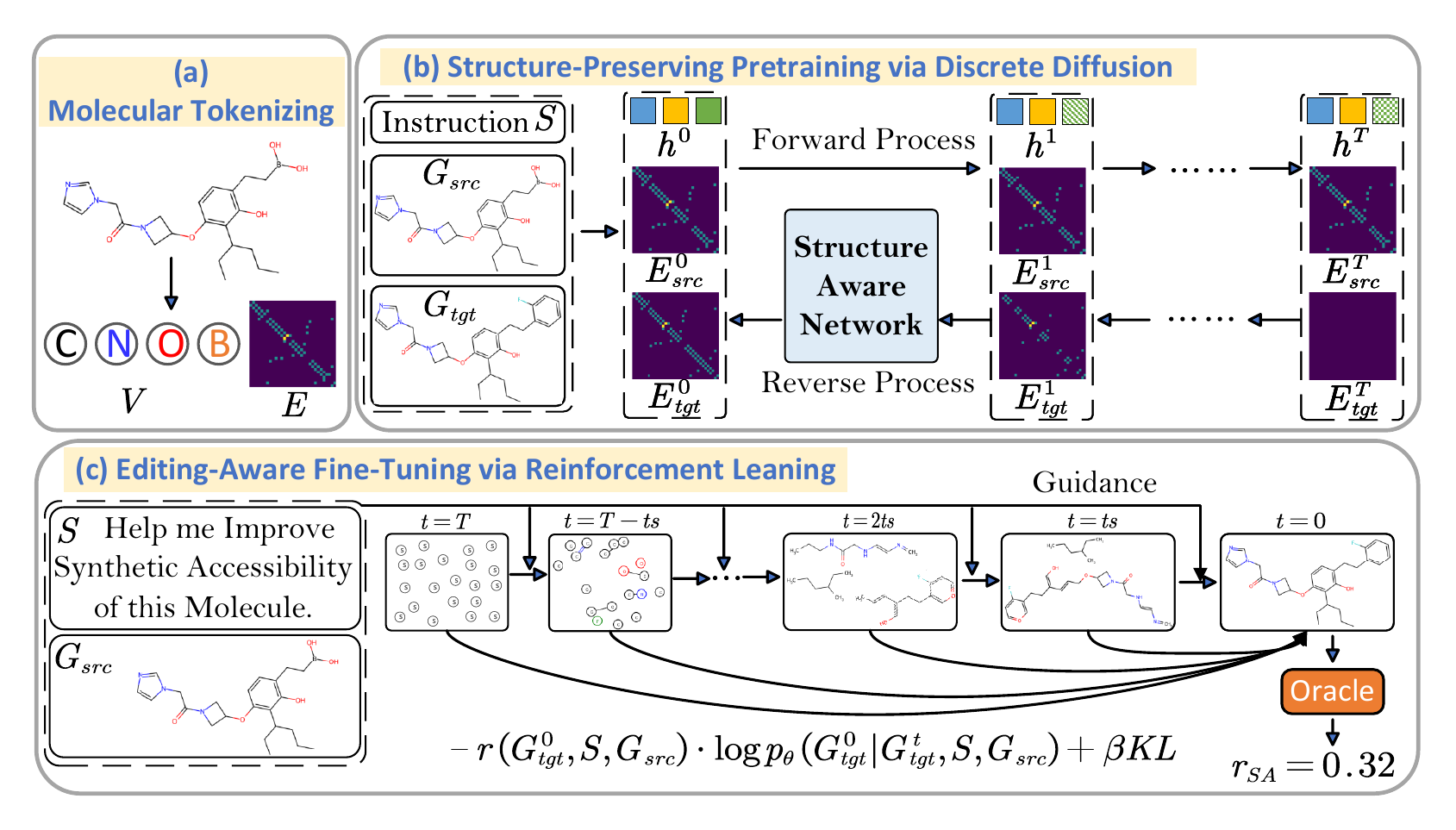}
    \vspace{-15pt}
    \caption{Overview of MolEditRL.}
    \label{fig:model}
    \vspace{-10pt}
\end{figure}

As illustrated in Figure~\ref{fig:model} (a), a molecule is represented as an attributed graph $G = (V, E)$, where $V$ denotes atom nodes with associated features, and $E$ denotes bond edges with bond-type attributes. The editing instruction is a sequence of tokens $S = [s_1,\ldots,s_n]$. Given a source molecule graph $G_{\text{src}}=(V_{src}, E_{src})$, the model aims to predict the target graph $G_{\text{tgt}}=(V_{tgt}, E_{tgt})$ that reflects the required edits. These components are embedded and concatenated into a unified input sequence:
\begin{equation}
h^0 = [\,h_1,\dots,h_n,\;h^{src}_{n+1},\dots,h^{src}_{n+k},\;h^{tgt}_{n+k+1},\dots,h^{tgt}_{n+k+m}\,] \,\in \mathbb{R}^{(n+k+m)\times d_h},
\end{equation}
where $h_1,\dots,h_n \in \mathbb{R}^{d_h}$ are embeddings for the instruction tokens, $h^{src}_{n+1},\dots,h^{src}_{n+k} \in \mathbb{R}^{d_h}$ encode source atoms, and $h^{tgt}_{n+k+1},\dots,h^{tgt}_{n+k+m} \in \mathbb{R}^{d_h}$ encode target atoms. The dimension $d_h$ is the model’s hidden size.

\subsection{Structure-Preserving Editing Network}

Recent work~\cite{xiang2024instruction} has explored aligning molecular graph structures with language semantics for text-conditioned generation. To enable precise and structure-aware molecular editing, we propose the Structure-Preserving Editing Network, which jointly encodes the semantic intent of natural language instructions and the topological features of source and target graphs. We initialize our transformer encoder with a pretrained RoBERTa model~\cite{liu2019roberta}, but enhance it with a structure-aware attention mechanism. This mechanism injects bond-level connectivity priors into attention scores via learnable bias terms that encode graph connectivity, guiding attention flows to preserve structural integrity while selectively updating target representations.
For tokens $i$ and $j$ at layer $l$, the raw attention score $\hat{A}^l_{i,j}$ is computed as:
\begin{equation}
\hat{A}^l_{i,j} \;=\; \frac{1}{\sqrt{d_k}}\,(h^l_i W_Q)\,\bigl(h^l_j W_K\bigr)^\top \;+\; b^l_{i,j},
\end{equation}
where $W_Q, W_K \in \mathbb{R}^{d_h \times d_k}$ are learnable projection matrices. To incorporate graph-level dependencies and better preserve structure during editing, we introduce $b^l_{i,j}$ to encode structural priors:
\begin{equation}
b^l_{i,j} \;=\;
\begin{cases}
E_{src}[i-n,j-n,:], & \text{if } n < i,j \le n+k, \\
E_{tgt}[i-(n+k),\,j-(n+k),:], & \text{if } l=0 \text{ and } n+k < i,j, \\
A^{l-1}_{i,j}, & \text{if } l>0 \text{ and } n+k < i,j, \\
0, & \text{otherwise}.
\end{cases}
\end{equation}
The bias term $b^l_{i,j}$ preserves structure by injecting source adjacency at all layers, using target adjacency at the first layer, and propagating attention from previous layers to maintain topology-aware attention flow. The attention weights are then normalized, and the updated hidden states are calculated as:
\begin{equation}
A^l_{i,j} = \frac{\exp\bigl(\hat{A}^l_{i,j}\bigr)}{\sum_{k}\exp\bigl(\hat{A}^l_{i,k}\bigr)}, 
\quad 
h^{l+1}_i = \sum_{j} A^l_{i,j}\,\bigl(h^l_j W_V\bigr) W_O,
\end{equation}
where $W_V \in \mathbb{R}^{d_h \times d_k}$ and $W_O \in \mathbb{R}^{d_k \times d_h}$ are learnable parameters.
After $L$ transformer layers, the final hidden states produce predictions for both node and edge attributes. For each node $v_i$, a masked language modeling (MLM) head $\text{MLM}_V$ outputs $\hat{p}(v_i)$. For edges, an MLM head $\text{MLM}_E$ operates on the learned bias or other representations, denoted $\hat{p}(e^t_{i,j})$. Edge predictions are symmetrized $\bigl(e_{i,j}+e_{j,i}\bigr)/2$ to respect bond-direction constraints.

\subsection{Structure-Preserving Pretraining via Discrete Diffusion}

As Figure~\ref{fig:model} (b), we pretrain the editing network via a discrete denoising diffusion process conditioned on the source molecule and instruction, enabling topology-aware generation.

\textbf{1. Forward Process.}
We define a discrete forward process that gradually corrupts the target molecular graph $G^0_{\text{tgt}}$ over $T$ timesteps. At each step $t$, atom and bond features are independently masked with probability $\beta(t) = (T - t + 1)^{-1}$:
\begin{equation}
q(G^{1:T}_{\text{tgt}}\!\mid G^{0}_{\text{tgt}}) \;=\; \prod\nolimits_{\substack{t=1}}^T q(G^{t}_{\text{tgt}}\!\mid G^{t-1}_{\text{tgt}}),
\quad
q(G^{t}_{\text{tgt}}\!\mid G^{t-1}_{\text{tgt}}) = \bigl(V^{t-1}_{\text{tgt}}\,{Q}^V_t,\; E^{t-1}_{\text{tgt}}\,{Q}^E_t\bigr),
\end{equation}
where $Q_t$ is a transition matrix that gradually increases the masking rate. At each step, each element remains the same with probability $1-\beta(t)$ or transitions to [MASK] with probability $\beta(t)$. This process gradually converts $G^0_{\text{tgt}}$ into a fully masked graph $G^{T}_{\text{tgt}}$.


\textbf{2. Reverse Process.}
To recover $G^{0}_{\text{tgt}}$ from the fully corrupted graph $G^{T}_{\text{tgt}}$, we train a denoising model $\phi_\theta$ to iteratively refine $G^{t}_{\text{tgt}}$ conditioned on the source molecule $G_{\text{src}}$ and instruction $S$:
\begin{equation}
p_\theta\bigl(G_{0:T-1} \mid G^{T}_{\text{tgt}},\, G_{\text{src}},\, S\bigr)
\;=\;
\prod\nolimits_{\substack{t=1}}^T p_\theta\bigl(G^{t-1}_{\text{tgt}}\mid G^{t}_{\text{tgt}},\;G_{\text{src}},\;S\bigr).
\end{equation}
At each timestep, $\phi_\theta$ predicts the denoised graph $G^{t-1}_{\text{tgt}}$ using the editing network described earlier.

\textbf{3. Training Objective.}
Although instruction $S$ is not corrupted during diffusion, we include a cross-entropy loss on instruction tokens to enforce semantic alignment with the predicted molecule.
\begin{equation}
\mathcal{L}_{pre}
=
\underbrace{
\sum_{i=n+1}^{n+k+m} \text{CE}\bigl(v_i,\;\hat{p}_i(v_i)\bigr)
}_{\text{Node prediction}}
\;+\;
\underbrace{
\sum_{i,j=n+k+1}^{n+k+m} \text{CE}\bigl(e_{i,j},\;\hat{p}_{i,j}(e^t_{i,j})\bigr)
}_{\text{Edge prediction}}
\;+\;
\underbrace{
\sum_{i=1}^{n} \text{CE}\bigl(s_i,\;\hat{p}_i(s_i)\bigr)
}_{\text{Text preservation}}.
\end{equation}

\textbf{4. Sampling.}
At inference time, we initialize a fully masked graph $G^{T}_{\text{tgt}}$ and iteratively apply the reverse process to reconstruct $G^{0}_{\text{tgt}}$. Each step consists of:
(1) \textit{Graph-Text Encoding}: The transformer $\phi_\theta$ encodes the current noisy graph $G^{t}_{\text{tgt}}$, source molecule $G_{\text{src}}$, and instruction $S$, producing logits $\hat{p}(V_{\text{tgt}}) \in \mathbb{R}^{m\times a}$ and $\hat{p}(E_{\text{tgt}}) \in \mathbb{R}^{m\times m\times b}$ over atom and bond types.
(2) \textit{Prediction}: Following $x_0$-parameterization~\cite{austin2021structured}, the model predicts the denoised graph as $\hat{G}^{0}_{\text{tgt}} = \arg\max \hat{p}(V_{\text{tgt}}, E_{\text{tgt}})$.
(3) \textit{Sampling}: The next graph $G^{t-1}_{\text{tgt}}$ is sampled from the posterior $q(G^{t-1}_{\text{tgt}} \mid G^{t}_{\text{tgt}}, \hat{G}^{0}_{\text{tgt}})$ by independently sampling atoms and bonds: $G^{t-1}_{\text{tgt}} \sim \prod_i p_\theta(v^{t-1}_i) \prod_{i,j} p_\theta(e^{t-1}_{i,j})$.



\subsection{Editing-Aware Fine-Tuning via Reinforcement Learning}

While the pretrained diffusion model captures molecular structure and ensures validity, it lacks explicit optimization for property-specific editing. We address this by introducing Editing-Aware Reinforcement Learning, which fine-tunes the model using rewards computed from well-established chemical toolkits (RDKit~\cite{bento2020open} and TDC~\cite{huang2021therapeutics}). A KL-regularized objective guides optimization toward desired properties while preserving structural consistency.

\textbf{1. MDP Formulation for Molecular Editing.}
We recast discrete graph denoising as a Markov Decision Process (MDP) tailored specifically for molecular editing:
(1) \textit{State:} $s_t = \bigl(S,\,G_{\text{src}},\,G^{T-t}_{\text{tgt}}\bigr)$ includes the instruction, source molecule, and current noisy target graph.  
(2) \textit{Action:} $a_t = G^{T-t-1}_{\text{tgt}}$ is the denoised graph at the next step.  
(3) \textit{Initial State:} Sampled as $P_0(s_0) = p(S)\,p(G_{\text{src}})\,q(G^{T}_{\text{tgt}})$, comprising an instruction, a source molecule, and a fully masked target graph.
(4) \textit{Transition:} $P(s_{t+1}\mid s_t, a_t)$ is deterministic given $a_t$.  
(5) \textit{Policy:} Defined by the denoising model $\pi_\theta(a_t \mid s_t)$.  
(6) \textit{Reward:} A scalar reward is assigned only at the final step ($t = T{-}1$) to evaluate editing success:
\begin{equation}
R(s_t, a_t) = 
\begin{cases}
r(G^0_{\text{tgt}}, S, G_{\text{src}}), & \text{if } t = T - 1, \\
0, & \text{otherwise},
\end{cases}
\end{equation}
where $r(\cdot)$ precisely measures property improvements through chemical-rule-based oracles provided by RDKit and TDC, guaranteeing reward accuracy and reliability.

\textbf{2. KL-Regularized RL Objective.}
To optimize molecular editing while preserving structural fidelity, we adopt a KL-regularized reinforcement learning objective. Formally, the objective is:
\begin{align}~\label{eq:objective}
    \mathcal{L}(\theta) &= - \mathbb{E}_{p(S)\,p(G_{\text{src}})} \, \mathbb{E}_{p_\theta(G^0_{\text{tgt}})} \bigl[ r(G^0_{\text{tgt}}, S, G_{\text{src}}) \bigr] \notag \\
    &\quad + \beta \sum\nolimits_{\substack{t=1}}^T \mathbb{E}_{p_\theta(G^t_{\text{tgt}})} \Bigl[ D_{\mathrm{KL}}\bigl( p_\theta(G^0_{\text{tgt}} \mid G^t_{\text{tgt}}, S, G_{\text{src}}) \,\|\, p_{\text{pre}}(G^0_{\text{tgt}} \mid G^t_{\text{tgt}}, S, G_{\text{src}}) \bigr) \Bigr],
\end{align}
where $p_\theta$ is the current policy’s distribution over denoised target molecules at timestep $t$, and $p_{\text{pre}}$ is the pretrained diffusion model, acting as a structure-aware prior. The coefficient $\beta$ balances reward maximization and structural consistency. To stabilize training, we normalize rewards within each batch: $\hat{A}_i = \frac{r_i - \text{mean}(r)}{\text{std}(r) + 10^{-6}}$. To reduce computation and improve efficiency, we apply policy updates at a fixed stride $t_s$, rather than every timestep. Specifically, we define the update set as: $\mathcal{T} = \{ t \in [1, T] \mid t \bmod t_s = 0 \}$, and compute gradients only at $t \in \mathcal{T}$. 
The resulting policy gradient becomes:
\begin{equation}\label{eq:policy_gradient}
    \resizebox{0.92\textwidth}{!}{$
\nabla_\theta J(\theta) = \mathbb{E}_{G_{\text{tgt}}^{0:T} \sim p_\theta} \left[
\hat{A} \cdot \sum\nolimits_{\substack{t \in \mathcal{T}}} \nabla_\theta \log p_\theta(\hat{G}^0_{\text{tgt}} \mid G^t_{\text{tgt}}, S, G_{\text{src}})
- \beta \sum\nolimits_{\substack{t \in \mathcal{T}}} \nabla_\theta D_{\mathrm{KL}}\left( p_\theta \,\|\, p_{\text{pre}} \right)
\right].
$}
\end{equation}

\textbf{3. Gradient Estimation via $x_0$-Parameterization.}
We estimate the gradient of the reward term (i.e., the first term of Eq.~\ref{eq:objective}). By the policy gradient theorem, the gradient is:
\begin{equation}\label{eq:DDPO}
\nabla_\theta \mathbb{E}_{G^{0:T}_{\text{tgt}}}[r] = \mathbb{E}_{G^{0:T}_{\text{tgt}}} \left[
r(G^0_{\text{tgt}}, S, G_{\text{src}}) \cdot 
\sum\nolimits_{\substack{t \in \mathcal{T}}} \nabla_\theta \log p_\theta(G^{t-1}_{\text{tgt}} \mid G^t_{\text{tgt}}, S, G_{\text{src}})
\right].
\end{equation}
Since rewards are only available at t{=}0, directly estimating the gradient suffers from high variance~\cite{liu2024graph}. To reduce this, we adopt $x_0$-parameterization~\cite{austin2021structured}, rewriting the reverse transition as:
\begin{equation}
    p_\theta(G^{t-1}_{\text{tgt}} \mid G^t_{\text{tgt}}, S, G_{\text{src}}) \approx 
    \sum\nolimits_{\substack{G^0_{\text{tgt}}}} 
    q(G^{t-1}_{\text{tgt}} \mid G^t_{\text{tgt}}, \hat{G}^0_{\text{tgt}})\, 
    p_\theta(\hat{G}^0_{\text{tgt}} \mid G^t_{\text{tgt}}, S, G_{\text{src}}),
    \end{equation}
where $q(\cdot)$ is a known corruption distribution in the forward process. This approximation yields:
\begin{equation}\label{eq:GDPO}
\nabla_\theta \log p_\theta(G^{t-1}_{\text{tgt}} \mid G^t_{\text{tgt}}, S, G_{\text{src}}) \approx \nabla_\theta \log p_\theta(\hat{G}^0_{\text{tgt}} \mid G^t_{\text{tgt}}, S, G_{\text{src}}).
\end{equation}
Under this formulation, the gradient of the reward term can be approximated via a reward-weighted cross-entropy loss over atoms and bonds:
\begin{equation}
    \resizebox{0.92\textwidth}{!}{$
 \sum_{t \in \mathcal{T}} r(G^0_{\text{tgt}}, S, G_{\text{src}}) \cdot \left(
    \sum_{i} \text{CE}\big(v^0_i,\; p_\theta(\cdot \mid G^t_{\text{tgt}}, S, G_{\text{src}})\big) + \sum_{i,j} \text{CE}\big(e^0_{i,j},\; p_\theta(\cdot \mid G^t_{\text{tgt}}, S, G_{\text{src}})\big)
    \right),
    $}
\end{equation}
where $v^0_i$ and $e^0_{i,j}$ are atoms and bonds in the final predicted molecule $G^0_{\text{tgt}}$, reused as supervision targets at each selected step $t \in \mathcal{T}$.


\section{Experiments}

\subsection{Data Construction}

We construct MolEdit, a large-scale and property-rich dataset specifically tailored for molecular editing with natural language instructions. Existing datasets, such as MolOpt-Instructions~\cite{ye2025drugassist} and MuMOInstruct~\cite{dey2025mathtt}, are limited in either property coverage, task diversity, or data scale. MolEdit addresses these gaps by extending the property set to 10 diverse chemical attributes—spanning biological activity, physicochemical characteristics, and synthetic accessibility—and defining 20 representative editing tasks (10 increases and 10 decreases). It contains 3 million high-quality molecular pairs (967K unique), each exhibiting substantial property shifts while maintaining high structural similarity (Tanimoto scores from 0.650 to 0.982). This provides a more realistic and comprehensive testbed for training and evaluating editing models. 
Further dataset construction details are provided in Appendix~\ref{sec:dataset_stats}. The model architecture is described in Appendix~\ref{sec:architecture}, and the training setup is detailed in Appendix~\ref{sec:training}.


\subsection{Baselines} 

We compare MolEditRL with five models in molecular editing:
(1) BioT5\cite{pei2023biot5} leverages SELFIES and a T5-style architecture for cross-modal learning between molecules and text.
(2) DrugAssist\cite{ye2025drugassist} is a Llama2-7B-based dialogue model for interactive molecule optimization.
(3) GeLLM$^3$O\cite{dey2025mathtt} uses instruction tuning on Mistral and Llama3 models for multi-property optimization; we evaluate both GeLLM$^3$O\_M and GeLLM$^3$O\_L, using the GeLLM$^3$O-P(6) variant trained on six properties.
(4) MolGen\cite{fang2024domain} is a domain-agnostic language model trained with chemical feedback to reduce invalid generations.
(5) REINVENT 4~\cite{loeffler2024reinvent} integrates reinforcement learning, transfer learning, and curriculum learning for molecular design using RNN and Transformer backbones.
\vspace{-10pt}

\subsection{Evaluation Metrics} 

To comprehensively assess molecular editing performance, we use the following metrics to evaluate chemical validity, editing accuracy under structural constraints, and overall molecular quality. Chemical validity and property values are computed using RDKit~\cite{bento2020open} and Therapeutics Data Commons (TDC)~\cite{huang2021therapeutics}, two widely used and trusted toolkits for molecular analysis:
(1) \textbf{Validity} is the proportion of generated molecules that are chemically valid, reflecting the model’s ability to produce syntactically correct molecular structures.
(2) \textbf{Overall Accuracy (\(\text{Acc}_{\text{all}}\)($\tau$)) and Valid Accuracy (\(\text{Acc}_{\text{valid}}\)($\tau$))} jointly measure editing success under structural similarity constraints. For a given Tanimoto threshold $\tau$, \(\text{Acc}_{\text{all}}\)($\tau$) is the percentage of all outputs that satisfy both the desired property changes and structural similarity $(\geq \tau)$; \(\text{Acc}_{\text{valid}}\)($\tau$) restricts this to valid molecules only. This distinction allows us to separate editing performance from chemical validity.
(3) \textbf{Fr{\'e}chet ChemNet Distance (FCD)}~\cite{preuer2018frechet} quantifies the distributional distance between generated and reference molecules. Lower FCD values indicate better alignment in chemical space, capturing both diversity and realism.
\vspace{-12pt}

\subsection{Experimental Results}

Table~\ref{maintable} compares molecular editing models on single-property and multi-property tasks.
MolEditRL consistently achieves the highest editing accuracy across all tasks and similarity thresholds ($\tau = 0.65$ and $0.15$). While SELFIES-based models (BioT5, MolGen) guarantee perfect chemical validity, they fail to preserve structural similarity, often achieving zero accuracy at $\tau = 0.65$, which reflects a lack of structural alignment.
DrugAssist, based on SMILES and LLM fine-tuning, maintains high validity but performs significantly worse than MolEditRL on both Acc$\text{all}$ and Acc$\text{valid}$. This indicates that chemical correctness alone is insufficient for precise, property-aligned editing. Although DrugAssist generates valid molecules, it struggles to retain scaffold similarity while optimizing properties.
All baseline models yield substantially higher FCD scores than MolEditRL, suggesting greater divergence from real molecule distributions. In contrast, MolEditRL generates molecules that are both valid and distributionally faithful, benefiting from structure-aware graph editing.
For multi-property tasks, we evaluate six scenarios aligned with real-world drug discovery objectives, such as improving stability and synthesis (Haccept$\downarrow$, SA$\downarrow$), balancing drug-likeness and accessibility (QED$\uparrow$, SA$\downarrow$), and resolving conflicting goals (Haccept$\uparrow$, MW$\uparrow$, QED$\downarrow$). MolEditRL consistently outperforms all baselines in these tasks, demonstrating strong generalization and effective handling of complex, multi-objective constraints. 
Generalization to unseen properties is discussed in Appendix~\ref{sec:unseen_properties}, and extended results on single-property and multi-property tasks are available in Appendix~\ref{sec:extended_singleprop} and Appendix~\ref{sec:extended_multitask}, respectively.
\vspace{-10pt}

\begin{table}[t]
\centering
\caption{Comparison of molecular editing models across tasks. Bold indicates best performance. Arrows ($\uparrow$, $\downarrow$) denote desired property increase or decrease.}
\resizebox{\textwidth}{!}{%
\begin{tabular}{ccccccccccccccc}
\hline
Model             & Task                                                                                                              & Validity & \begin{tabular}[c]{@{}c@{}}\(\text{Acc}_{\text{all}}\)\\ (0.65)\end{tabular} & \begin{tabular}[c]{@{}c@{}}\(\text{Acc}_{\text{valid}}\)\\ (0.65)\end{tabular} & \begin{tabular}[c]{@{}c@{}}\(\text{Acc}_{\text{all}}\)\\ (0.15)\end{tabular} & \begin{tabular}[c]{@{}c@{}}\(\text{Acc}_{\text{valid}}\)\\ (0.15)\end{tabular} & FCD              & Task                                                                                                              & Validity & \begin{tabular}[c]{@{}c@{}}\(\text{Acc}_{\text{all}}\)\\ (0.65)\end{tabular} & \begin{tabular}[c]{@{}c@{}}\(\text{Acc}_{\text{valid}}\)\\ (0.65)\end{tabular} & \begin{tabular}[c]{@{}c@{}}\(\text{Acc}_{\text{all}}\)\\ (0.15)\end{tabular} & \begin{tabular}[c]{@{}c@{}}\(\text{Acc}_{\text{valid}}\)\\ (0.15)\end{tabular} & FCD              \\ \hline
BioT5       & \multirow{7}{*}{GSK3$\beta{\uparrow}$}                                                                            & 1        & 0                                                                            & 0                                                                              & 0.328                                                                        & 0.328                                                                          & 15.004          & \multirow{7}{*}{Rotbonds${\downarrow}$}                                                                           & 1        & 0                                                                            & 0                                                                              & 0.304                                                                        & 0.304                                                                          & 17.861          \\
MolGen      &                                                                                                                   & 1        & 0.024                                                                        & 0.024                                                                          & 0.315                                                                        & 0.315                                                                          & 11.613          &                                                                                                                   & 1        & 0.022                                                                        & 0.022                                                                          & 0.312                                                                        & 0.312                                                                          & 14.243          \\
REINVENT4         &                                                                                                                   & 0.722    & 0.130                                                                         & 0.180                                                                         & 0.272                                                                        & 0.377                                                                         & 10.977          &                                                                                                                   & 0.784    & 0.430                                                                         & 0.549                                                                         & 0.582                                                                        & 0.7423                                                                         & 9.3394           \\
DrugAssist &                                                                                                                   & 0.976    & 0.236                                                                        & 0.2418                                                                         & 0.304                                                                        & 0.3115                                                                         & 9.4182           &                                                                                                                   & 0.9779   & 0.4458                                                                       & 0.4559                                                                         & 0.5321                                                                       & 0.5441                                                                         & 9.3008           \\
GeLLM$^3$O\_M  &                                                                                                                   & 0.924    & 0.164                                                                        & 0.1775                                                                         & 0.354                                                                        & 0.3831                                                                         & 10.326           &                                                                                                                   & 0.896    & 0.054                                                                        & 0.0603                                                                         & 0.306                                                                        & 0.3415                                                                         & 10.3557          \\
GeLLM$^3$O\_L  &                                                                                                                   & 0.902    & 0.114                                                                        & 0.1264                                                                         & 0.294                                                                        & 0.3259                                                                         & 9.7367           &                                                                                                                   & 0.876    & 0.138                                                                        & 0.1575                                                                         & 0.506                                                                        & 0.5776                                                                         & 8.6119           \\
MolEditRL   &                                                                                                                   & 0.952    & \textbf{0.342}                                                               & \textbf{0.3592}                                                                & \textbf{0.514}                                                               & \textbf{0.5399}                                                                & \textbf{7.9953}  &                                                                                                                   & 0.984    & \textbf{0.634}                                                               & \textbf{0.6443}                                                                & \textbf{0.83}                                                                & \textbf{0.8435}                                                                & \textbf{6.9724}  \\ \hline
BioT5             & \multirow{7}{*}{MW${\uparrow}$}                                                                                   & 1        & 0                                                                            & 0                                                                              & 0.354                                                                        & 0.354                                                                          & 15.1392          & \multirow{7}{*}{SA${\downarrow}$}                                                                                 & 1        & 0                                                                            & 0                                                                              & 0.428                                                                        & 0.428                                                                          & 13.3982          \\
MolGen            &                                                                                                                   & 1        & 0.036                                                                        & 0.036                                                                          & 0.362                                                                        & 0.362                                                                          & 12.842           &                                                                                                                   & 1        & 0.016                                                                        & 0.016                                                                          & 0.306                                                                        & 0.306                                                                          & 14.1903          \\
REINVENT4         &                                                                                                                   & 0.654    & 0.068                                                                        & 0.104                                                                          & 0.34                                                                         & 0.5199                                                                         & 12.1462          &                                                                                                                   & 0.582    & 0.01                                                                         & 0.0172                                                                         & 0.024                                                                        & 0.0412                                                                         & 21.3067          \\
DrugAssist        &                                                                                                                   & 0.98     & 0.3166                                                                       & 0.3231                                                                         & 0.4068                                                                       & 0.4151                                                                         & 8.5049           &                                                                                                                   & 0.988    & 0.5371                                                                       & 0.5436                                                                         & 0.6092                                                                       & 0.6166                                                                         & 9.046            \\
GeLLM$^3$O\_M         &                                                                                                                   & 0.906    & 0.322                                                                        & 0.3554                                                                         & 0.572                                                                        & 0.6313                                                                         & 7.4434           &                                                                                                                   & 0.916    & 0.35                                                                         & 0.3821                                                                         & 0.662                                                                        & 0.7227                                                                         & 7.8452           \\
GeLLM$^3$O\_L         &                                                                                                                   & 0.884    & 0.232                                                                        & 0.2624                                                                         & 0.472                                                                        & 0.5339                                                                         & 6.9281           &                                                                                                                   & 0.888    & 0.238                                                                        & 0.268                                                                          & 0.64                                                                         & 0.7207                                                                         & 7.1048           \\
MolEditRL         &                                                                                                                   & 0.96     & \textbf{0.404}                                                               & \textbf{0.4208}                                                                & \textbf{0.856}                                                               & \textbf{0.8917}                                                                & \textbf{6.5879}  &                                                                                                                   & 0.988    & \textbf{0.628}                                                               & \textbf{0.6356}                                                                & \textbf{0.828}                                                               & \textbf{0.8381}                                                                & \textbf{7.1027}  \\ \hline
BioT5             & \multirow{5}{*}{\begin{tabular}[c]{@{}c@{}}Haccept${\downarrow}$\\ SA${\downarrow}$\end{tabular}}                 & 1        & 0                                                                            & 0                                                                              & 0.306                                                                        & 0.306                                                                          & 17.2356          & \multirow{5}{*}{\begin{tabular}[c]{@{}c@{}}QED${\uparrow}$\\ SA${\downarrow}$\end{tabular}}                       & 1        & 0                                                                            & 0                                                                              & 0.272                                                                        & 0.272                                                                          & 17.2143          \\
MolGen      &                                                                                                                   & 1        & 0.015                                                                        & 0.015                                                                          & 0.196                                                                        & 0.196                                                                          & 15.3624          &                                                                                                                   & 1        & 0.017                                                                        & 0.017                                                                          & 0.254                                                                        & 0.254                                                                          & 16.1782          \\
REINVENT4         &                                                                                                                   & 0.659    & 0.098                                                                         & 0.179                                                                         & 0.218                                                                        & 0.2781                                                                         & 13.3826          &                                                                                                                   & 0.701    & 0.143                                                                         & 0.2411                                                                         & 0.279                                                                        & 0.3491                                                                         & 11.702           \\
DrugAssist        &                                                                                                                   & 0.988    & 0.2425                                                                       & 0.2454                                                                         & 0.2725                                                                       & 0.2759                                                                         & 17.4653          &                                                                                                                   & 0.98     & 0.532                                                                        & 0.5429                                                                         & 0.578                                                                        & 0.5898                                                                         & 9.6831           \\
GeLLM$^3$O\_M         &                                                                                                                   & 0.872    & 0.132                                                                        & 0.1514                                                                         & 0.304                                                                        & 0.3486                                                                         & 13.6238          &                                                                                                                   & 0.882    & 0.012                                                                        & 0.0136                                                                         & 0.224                                                                        & 0.254                                                                          & 13.1972          \\
GeLLM$^3$O\_L         &                                                                                                                   & 0.92     & 0.118                                                                        & 0.1283                                                                         & 0.388                                                                        & 0.4217                                                                         & 11.685           &                                                                                                                   & 0.904    & 0.206                                                                        & 0.2279                                                                         & 0.586                                                                        & 0.6482                                                                         & 8.7621           \\
MolEditRL         &                                                                                                                   & 0.972    & \textbf{0.346}                                                               & \textbf{0.356}                                                                 & \textbf{0.51}                                                                & \textbf{0.5247}                                                                & \textbf{11.3769} &                                                                                                                   & 0.974    & \textbf{0.632}                                                               & \textbf{0.6489}                                                                & \textbf{0.788}                                                               & \textbf{0.809}                                                                 & \textbf{7.5393}  \\ \hline
BioT5             & \multirow{5}{*}{\begin{tabular}[c]{@{}c@{}}Haccept${\downarrow}$\\ LogP${\uparrow}$\end{tabular}}                 & 1        & 0                                                                            & 0                                                                              & 0.36                                                                         & 0.36                                                                           & 16.9514          & \multirow{5}{*}{\begin{tabular}[c]{@{}c@{}}Haccept${\downarrow}$\\ MW${\downarrow}$\end{tabular}}                 & 1        & 0                                                                            & 0                                                                              & 0.342                                                                        & 0.342                                                                          & 17.3017          \\
MolGen      &                                                                                                                   & 1        & 0.072                                                                        & 0.072                                                                          & 0.125                                                                        & 0.125                                                                          & 12.5741          &                                                                                                                   & 1        & 0.051                                                                        & 0.051                                                                          & 0.109                                                                        & 0.109                                                                          & 13.1683          \\
REINVENT4         &                                                                                                                   & 0.581    & 0.173                                                                         & 0.197                                                                         & 0.305                                                                        & 0.3854                                                                         & 12.3519          &                                                                                                                   & 0.562    & 0.091                                                                         & 0.1474                                                                         & 0.179                                                                        & 0.2381                                                                         & 12.7407           \\
DrugAssist        &                                                                                                                   & 0.984    & 0.302                                                                        & 0.308                                                                          & 0.43                                                                         & 0.437                                                                          & 12.3302          &                                                                                                                   & 0.984    & 0.2224                                                                       & 0.2261                                                                         & 0.2465                                                                       & 0.2505                                                                         & 19.3008          \\
GeLLM$^3$O\_M         &                                                                                                                   & 0.906    & 0.224                                                                        & 0.2472                                                                         & 0.486                                                                        & 0.5364                                                                         & 10.6775          &                                                                                                                   & 0.91     & 0.176                                                                        & 0.1934                                                                         & 0.37                                                                         & 0.4066                                                                         & 11.9159          \\
GeLLM$^3$O\_L         &                                                                                                                   & 0.904    & 0.13                                                                         & 0.1438                                                                         & 0.354                                                                        & 0.3916                                                                         & 11.2367          &                                                                                                                   & 0.932    & 0.128                                                                        & 0.1373                                                                         & 0.432                                                                        & 0.4635                                                                         & 11.6835          \\
MolEditRL         &                                                                                                                   & 0.946    & \textbf{0.316}                                                               & \textbf{0.334}                                                                 & \textbf{0.8}                                                                 & \textbf{0.8457}                                                                & \textbf{10.1124} &                                                                                                                   & 0.942    & \textbf{0.252}                                                               & \textbf{0.2675}                                                                & \textbf{0.66}                                                                & \textbf{0.7006}                                                                & \textbf{11.3373} \\ \hline
BioT5             & \multirow{5}{*}{\begin{tabular}[c]{@{}c@{}}DRD2${\downarrow}$\\ MW${\downarrow}$\\ SA${\downarrow}$\end{tabular}} & 1        & 0                                                                            & 0                                                                              & 0.136                                                                        & 0.136                                                                          & 24.3202          & \multirow{5}{*}{\begin{tabular}[c]{@{}c@{}}Haccept${\uparrow}$\\ MW${\uparrow}$\\ QED${\downarrow}$\end{tabular}} & 1        & 0                                                                            & 0                                                                              & 0.088                                                                        & 0.088                                                                          & 26.236           \\
MolGen      &                                                                                                                   & 1        & 0.039                                                                        & 0.039                                                                          & 0.128                                                                        & 0.128                                                                          & 11.9482          &                                                                                                                   & 1        & 0.033                                                                        & 0.033                                                                          & 0.117                                                                        & 0.117                                                                          & 13.7517          \\
REINVENT4         &                                                                                                                   & 0.522    & 0.093                                                                         & 0.2302                                                                         & 0.173                                                                        & 0.283                                                                         & 11.4873          &                                                                                                                   & 0.638    & 0.017                                                                         & 0.163                                                                         & 0.131                                                                        & 0.2057                                                                         & 12.085           \\
DrugAssist        &                                                                                                                   & 0.98     & 0.422                                                                        & 0.4306                                                                         & 0.464                                                                        & 0.4735                                                                         & 9.8929           &                                                                                                                   & 0.956    & 0.23                                                                         & 0.2406                                                                         & 0.27                                                                         & 0.2824                                                                         & 11.7156          \\
GeLLM$^3$O\_M         &                                                                                                                   & 0.9      & 0.08                                                                         & 0.0889                                                                         & 0.186                                                                        & 0.2067                                                                         & 12.348           &                                                                                                                   & 0.906    & 0.01                                                                         & 0.011                                                                          & 0.076                                                                        & 0.0839                                                                         & 16.2161          \\
GeLLM$^3$O\_L         &                                                                                                                   & 0.918    & 0.108                                                                        & 0.1176                                                                         & 0.288                                                                        & 0.3137                                                                         & 11.2519          &                                                                                                                   & 0.886    & 0.04                                                                         & 0.0451                                                                         & 0.084                                                                        & 0.0948                                                                         & 15.7028          \\
MolEditRL         &                                                                                                                   & 0.986    & \textbf{0.518}                                                               & \textbf{0.5254}                                                                & \textbf{0.724}                                                               & \textbf{0.7343}                                                                & \textbf{7.2758}  &                                                                                                                   & 0.958    & \textbf{0.43}                                                                & \textbf{0.4489}                                                                & \textbf{0.756}                                                               & \textbf{0.7891}                                                                & \textbf{9.7922}  \\ \hline
\vspace{-25pt}
\end{tabular}}
\label{maintable}
\end{table}

\subsection{Multi-Property Editing Performance}

\begin{figure}[htbp]
    \centering
    \begin{subfigure}[b]{0.5\textwidth}
      \centering
      \includegraphics[width=\textwidth]{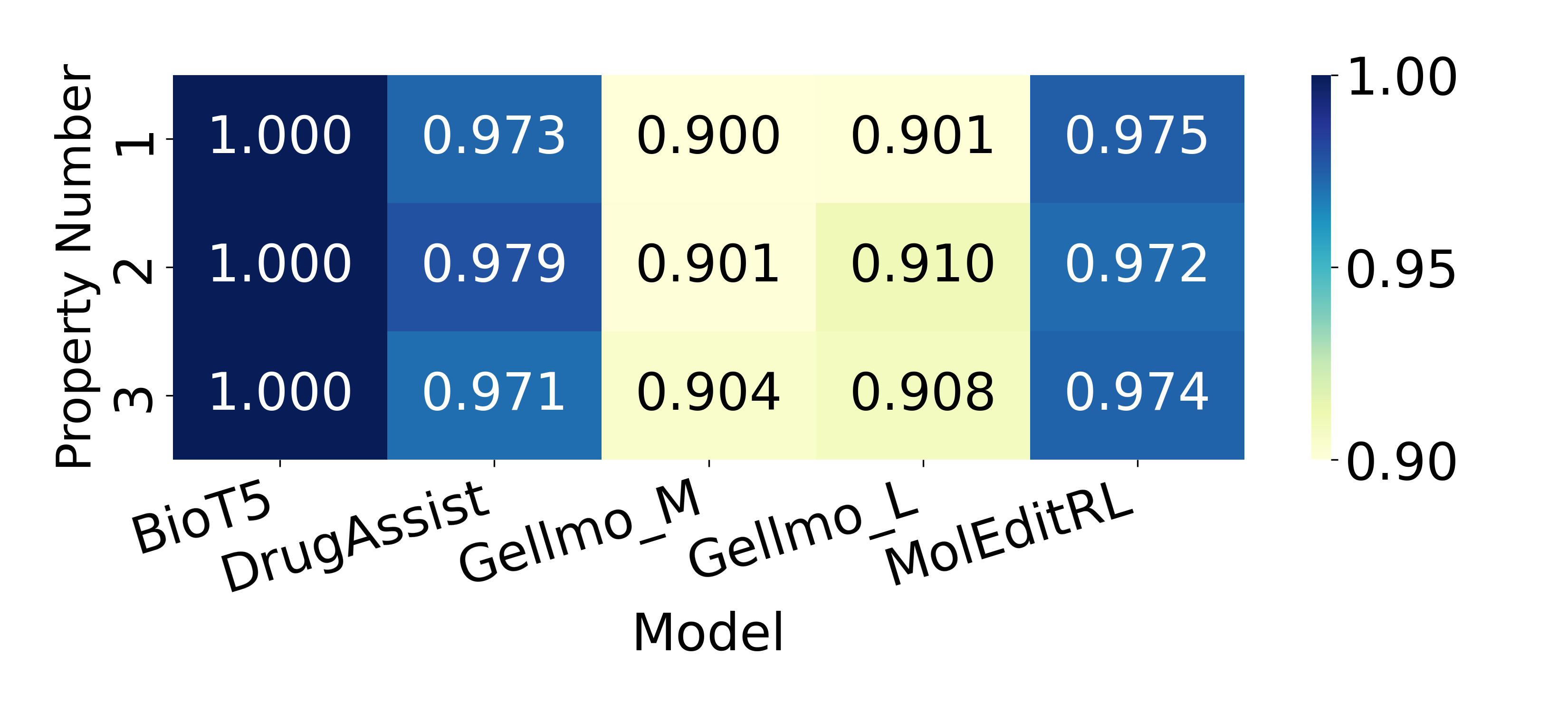}
      \vspace{-20pt}
      \caption{Mean validity.}
    \end{subfigure}%
    \begin{subfigure}[b]{0.5\textwidth}
      \centering
      \includegraphics[width=\textwidth]{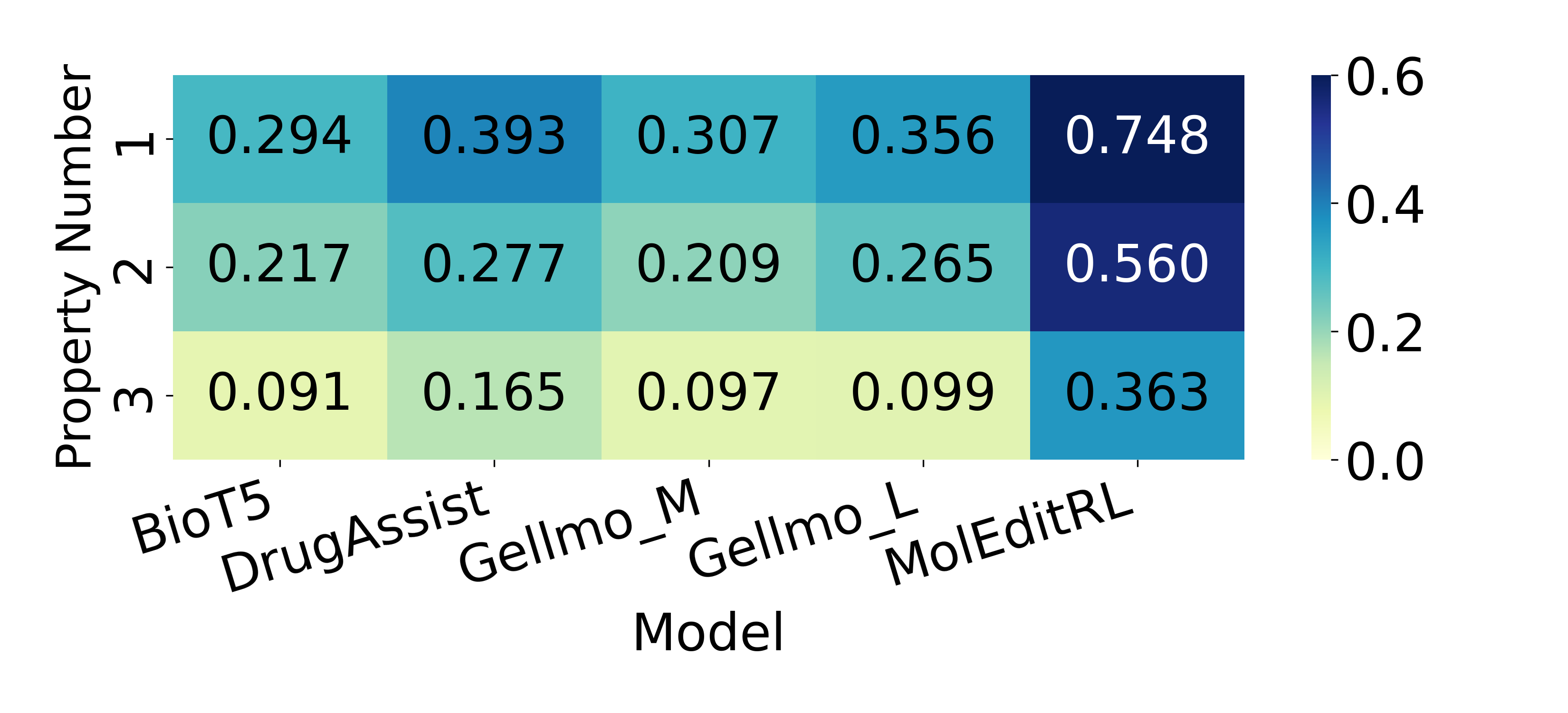}
      \vspace{-20pt}
      \caption{Mean accuracy at $\tau{=}0.15$ ($\text{Acc}_{\text{all}}$).}
    \end{subfigure}
    \caption{Performance by number of edited properties.}
    \vspace{-10pt}
    \label{fig:Ablation3}
  \end{figure}

To assess model robustness under increasing task complexity, we evaluate performance on molecular editing tasks involving 1, 2, or 3 simultaneous property changes. For each setting, 10 combinations of editing objectives are randomly sampled, and the results are averaged.
Figure~\ref{fig:Ablation3}(a) shows mean chemical validity, while (b) presents mean editing accuracy (Acc$_\text{all}$) at $\tau = 0.15$. As expected, accuracy drops for all models as the number of target properties increases, reflecting the challenge of jointly satisfying multiple constraints while preserving molecular structure. Some models maintain high validity but suffer from very low accuracy, indicating that generating chemically plausible molecules alone is insufficient for precise, property-aligned edits.
MolEditRL consistently outperforms all baselines across all settings. In the most difficult 3-property scenario, it achieves an average accuracy of 0.363, more than double the second-best baseline (DrugAssist, 0.165). These results demonstrate the effectiveness of our structure-aware diffusion framework and reinforcement learning fine-tuning in enabling scalable and precise instruction-based molecular editing.

\subsection{Effect of Fine-Tuning Strategies and KL Regularization}

\begin{figure}[htbp]
    \centering
    \begin{subfigure}[b]{0.5\textwidth}
      \centering
      \includegraphics[width=\textwidth]{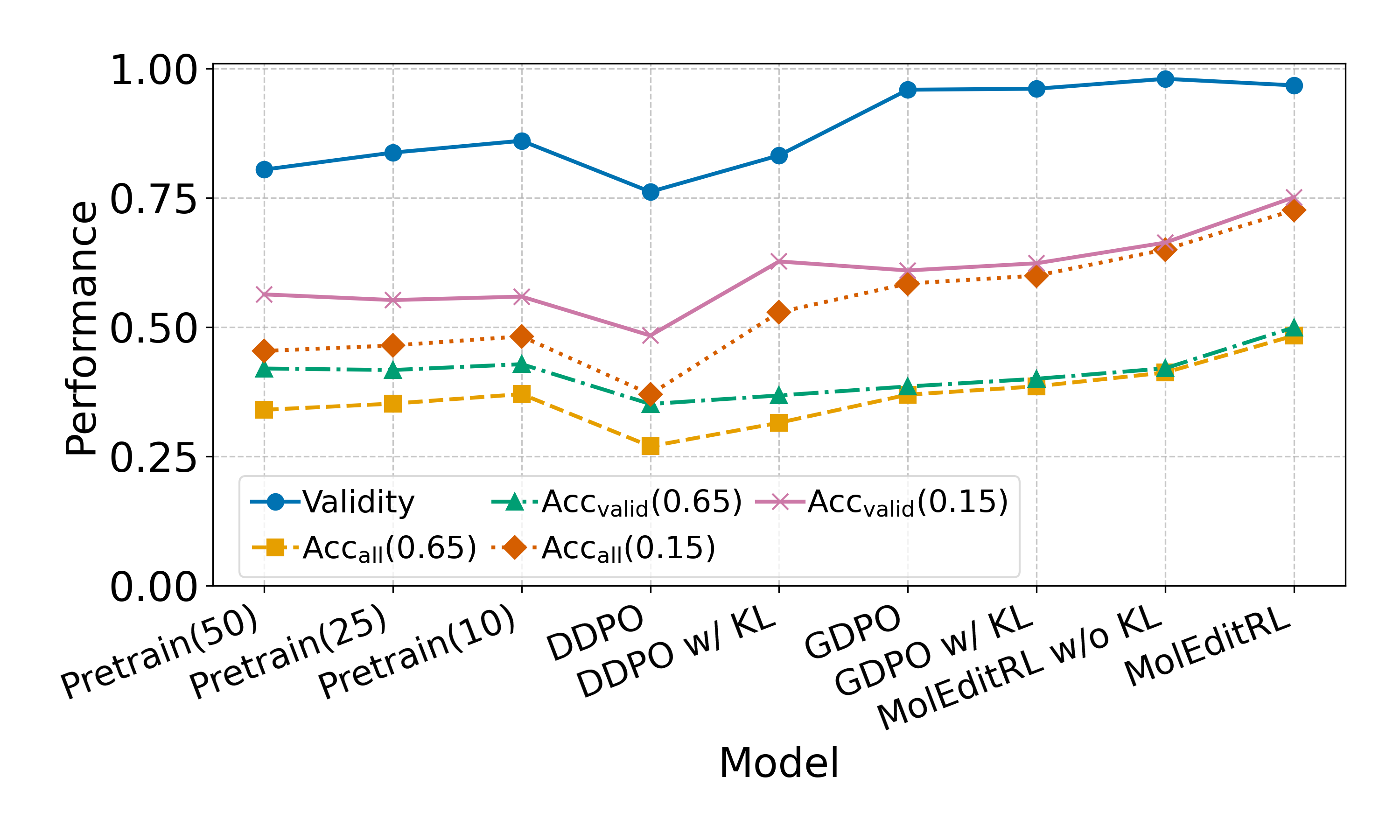}
    \vspace{-20pt}
      \caption{Accuracy across models.}
    \end{subfigure}%
    \begin{subfigure}[b]{0.5\textwidth}
      \centering
      \includegraphics[width=\textwidth]{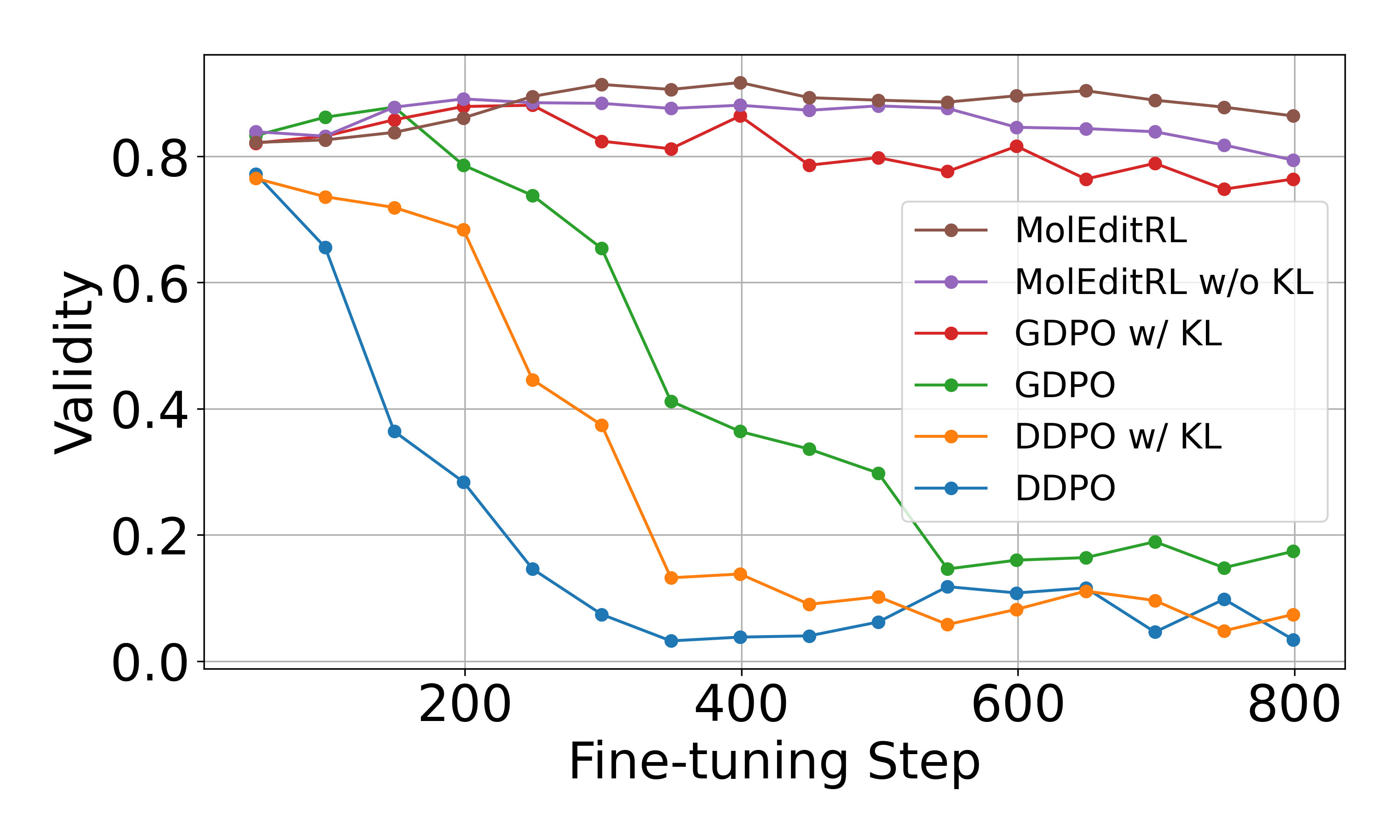}
    \vspace{-20pt}
      \caption{Training stability.}
    \end{subfigure}
    \caption{Impact of step size, fine-tuning strategy, and KL regularization.}
    \label{fig:Ablation1}
  \end{figure}

To assess the impact of key design choices, we conduct ablation studies on denoising step size, RL fine-tuning strategies, and KL regularization (Figure~\ref{fig:Ablation1}). Subfigure (a) shows overall editing accuracy and validity; (b) shows training stability. In (a), "Pretrain($x$)" refers to models without RL fine-tuning, where $x \in \{50, 25, 10\}$ denotes the denoising step size. Smaller $x$ (i.e., more reverse steps) generally improves accuracy but increases cost. In all main experiments, we use $t_s=50$ as the policy update stride, computing gradients only at timesteps $t$ where $t \bmod t_s = 0$ for efficiency.
We compare two RL fine-tuning strategies: DDPO\cite{black2023training}, which applies REINFORCE independently at each denoising step (Eq.\ref{eq:DDPO}), and GDPO\cite{liu2024graph}, which leverages the $x_0$-parameterization to optimize the final output (Eq.\ref{eq:GDPO}).
DDPO performs joint fine-tuning across all editing tasks using a single model. However, since intermediate molecules in the diffusion process are often chemically meaningless, step-wise optimization leads to instability and poor performance. GDPO improves stability by optimizing only the final output but requires training a separate model for each task, hindering scalability and generalization.
Neither method enforces structural constraints to preserve molecular validity and similarity during fine-tuning.
In contrast, MolEditRL introduces KL-regularized optimization over the entire diffusion process, enabling stable, structure-aware fine-tuning across diverse editing tasks. As shown in our experiments, MolEditRL consistently achieves higher accuracy and validity, demonstrating its effectiveness for instruction-driven molecular editing.
As shown in Figure~\ref{fig:Ablation1}(b), DDPO exhibits rapid degradation in chemical validity during training, regardless of the presence of KL regularization. This suggests that its step-wise optimization strategy (Eq.~\ref{eq:DDPO}) is inherently unstable for molecular editing. In contrast, GDPO shows improved stability when combined with KL regularization, which helps mitigate early overfitting. However, its task-specific fine-tuning design limits overall performance, which plateaus below that of MolEditRL.
MolEditRL maintains consistently high accuracy and validity throughout training. The addition of KL regularization further enhances stability without compromising performance, underscoring the robustness of our method across diverse and complex editing tasks.

\begin{figure}[t]
  \centering
  \begin{subfigure}[b]{0.48\textwidth}
    \centering
    \includegraphics[width=\textwidth]{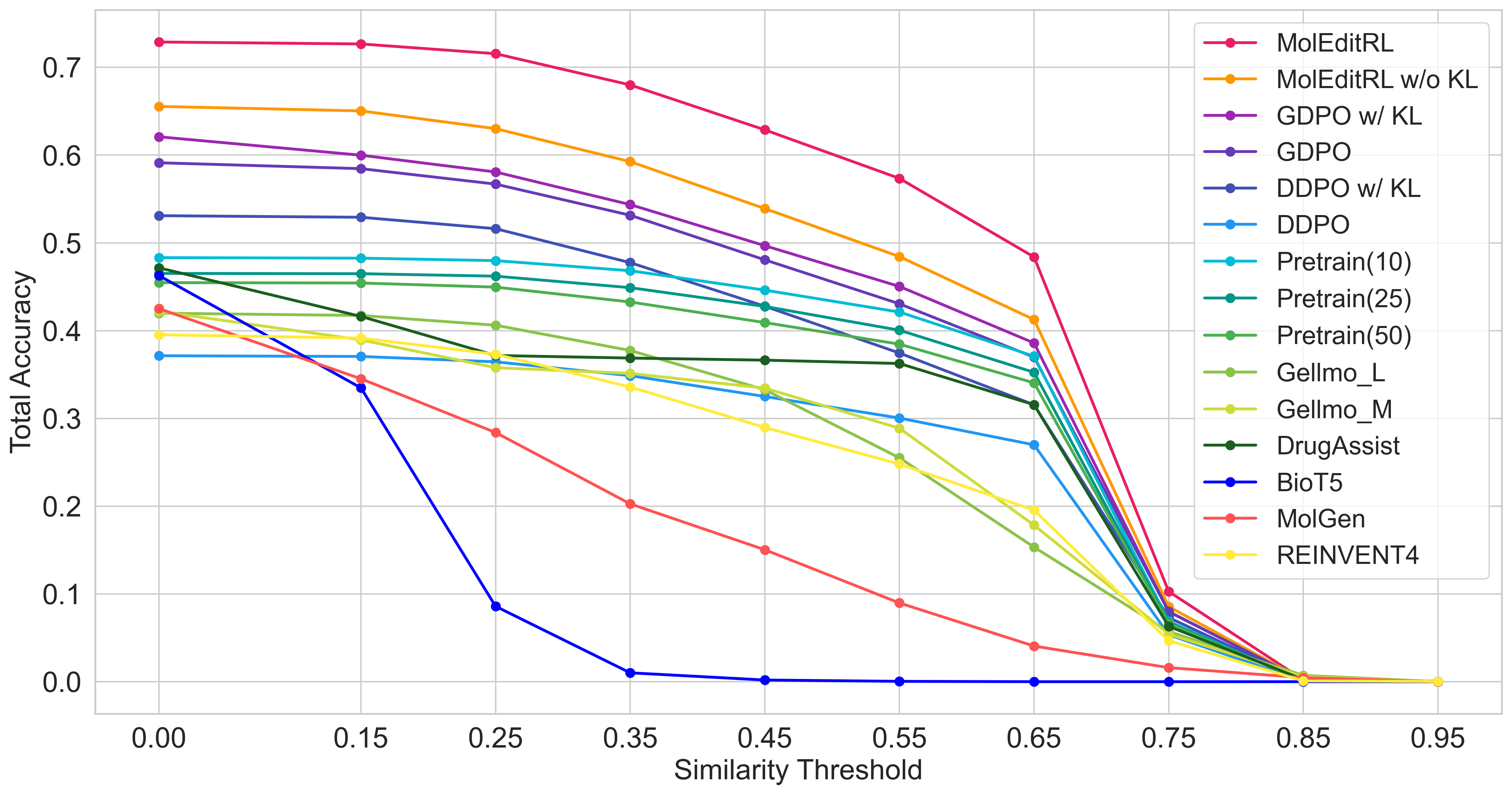}
    \vspace{-15pt}
    \caption{Accuracy under increasing similarity thresholds.}
  \end{subfigure}%
  \begin{subfigure}[b]{0.48\textwidth}
    \centering
    \includegraphics[width=\textwidth]{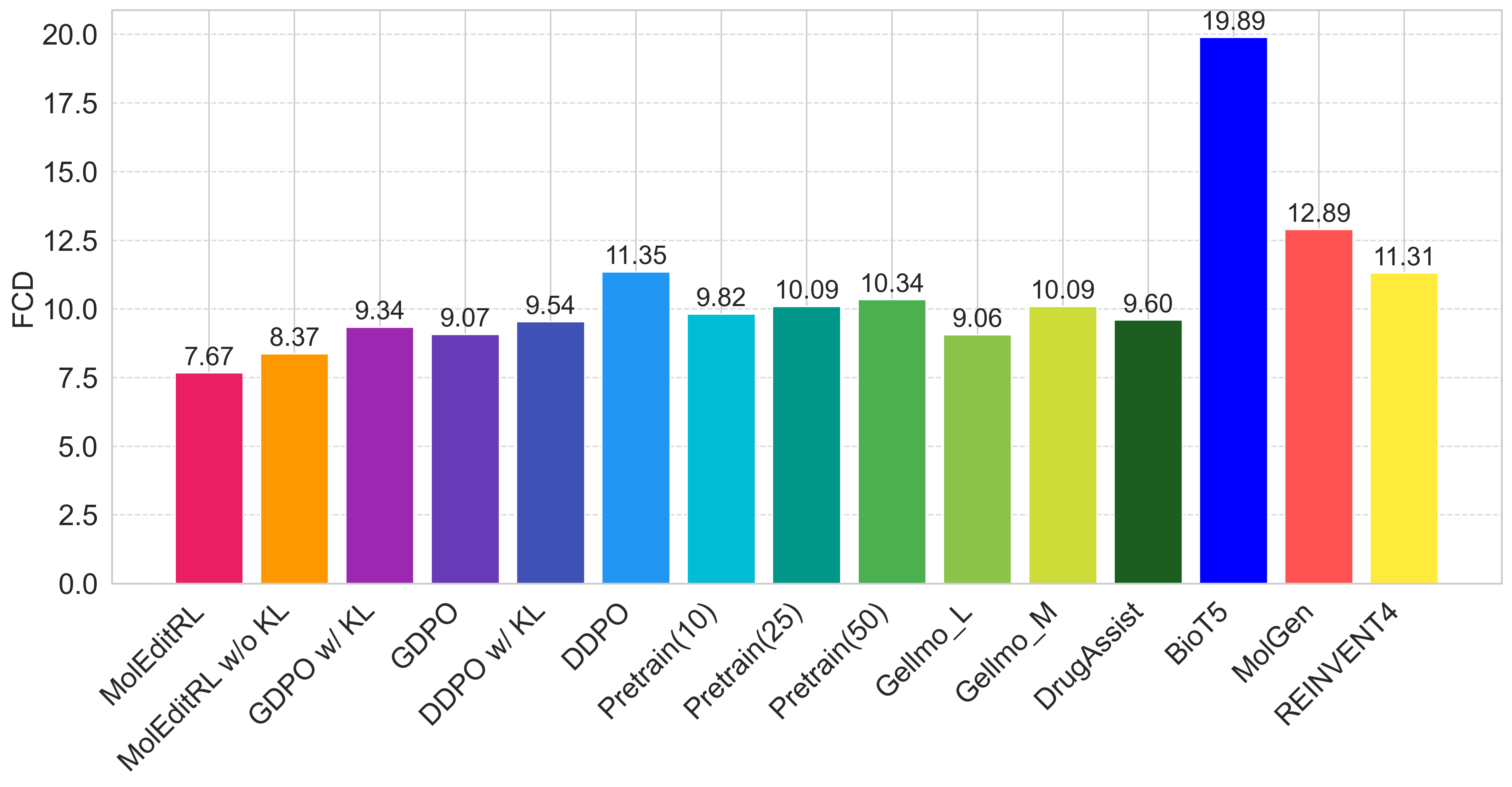}
    \vspace{-15pt}
    \caption{Fréchet ChemNet Distance across models.}
  \end{subfigure}
  \vspace{-5pt}
  \caption{Performance comparison under structural constraints.}
  \label{fig:Ablation2}
  \vspace{-12pt}
\end{figure}

\subsection{Structure Fidelity and Distributional Quality}

To comprehensively assess model performance, we evaluate both editing accuracy under structural similarity constraints and the distributional fidelity of the generated molecules. Figure~\ref{fig:Ablation2} reports results averaged over 20 single-property editing tasks.
Figure~\ref{fig:Ablation2}(a) shows $\text{Acc}_{\text{all}}$ across different Tanimoto similarity thresholds. MolEditRL consistently achieves the highest accuracy at all thresholds, demonstrating its ability to generate molecules that satisfy desired property changes while preserving structural similarity. In contrast, LLM-based baselines such as BioT5 and MolGen perform substantially worse, especially under stricter similarity constraints.
Figure~\ref{fig:Ablation2}(b) reports Fréchet ChemNet Distance (FCD) at a fixed threshold of 0.15. Lower FCD indicates better alignment between the distributions of generated and real molecules. Consistent with the accuracy results in (a), MolEditRL achieves the lowest FCD, highlighting its ability to produce chemically realistic and distributionally faithful molecules.

\begin{figure}[htbp]
    \centering
    \begin{subfigure}[b]{0.13\textwidth}
     \centering
     \includegraphics[width=\linewidth,height=5cm,keepaspectratio]{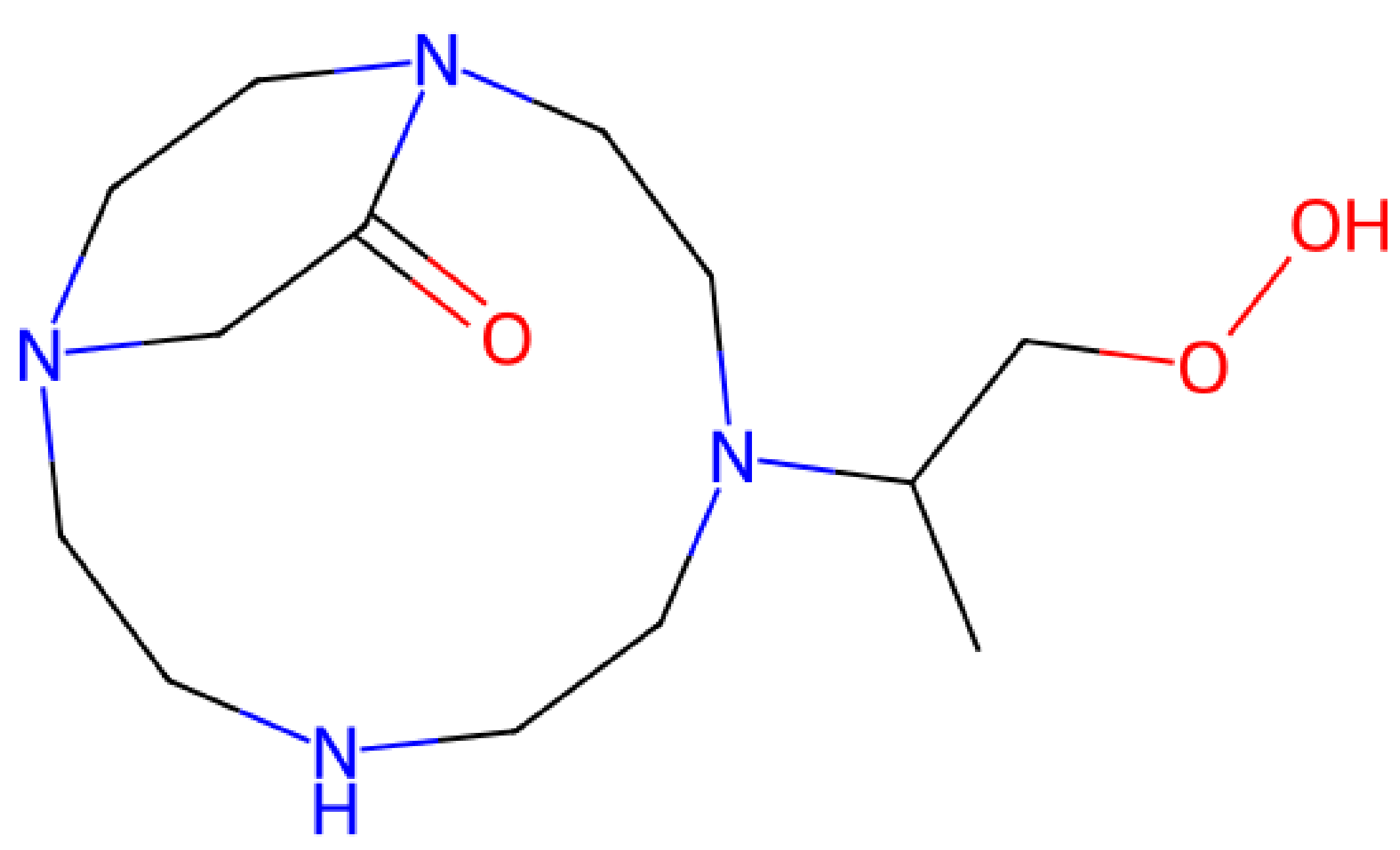}
     \caption{Source}
    \end{subfigure}%
    \hfill
    \begin{subfigure}[b]{0.41\textwidth}
     \centering
     \includegraphics[width=\linewidth,height=5cm,keepaspectratio]{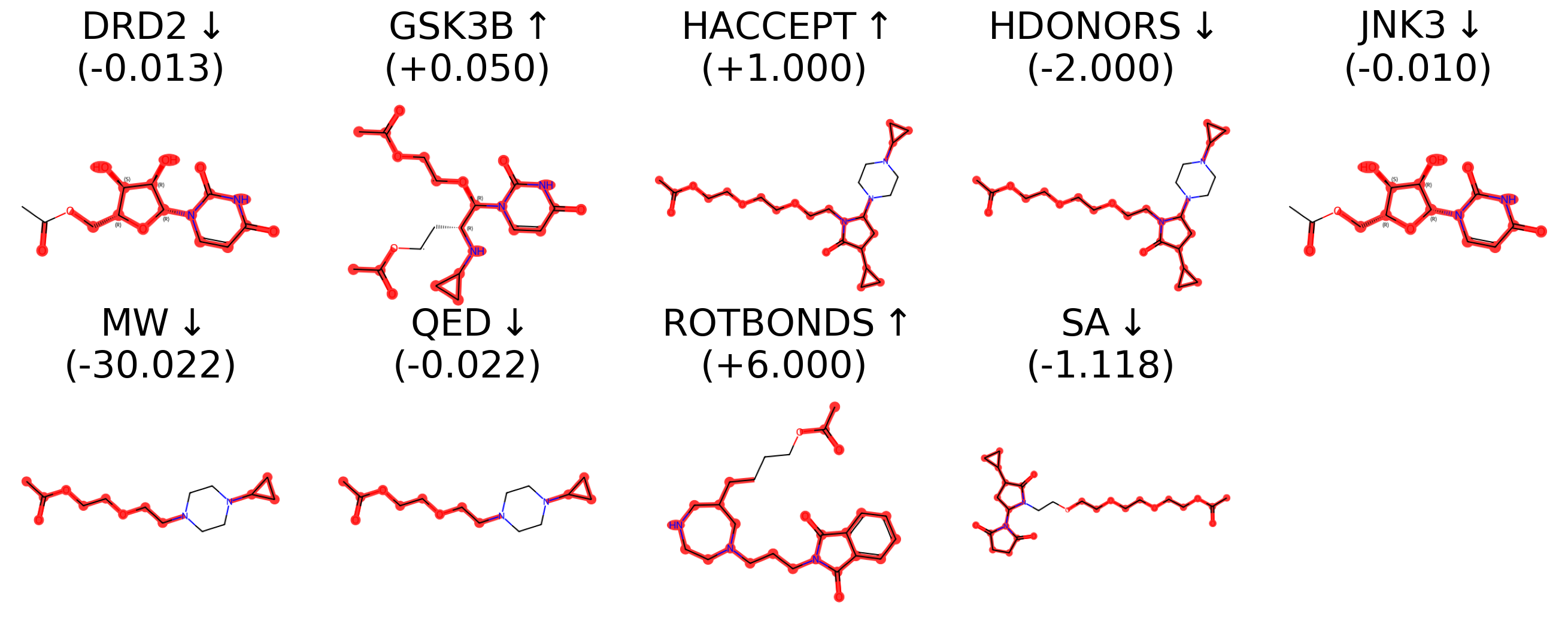}
     \caption{BioT5 (9/20 successful edits)}
    \end{subfigure}%
    \hfill
    \begin{subfigure}[b]{0.42\textwidth}
     \centering
     \includegraphics[width=\linewidth,height=5cm,keepaspectratio]{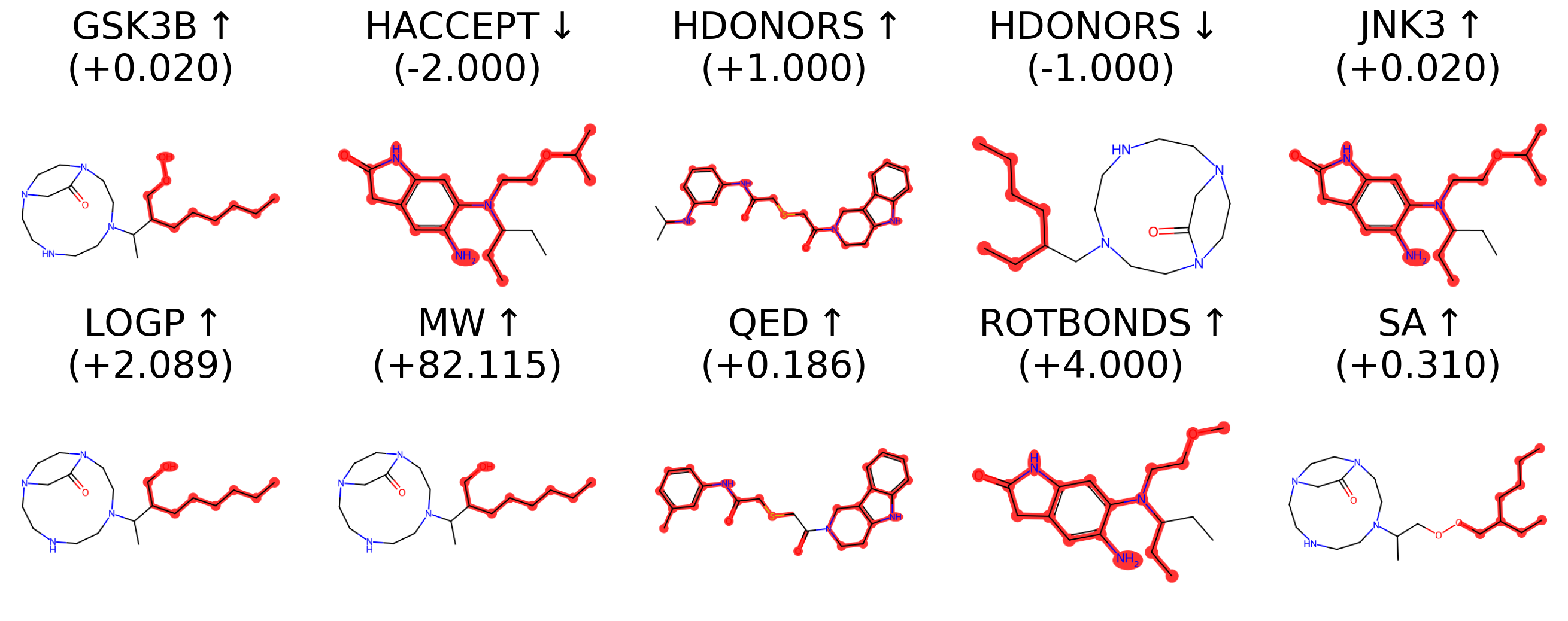}
     \caption{DrugAssist (10/20 successful edits)}
    \end{subfigure}
    \begin{subfigure}[b]{0.44\textwidth}
     \centering
     \includegraphics[width=\linewidth,height=5cm,keepaspectratio]{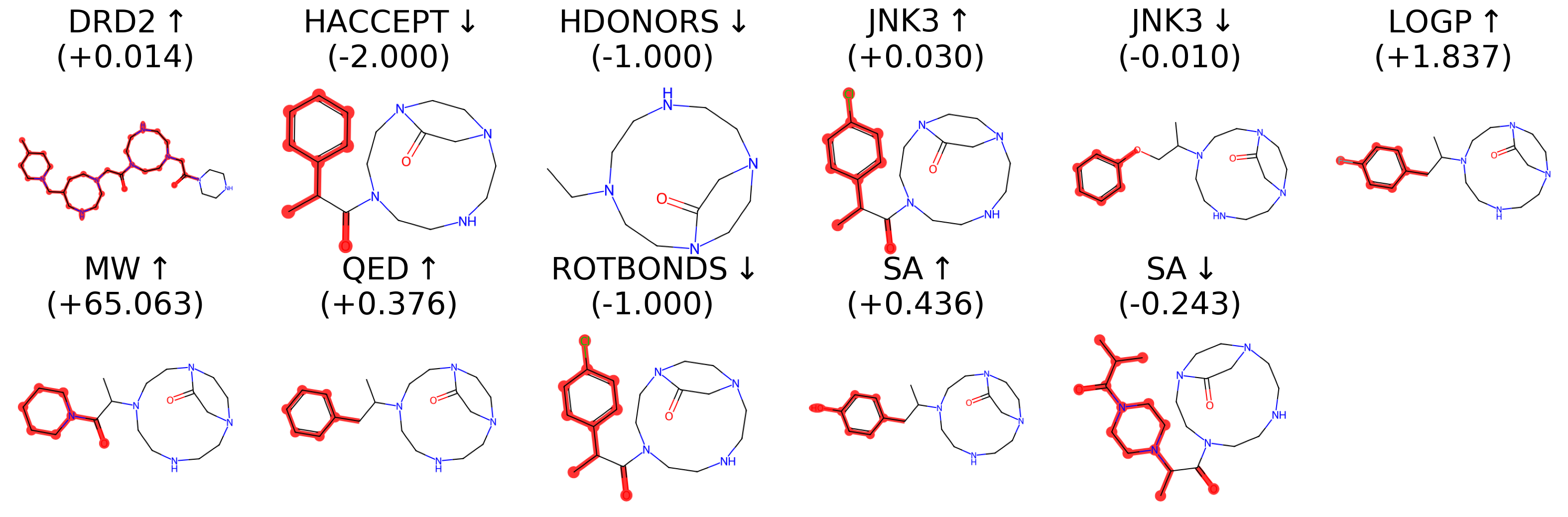}
     \caption{GeLLMO\_L (11/20 successful edits)}
    \end{subfigure}%
    \hfill
    \begin{subfigure}[b]{0.52\textwidth}
     \centering
     \includegraphics[width=\linewidth,height=5cm,keepaspectratio]{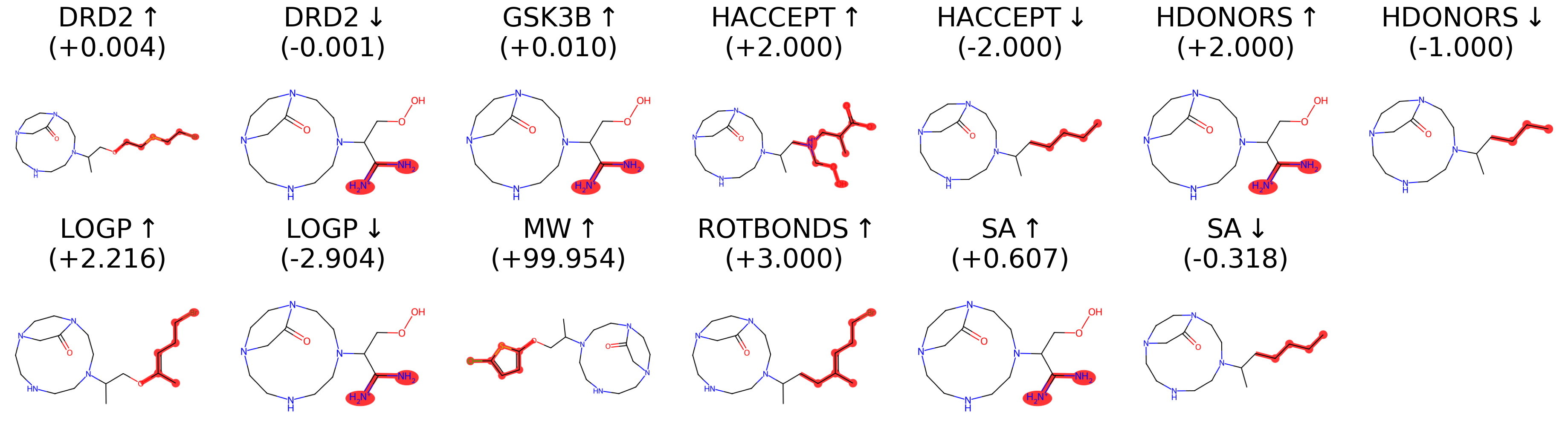}
     \caption{MolEditRL (13/20 successful edits)}
     \vspace{-4pt}
    \end{subfigure}
    \caption{Visualization of edits. Red highlights indicate structural changes from the source.}
    \label{fig:visualize}
   \end{figure}

\vspace{-10pt}
\subsection{Qualitative Analysis of Molecular Editing}
To better understand model behavior and editing fidelity, we visualize successful molecular edits from four representative models on 20 single-property editing tasks using the same source molecule. As shown in Figure~\ref{fig:visualize}, MolEditRL achieves the highest task success rate, successfully editing more properties than other baselines. Moreover, MolEditRL is the only model that consistently preserves the core scaffold of the source molecule across all edits, demonstrating its strong structural controllability.
In contrast, BioT5 and DrugAssist often generate structurally divergent molecules, with frequent scaffold disruptions that compromise similarity. GeLLMO\_L achieves better structural alignment, but still exhibit cases where major structural components are altered. These results qualitatively support our quantitative findings and highlight the effectiveness of MolEditRL’s structure-aware diffusion and full-trajectory reinforcement learning strategy in generating precise and high-fidelity molecular edits. Additional visualizations are provided in Appendix~\ref{sec:visualization}.

\vspace{-6pt}
\section{Conclusion}
\vspace{-6pt}
We introduce MolEditRL, a novel framework that integrates discrete graph diffusion with reinforcement learning to enable precise, structure-preserving molecular edits. It achieves state-of-the-art performance on the MolEdit-Instruct benchmark while using significantly fewer parameters.

\bibliographystyle{ACM-Reference-Format}
\bibliography{sample-base}


\begin{thebibliography}{39}


\ifx \showCODEN    \undefined \def \showCODEN     #1{\unskip}     \fi
\ifx \showISBNx    \undefined \def \showISBNx     #1{\unskip}     \fi
\ifx \showISBNxiii \undefined \def \showISBNxiii  #1{\unskip}     \fi
\ifx \showISSN     \undefined \def \showISSN      #1{\unskip}     \fi
\ifx \showLCCN     \undefined \def \showLCCN      #1{\unskip}     \fi
\ifx \shownote     \undefined \def \shownote      #1{#1}          \fi
\ifx \showarticletitle \undefined \def \showarticletitle #1{#1}   \fi
\ifx \showURL      \undefined \def \showURL       {\relax}        \fi
\providecommand\bibfield[2]{#2}
\providecommand\bibinfo[2]{#2}
\providecommand\natexlab[1]{#1}
\providecommand\showeprint[2][]{arXiv:#2}

\bibitem[Ar{\'u}s-Pous et~al\mbox{.}(2019)]%
        {arus2019randomized}
\bibfield{author}{\bibinfo{person}{Josep Ar{\'u}s-Pous}, \bibinfo{person}{Simon~Viet Johansson}, \bibinfo{person}{Oleksii Prykhodko}, \bibinfo{person}{Esben~Jannik Bjerrum}, \bibinfo{person}{Christian Tyrchan}, \bibinfo{person}{Jean-Louis Reymond}, \bibinfo{person}{Hongming Chen}, {and} \bibinfo{person}{Ola Engkvist}.} \bibinfo{year}{2019}\natexlab{}.
\newblock \showarticletitle{Randomized SMILES strings improve the quality of molecular generative models}.
\newblock \bibinfo{journal}{\emph{Journal of cheminformatics}}  \bibinfo{volume}{11} (\bibinfo{year}{2019}), \bibinfo{pages}{1--13}.
\newblock


\bibitem[Austin et~al\mbox{.}(2021)]%
        {austin2021structured}
\bibfield{author}{\bibinfo{person}{Jacob Austin}, \bibinfo{person}{Daniel~D Johnson}, \bibinfo{person}{Jonathan Ho}, \bibinfo{person}{Daniel Tarlow}, {and} \bibinfo{person}{Rianne Van Den~Berg}.} \bibinfo{year}{2021}\natexlab{}.
\newblock \showarticletitle{Structured denoising diffusion models in discrete state-spaces}.
\newblock \bibinfo{journal}{\emph{NeurIPS}}  \bibinfo{volume}{34} (\bibinfo{year}{2021}), \bibinfo{pages}{17981--17993}.
\newblock


\bibitem[Bento et~al\mbox{.}(2020)]%
        {bento2020open}
\bibfield{author}{\bibinfo{person}{A~Patr{\'\i}cia Bento}, \bibinfo{person}{Anne Hersey}, \bibinfo{person}{Eloy F{\'e}lix}, \bibinfo{person}{Greg Landrum}, \bibinfo{person}{Anna Gaulton}, \bibinfo{person}{Francis Atkinson}, \bibinfo{person}{Louisa~J Bellis}, \bibinfo{person}{Marleen De~Veij}, {and} \bibinfo{person}{Andrew~R Leach}.} \bibinfo{year}{2020}\natexlab{}.
\newblock \showarticletitle{An open source chemical structure curation pipeline using RDKit}.
\newblock \bibinfo{journal}{\emph{Journal of Cheminformatics}}  \bibinfo{volume}{12} (\bibinfo{year}{2020}), \bibinfo{pages}{1--16}.
\newblock


\bibitem[Black et~al\mbox{.}(2023)]%
        {black2023training}
\bibfield{author}{\bibinfo{person}{Kevin Black}, \bibinfo{person}{Michael Janner}, \bibinfo{person}{Yilun Du}, \bibinfo{person}{Ilya Kostrikov}, {and} \bibinfo{person}{Sergey Levine}.} \bibinfo{year}{2023}\natexlab{}.
\newblock \showarticletitle{Training diffusion models with reinforcement learning}.
\newblock \bibinfo{journal}{\emph{arXiv preprint arXiv:2305.13301}} (\bibinfo{year}{2023}).
\newblock


\bibitem[Chen et~al\mbox{.}(2021)]%
        {chen2021deep}
\bibfield{author}{\bibinfo{person}{Ziqi Chen}, \bibinfo{person}{Martin~Renqiang Min}, \bibinfo{person}{Srinivasan Parthasarathy}, {and} \bibinfo{person}{Xia Ning}.} \bibinfo{year}{2021}\natexlab{}.
\newblock \showarticletitle{A deep generative model for molecule optimization via one fragment modification}.
\newblock \bibinfo{journal}{\emph{Nature machine intelligence}} \bibinfo{volume}{3}, \bibinfo{number}{12} (\bibinfo{year}{2021}), \bibinfo{pages}{1040--1049}.
\newblock


\bibitem[Dalke et~al\mbox{.}(2018)]%
        {dalke2018mmpdb}
\bibfield{author}{\bibinfo{person}{Andrew Dalke}, \bibinfo{person}{Jerome Hert}, {and} \bibinfo{person}{Christian Kramer}.} \bibinfo{year}{2018}\natexlab{}.
\newblock \showarticletitle{mmpdb: An open-source matched molecular pair platform for large multiproperty data sets}.
\newblock \bibinfo{journal}{\emph{Journal of chemical information and modeling}} \bibinfo{volume}{58}, \bibinfo{number}{5} (\bibinfo{year}{2018}), \bibinfo{pages}{902--910}.
\newblock


\bibitem[De~Cao and Kipf(2018)]%
        {de2018molgan}
\bibfield{author}{\bibinfo{person}{Nicola De~Cao} {and} \bibinfo{person}{Thomas Kipf}.} \bibinfo{year}{2018}\natexlab{}.
\newblock \showarticletitle{MolGAN: An implicit generative model for small molecular graphs}.
\newblock \bibinfo{journal}{\emph{arXiv preprint arXiv:1805.11973}} (\bibinfo{year}{2018}).
\newblock


\bibitem[Dey et~al\mbox{.}(2025)]%
        {dey2025mathtt}
\bibfield{author}{\bibinfo{person}{Vishal Dey}, \bibinfo{person}{Xiao Hu}, {and} \bibinfo{person}{Xia Ning}.} \bibinfo{year}{2025}\natexlab{}.
\newblock \showarticletitle{$ \mathtt{GeLLM^{3O}}$: Generalizing Large Language Models for Multi-property Molecule Optimization}.
\newblock \bibinfo{journal}{\emph{arXiv preprint arXiv:2502.13398}} (\bibinfo{year}{2025}).
\newblock


\bibitem[Fang et~al\mbox{.}(2024)]%
        {fang2024domain}
\bibfield{author}{\bibinfo{person}{Yin Fang}, \bibinfo{person}{Ningyu Zhang}, \bibinfo{person}{Zhuo Chen}, \bibinfo{person}{Lingbing Guo}, \bibinfo{person}{Xiaohui Fan}, {and} \bibinfo{person}{Huajun Chen}.} \bibinfo{year}{2024}\natexlab{}.
\newblock \showarticletitle{Domain-Agnostic Molecular Generation with Chemical Feedback}. In \bibinfo{booktitle}{\emph{The Twelfth International Conference on Learning Representations}}.
\newblock


\bibitem[Fu et~al\mbox{.}(2021)]%
        {fu2021mimosa}
\bibfield{author}{\bibinfo{person}{Tianfan Fu}, \bibinfo{person}{Cao Xiao}, \bibinfo{person}{Xinhao Li}, \bibinfo{person}{Lucas~M Glass}, {and} \bibinfo{person}{Jimeng Sun}.} \bibinfo{year}{2021}\natexlab{}.
\newblock \showarticletitle{Mimosa: Multi-constraint molecule sampling for molecule optimization}. In \bibinfo{booktitle}{\emph{Proceedings of the AAAI Conference on Artificial Intelligence}}, Vol.~\bibinfo{volume}{35}. \bibinfo{pages}{125--133}.
\newblock


\bibitem[Hansch(1969)]%
        {hansch1969quantitative}
\bibfield{author}{\bibinfo{person}{Corwin Hansch}.} \bibinfo{year}{1969}\natexlab{}.
\newblock \showarticletitle{Quantitative approach to biochemical structure-activity relationships}.
\newblock \bibinfo{journal}{\emph{Accounts of chemical research}} \bibinfo{volume}{2}, \bibinfo{number}{8} (\bibinfo{year}{1969}), \bibinfo{pages}{232--239}.
\newblock


\bibitem[He et~al\mbox{.}(2021)]%
        {he2021molecular}
\bibfield{author}{\bibinfo{person}{Jiazhen He}, \bibinfo{person}{Huifang You}, \bibinfo{person}{Emil Sandstr{\"o}m}, \bibinfo{person}{Eva Nittinger}, \bibinfo{person}{Esben~Jannik Bjerrum}, \bibinfo{person}{Christian Tyrchan}, \bibinfo{person}{Werngard Czechtizky}, {and} \bibinfo{person}{Ola Engkvist}.} \bibinfo{year}{2021}\natexlab{}.
\newblock \showarticletitle{Molecular optimization by capturing chemist’s intuition using deep neural networks}.
\newblock \bibinfo{journal}{\emph{Journal of cheminformatics}}  \bibinfo{volume}{13} (\bibinfo{year}{2021}), \bibinfo{pages}{1--17}.
\newblock


\bibitem[Huang et~al\mbox{.}(2021)]%
        {huang2021therapeutics}
\bibfield{author}{\bibinfo{person}{Kexin Huang}, \bibinfo{person}{Tianfan Fu}, \bibinfo{person}{Wenhao Gao}, \bibinfo{person}{Yue Zhao}, \bibinfo{person}{Yusuf Roohani}, \bibinfo{person}{Jure Leskovec}, \bibinfo{person}{Connor~W Coley}, \bibinfo{person}{Cao Xiao}, \bibinfo{person}{Jimeng Sun}, {and} \bibinfo{person}{Marinka Zitnik}.} \bibinfo{year}{2021}\natexlab{}.
\newblock \showarticletitle{Therapeutics data commons: Machine learning datasets and tasks for drug discovery and development}.
\newblock \bibinfo{journal}{\emph{arXiv preprint arXiv:2102.09548}} (\bibinfo{year}{2021}).
\newblock


\bibitem[Hui et~al\mbox{.}(2022)]%
        {hui2022molecular}
\bibfield{author}{\bibinfo{person}{Chunngai Hui}, \bibinfo{person}{Zhuo Wang}, \bibinfo{person}{Shiping Wang}, {and} \bibinfo{person}{Chunfa Xu}.} \bibinfo{year}{2022}\natexlab{}.
\newblock \showarticletitle{Molecular editing in natural product synthesis}.
\newblock \bibinfo{journal}{\emph{Organic Chemistry Frontiers}} \bibinfo{volume}{9}, \bibinfo{number}{5} (\bibinfo{year}{2022}), \bibinfo{pages}{1451--1457}.
\newblock


\bibitem[Jin et~al\mbox{.}(2018)]%
        {jin2018junction}
\bibfield{author}{\bibinfo{person}{Wengong Jin}, \bibinfo{person}{Regina Barzilay}, {and} \bibinfo{person}{Tommi Jaakkola}.} \bibinfo{year}{2018}\natexlab{}.
\newblock \showarticletitle{Junction tree variational autoencoder for molecular graph generation}. In \bibinfo{booktitle}{\emph{ICML}}. PMLR, \bibinfo{pages}{2323--2332}.
\newblock


\bibitem[Jin et~al\mbox{.}(2020)]%
        {jin2020hierarchical}
\bibfield{author}{\bibinfo{person}{Wengong Jin}, \bibinfo{person}{Regina Barzilay}, {and} \bibinfo{person}{Tommi Jaakkola}.} \bibinfo{year}{2020}\natexlab{}.
\newblock \showarticletitle{Hierarchical generation of molecular graphs using structural motifs}. In \bibinfo{booktitle}{\emph{ICML}}. PMLR, \bibinfo{pages}{4839--4848}.
\newblock


\bibitem[Krenn et~al\mbox{.}(2020)]%
        {krenn2020self}
\bibfield{author}{\bibinfo{person}{Mario Krenn}, \bibinfo{person}{Florian H{\"a}se}, \bibinfo{person}{AkshatKumar Nigam}, \bibinfo{person}{Pascal Friederich}, {and} \bibinfo{person}{Alan Aspuru-Guzik}.} \bibinfo{year}{2020}\natexlab{}.
\newblock \showarticletitle{Self-referencing embedded strings (SELFIES): A 100\% robust molecular string representation}.
\newblock \bibinfo{journal}{\emph{Machine Learning: Science and Technology}} \bibinfo{volume}{1}, \bibinfo{number}{4} (\bibinfo{year}{2020}), \bibinfo{pages}{045024}.
\newblock


\bibitem[Kusner et~al\mbox{.}(2017)]%
        {kusner2017grammar}
\bibfield{author}{\bibinfo{person}{Matt~J Kusner}, \bibinfo{person}{Brooks Paige}, {and} \bibinfo{person}{Jos{\'e}~Miguel Hern{\'a}ndez-Lobato}.} \bibinfo{year}{2017}\natexlab{}.
\newblock \showarticletitle{Grammar variational autoencoder}. In \bibinfo{booktitle}{\emph{International conference on machine learning}}. PMLR, \bibinfo{pages}{1945--1954}.
\newblock


\bibitem[Le and Chawla(2024)]%
        {le2024utilizing}
\bibfield{author}{\bibinfo{person}{Khiem Le} {and} \bibinfo{person}{Nitesh~V Chawla}.} \bibinfo{year}{2024}\natexlab{}.
\newblock \showarticletitle{Utilizing Large Language Models in an iterative paradigm with domain feedback for molecule optimization}.
\newblock \bibinfo{journal}{\emph{arXiv preprint arXiv:2410.13147}} (\bibinfo{year}{2024}).
\newblock


\bibitem[Liu et~al\mbox{.}(2023)]%
        {liu2023multi}
\bibfield{author}{\bibinfo{person}{Shengchao Liu}, \bibinfo{person}{Weili Nie}, \bibinfo{person}{Chengpeng Wang}, \bibinfo{person}{Jiarui Lu}, \bibinfo{person}{Zhuoran Qiao}, \bibinfo{person}{Ling Liu}, \bibinfo{person}{Jian Tang}, \bibinfo{person}{Chaowei Xiao}, {and} \bibinfo{person}{Animashree Anandkumar}.} \bibinfo{year}{2023}\natexlab{}.
\newblock \showarticletitle{Multi-modal molecule structure--text model for text-based retrieval and editing}.
\newblock \bibinfo{journal}{\emph{Nature Machine Intelligence}} \bibinfo{volume}{5}, \bibinfo{number}{12} (\bibinfo{year}{2023}), \bibinfo{pages}{1447--1457}.
\newblock


\bibitem[Liu et~al\mbox{.}(2024b)]%
        {liu2024conversational}
\bibfield{author}{\bibinfo{person}{Shengchao Liu}, \bibinfo{person}{Jiongxiao Wang}, \bibinfo{person}{Yijin Yang}, \bibinfo{person}{Chengpeng Wang}, \bibinfo{person}{Ling Liu}, \bibinfo{person}{Hongyu Guo}, {and} \bibinfo{person}{Chaowei Xiao}.} \bibinfo{year}{2024}\natexlab{b}.
\newblock \showarticletitle{Conversational drug editing using retrieval and domain feedback}. In \bibinfo{booktitle}{\emph{The Twelfth International Conference on Learning Representations}}.
\newblock


\bibitem[Liu(2019)]%
        {liu2019roberta}
\bibfield{author}{\bibinfo{person}{Yinhan Liu}.} \bibinfo{year}{2019}\natexlab{}.
\newblock \showarticletitle{Roberta: A robustly optimized bert pretraining approach}.
\newblock \bibinfo{journal}{\emph{arXiv preprint arXiv:1907.11692}}  \bibinfo{volume}{364} (\bibinfo{year}{2019}).
\newblock


\bibitem[Liu et~al\mbox{.}(2024a)]%
        {liu2024graph}
\bibfield{author}{\bibinfo{person}{Yijing Liu}, \bibinfo{person}{Chao Du}, \bibinfo{person}{Tianyu Pang}, \bibinfo{person}{Chongxuan Li}, \bibinfo{person}{Min Lin}, {and} \bibinfo{person}{Wei Chen}.} \bibinfo{year}{2024}\natexlab{a}.
\newblock \showarticletitle{Graph diffusion policy optimization}.
\newblock \bibinfo{journal}{\emph{NeurIPS}}  \bibinfo{volume}{37} (\bibinfo{year}{2024}), \bibinfo{pages}{9585--9611}.
\newblock


\bibitem[Loeffler et~al\mbox{.}(2024)]%
        {loeffler2024reinvent}
\bibfield{author}{\bibinfo{person}{Hannes~H Loeffler}, \bibinfo{person}{Jiazhen He}, \bibinfo{person}{Alessandro Tibo}, \bibinfo{person}{Jon~Paul Janet}, \bibinfo{person}{Alexey Voronov}, \bibinfo{person}{Lewis~H Mervin}, {and} \bibinfo{person}{Ola Engkvist}.} \bibinfo{year}{2024}\natexlab{}.
\newblock \showarticletitle{Reinvent 4: Modern AI--driven generative molecule design}.
\newblock \bibinfo{journal}{\emph{Journal of Cheminformatics}} \bibinfo{volume}{16}, \bibinfo{number}{1} (\bibinfo{year}{2024}), \bibinfo{pages}{20}.
\newblock


\bibitem[Ma et~al\mbox{.}(2024)]%
        {ma2024rational}
\bibfield{author}{\bibinfo{person}{Chunhua Ma}, \bibinfo{person}{Craig~W Lindsley}, \bibinfo{person}{Junbiao Chang}, {and} \bibinfo{person}{Bin Yu}.} \bibinfo{year}{2024}\natexlab{}.
\newblock \bibinfo{title}{Rational molecular editing: a new paradigm in drug discovery}.
\newblock \bibinfo{numpages}{11459--11466}~pages.
\newblock


\bibitem[Noutahi et~al\mbox{.}(2024)]%
        {noutahi2024gotta}
\bibfield{author}{\bibinfo{person}{Emmanuel Noutahi}, \bibinfo{person}{Cristian Gabellini}, \bibinfo{person}{Michael Craig}, \bibinfo{person}{Jonathan~SC Lim}, {and} \bibinfo{person}{Prudencio Tossou}.} \bibinfo{year}{2024}\natexlab{}.
\newblock \showarticletitle{Gotta be SAFE: a new framework for molecular design}.
\newblock \bibinfo{journal}{\emph{Digital Discovery}} \bibinfo{volume}{3}, \bibinfo{number}{4} (\bibinfo{year}{2024}), \bibinfo{pages}{796--804}.
\newblock


\bibitem[Olivecrona et~al\mbox{.}(2017)]%
        {olivecrona2017molecular}
\bibfield{author}{\bibinfo{person}{Marcus Olivecrona}, \bibinfo{person}{Thomas Blaschke}, \bibinfo{person}{Ola Engkvist}, {and} \bibinfo{person}{Hongming Chen}.} \bibinfo{year}{2017}\natexlab{}.
\newblock \showarticletitle{Molecular de-novo design through deep reinforcement learning}.
\newblock \bibinfo{journal}{\emph{Journal of cheminformatics}}  \bibinfo{volume}{9} (\bibinfo{year}{2017}), \bibinfo{pages}{1--14}.
\newblock


\bibitem[Pei et~al\mbox{.}(2023)]%
        {pei2023biot5}
\bibfield{author}{\bibinfo{person}{Qizhi Pei}, \bibinfo{person}{Wei Zhang}, \bibinfo{person}{Jinhua Zhu}, \bibinfo{person}{Kehan Wu}, \bibinfo{person}{Kaiyuan Gao}, \bibinfo{person}{Lijun Wu}, \bibinfo{person}{Yingce Xia}, {and} \bibinfo{person}{Rui Yan}.} \bibinfo{year}{2023}\natexlab{}.
\newblock \showarticletitle{Biot5: Enriching cross-modal integration in biology with chemical knowledge and natural language associations}.
\newblock \bibinfo{journal}{\emph{arXiv preprint arXiv:2310.07276}} (\bibinfo{year}{2023}).
\newblock


\bibitem[Popova et~al\mbox{.}(2018)]%
        {popova2018deep}
\bibfield{author}{\bibinfo{person}{Mariya Popova}, \bibinfo{person}{Olexandr Isayev}, {and} \bibinfo{person}{Alexander Tropsha}.} \bibinfo{year}{2018}\natexlab{}.
\newblock \showarticletitle{Deep reinforcement learning for de novo drug design}.
\newblock \bibinfo{journal}{\emph{Science advances}} \bibinfo{volume}{4}, \bibinfo{number}{7} (\bibinfo{year}{2018}), \bibinfo{pages}{eaap7885}.
\newblock


\bibitem[Preuer et~al\mbox{.}(2018)]%
        {preuer2018frechet}
\bibfield{author}{\bibinfo{person}{Kristina Preuer}, \bibinfo{person}{Philipp Renz}, \bibinfo{person}{Thomas Unterthiner}, \bibinfo{person}{Sepp Hochreiter}, {and} \bibinfo{person}{Gunter Klambauer}.} \bibinfo{year}{2018}\natexlab{}.
\newblock \showarticletitle{Fr{\'e}chet ChemNet distance: a metric for generative models for molecules in drug discovery}.
\newblock \bibinfo{journal}{\emph{Journal of chemical information and modeling}} \bibinfo{volume}{58}, \bibinfo{number}{9} (\bibinfo{year}{2018}), \bibinfo{pages}{1736--1741}.
\newblock


\bibitem[Shi et~al\mbox{.}(2020)]%
        {shi2020graphaf}
\bibfield{author}{\bibinfo{person}{Chence Shi}, \bibinfo{person}{Minkai Xu}, \bibinfo{person}{Zhaocheng Zhu}, \bibinfo{person}{Weinan Zhang}, \bibinfo{person}{Ming Zhang}, {and} \bibinfo{person}{Jian Tang}.} \bibinfo{year}{2020}\natexlab{}.
\newblock \showarticletitle{Graphaf: a flow-based autoregressive model for molecular graph generation}.
\newblock \bibinfo{journal}{\emph{arXiv preprint arXiv:2001.09382}} (\bibinfo{year}{2020}).
\newblock


\bibitem[Sridharan et~al\mbox{.}(2024)]%
        {sridharan2024deep}
\bibfield{author}{\bibinfo{person}{Bhuvanesh Sridharan}, \bibinfo{person}{Animesh Sinha}, \bibinfo{person}{Jai Bardhan}, \bibinfo{person}{Rohit Modee}, \bibinfo{person}{Masahiro Ehara}, {and} \bibinfo{person}{U~Deva Priyakumar}.} \bibinfo{year}{2024}\natexlab{}.
\newblock \showarticletitle{Deep reinforcement learning in chemistry: A review}.
\newblock \bibinfo{journal}{\emph{Journal of Computational Chemistry}} \bibinfo{volume}{45}, \bibinfo{number}{22} (\bibinfo{year}{2024}), \bibinfo{pages}{1886--1898}.
\newblock


\bibitem[Wang et~al\mbox{.}(2022)]%
        {wang2022deep}
\bibfield{author}{\bibinfo{person}{Mingyang Wang}, \bibinfo{person}{Zhe Wang}, \bibinfo{person}{Huiyong Sun}, \bibinfo{person}{Jike Wang}, \bibinfo{person}{Chao Shen}, \bibinfo{person}{Gaoqi Weng}, \bibinfo{person}{Xin Chai}, \bibinfo{person}{Honglin Li}, \bibinfo{person}{Dongsheng Cao}, {and} \bibinfo{person}{Tingjun Hou}.} \bibinfo{year}{2022}\natexlab{}.
\newblock \showarticletitle{Deep learning approaches for de novo drug design: An overview}.
\newblock \bibinfo{journal}{\emph{Current opinion in structural biology}}  \bibinfo{volume}{72} (\bibinfo{year}{2022}), \bibinfo{pages}{135--144}.
\newblock


\bibitem[Wu et~al\mbox{.}(2024)]%
        {wu2024leveraging}
\bibfield{author}{\bibinfo{person}{Zhenxing Wu}, \bibinfo{person}{Odin Zhang}, \bibinfo{person}{Xiaorui Wang}, \bibinfo{person}{Li Fu}, \bibinfo{person}{Huifeng Zhao}, \bibinfo{person}{Jike Wang}, \bibinfo{person}{Hongyan Du}, \bibinfo{person}{Dejun Jiang}, \bibinfo{person}{Yafeng Deng}, \bibinfo{person}{Dongsheng Cao}, {et~al\mbox{.}}} \bibinfo{year}{2024}\natexlab{}.
\newblock \showarticletitle{Leveraging language model for advanced multiproperty molecular optimization via prompt engineering}.
\newblock \bibinfo{journal}{\emph{Nature Machine Intelligence}} (\bibinfo{year}{2024}), \bibinfo{pages}{1--11}.
\newblock


\bibitem[Xiang et~al\mbox{.}(2024)]%
        {xiang2024instruction}
\bibfield{author}{\bibinfo{person}{Yuran Xiang}, \bibinfo{person}{Haiteng Zhao}, \bibinfo{person}{Chang Ma}, {and} \bibinfo{person}{Zhi-Hong Deng}.} \bibinfo{year}{2024}\natexlab{}.
\newblock \showarticletitle{Instruction-Based Molecular Graph Generation with Unified Text-Graph Diffusion Model}.
\newblock \bibinfo{journal}{\emph{arXiv preprint arXiv:2408.09896}} (\bibinfo{year}{2024}).
\newblock


\bibitem[Xiong et~al\mbox{.}(2024)]%
        {xiong2024text}
\bibfield{author}{\bibinfo{person}{Yida Xiong}, \bibinfo{person}{Kun Li}, \bibinfo{person}{Weiwei Liu}, \bibinfo{person}{Jia Wu}, \bibinfo{person}{Bo Du}, \bibinfo{person}{Shirui Pan}, {and} \bibinfo{person}{Wenbin Hu}.} \bibinfo{year}{2024}\natexlab{}.
\newblock \showarticletitle{Text-Guided Multi-Property Molecular Optimization with a Diffusion Language Model}.
\newblock \bibinfo{journal}{\emph{arXiv preprint arXiv:2410.13597}} (\bibinfo{year}{2024}).
\newblock


\bibitem[Ye et~al\mbox{.}(2025)]%
        {ye2025drugassist}
\bibfield{author}{\bibinfo{person}{Geyan Ye}, \bibinfo{person}{Xibao Cai}, \bibinfo{person}{Houtim Lai}, \bibinfo{person}{Xing Wang}, \bibinfo{person}{Junhong Huang}, \bibinfo{person}{Longyue Wang}, \bibinfo{person}{Wei Liu}, {and} \bibinfo{person}{Xiangxiang Zeng}.} \bibinfo{year}{2025}\natexlab{}.
\newblock \showarticletitle{Drugassist: A large language model for molecule optimization}.
\newblock \bibinfo{journal}{\emph{Briefings in Bioinformatics}} \bibinfo{volume}{26}, \bibinfo{number}{1} (\bibinfo{year}{2025}), \bibinfo{pages}{bbae693}.
\newblock


\bibitem[You et~al\mbox{.}(2018)]%
        {you2018graph}
\bibfield{author}{\bibinfo{person}{Jiaxuan You}, \bibinfo{person}{Bowen Liu}, \bibinfo{person}{Zhitao Ying}, \bibinfo{person}{Vijay Pande}, {and} \bibinfo{person}{Jure Leskovec}.} \bibinfo{year}{2018}\natexlab{}.
\newblock \showarticletitle{Graph convolutional policy network for goal-directed molecular graph generation}.
\newblock \bibinfo{journal}{\emph{Advances in neural information processing systems}}  \bibinfo{volume}{31} (\bibinfo{year}{2018}).
\newblock


\bibitem[Zhou et~al\mbox{.}(2019)]%
        {zhou2019optimization}
\bibfield{author}{\bibinfo{person}{Zhenpeng Zhou}, \bibinfo{person}{Steven Kearnes}, \bibinfo{person}{Li Li}, \bibinfo{person}{Richard~N Zare}, {and} \bibinfo{person}{Patrick Riley}.} \bibinfo{year}{2019}\natexlab{}.
\newblock \showarticletitle{Optimization of molecules via deep reinforcement learning}.
\newblock \bibinfo{journal}{\emph{Scientific reports}} \bibinfo{volume}{9}, \bibinfo{number}{1} (\bibinfo{year}{2019}), \bibinfo{pages}{10752}.
\newblock


\end{thebibliography}

\clearpage
\appendix

\section*{Technical Appendices and Supplementary Material}

This appendix provides extended technical and experimental details that support the findings in the main paper. It is organized as follows:

(1) Appendix~\ref{sec:architecture}: Technical specifications of the discrete diffusion model used in MolEditRL, including architectural modules and embedding strategies.

(2) Appendix~\ref{sec:training}: Training setup and hyperparameter configurations for both pretraining and reinforcement learning stages.

(3) Appendix~\ref{sec:dataset_stats}: Details of the MolEdit-Instruct dataset construction, including property definitions, distribution statistics, and task coverage.

(4) Appendix~\ref{limitation}: Discussion of limitations and future directions, including extending to biomolecules and dialogue-based editing.

(5) Appendix~\ref{sec:unseen_properties}: Generalization experiments on properties not included in the pretraining data (e.g., BBBP, HIA, hERG), demonstrating MolEditRL’s adaptability.

(6) Appendix~\ref{sec:extended_singleprop} and Appendix~\ref{sec:extended_multitask}: Additional results on single-property and multi-property editing tasks to validate scalability and robustness.

(7) Appendix~\ref{sec:visualization}: More qualitative visualizations comparing structural edits across different models and editing tasks.

\section{Model Architecture}~\label{sec:architecture}

Table~\ref{tab:technical} details the structure-aware diffusion model used in MolEditRL. The architecture includes token and edge embeddings, a RoBERTa-based transformer with graph-aware attention, a discrete diffusion module for masked denoising, and a reinforcement learning component guided by property-based rewards. During inference, we employ top-$k$ sampling and apply policy updates at fixed stride intervals to improve efficiency.

\begin{table}[htbp]
\centering
\caption{Technical specifications of the structure-aware diffusion model in MolEditRL.}
\resizebox{\textwidth}{!}{
\begin{tabular}{>{\centering\arraybackslash}p{2.5cm}c>{\arraybackslash}p{8cm}}
\toprule
\textbf{Module Types} & \textbf{Dimensions} & \textbf{Structures} \\
\midrule
Input Layer & -- & Source Tokens [batch, seq\_len] $\rightarrow$ Concat \\
\midrule
Embedding & Vocab Size = 51,933 & TokenEmbedding [seq\_len, 768] + PositionEmbedding [seq\_len, 768] \\
\midrule
Edge Embedding & Edge Types = 6 & EdgeEmbedding [nodes, nodes, 768] \\
\midrule
\multirow{3}{*}{RoBERTa} & \multirow{3}{*}{12 Layers} & Input [batch, seq\_len, 768] \\
 &  & $\downarrow$ Self-Attention (12 × 64) $\rightarrow$ LayerNorm + Residual \\
 &  & $\downarrow$ FFN (768→3072→768) $\rightarrow$ LayerNorm + Residual \\
\midrule
\multirow{3}{*}{Diffusion} & \multirow{3}{*}{2000 steps} & Forward: Input $\rightarrow$ Masked Tokens \\
 &  & $\downarrow$ Reverse (stride = 50) \\
 &  & $\downarrow$ Denoising Network \\
\midrule
\multirow{2}{*}{Prediction} & seq\_len × 51,933 & AtomLogits [batch, seq\_len, 51933] \\
 & nodes × nodes × 6 & EdgeLogits [batch, nodes, nodes, 6] \\
\midrule
\multirow{2}{*}{Sampling} & \multirow{2}{*}{Top-k = 15} & Atom Categorical Sampling \\
 &  & Edge Structure Sampling \\
\midrule
\multirow{2}{*}{Property-Guided} & \multirow{2}{*}{--} & Reward Calculation (0, 0.2, 1.0) \\
 &  & Advantage Function $\rightarrow$ Loss Weighting \\
\bottomrule
\end{tabular}}
\label{tab:technical}
\end{table}

\section{Training and Hyperparameter Setup}~\label{sec:training}

We train MolEditRL on a multi-GPU cluster using PyTorch with Distributed Data Parallel (DDP). The model is initialized from a RoBERTa-base encoder with 12 layers, 12 heads, and hidden size 768. The tokenizer is extended to 51,933 tokens to accommodate molecular and instruction-specific vocabulary, and the embedding layer is resized accordingly.
For optimization, we use the AdamW optimizer with a learning rate of 5e-5, weight decay of 0.01, and a linear warm-up scheduler over 10,000 steps. Mixed precision (FP16) is enabled to reduce memory usage and accelerate training.
During pretraining, the model is trained for 100 epochs with a per-GPU batch size of 16. A discrete diffusion schedule with 2,000 denoising steps is used, following a mutual noise schedule $\beta_t = 1/(T - t)$ where $T=2000$. We apply word- and edge-level frequency weighting with sinusoidal modulation ($\lambda=0.3$) to guide denoising dynamics. The edge vocabulary includes 6 bond types.
During reinforcement learning fine-tuning, rewards are computed using property oracles (e.g., RDKit, TDC). We use top-$k$ sampling with $k=15$ and a temperature of 1.0 during evaluation. To improve efficiency, the policy is updated every $t_s = 50$ steps, resulting in 40 updates over the 2000-step diffusion process. For consistency, inference also runs for 40 denoising steps, starting from a fully masked graph and progressively reconstructing the final molecule.
All experiments were conducted on a single NVIDIA A6000 GPU using PyTorch and DGL. Pretraining on the MolEdit-Instruct dataset (3M examples) took approximately 100 hours. RL fine-tuning for each task required 1–2 hours.

\section{Dataset Statistics}~\label{sec:dataset_stats}

Our dataset is constructed following a procedure similar to DrugAssist~\cite{ye2025drugassist}, involving three main steps: (1) drug-like molecules are filtered from public databases such as ZINC and ChEMBL based on Lipinski’s Rule of Five; (2) Matched Molecular Pairs (MMP) are extracted using BRICS fragmentation to identify structurally similar molecule pairs with local edits; and (3) pairs showing significant property shifts are retained, and corresponding natural language instructions are generated to describe the desired property modifications.
Table~\ref{tab:property_stats} presents comprehensive statistics for each molecular property in our MolEdit dataset. The physicochemical properties in our dataset are carefully selected in accordance with Lipinski's Rule of Five, a key set of guidelines for drug-like molecules that includes constraints on molecular weight ($\leq$ 500 Da), LogP ($\leq$ 5), hydrogen bond donors ($\leq$ 5), and hydrogen bond acceptors ($\leq$ 10). These constraints are reflected in the value ranges of our dataset properties. The dataset covers both biological activity properties and physicochemical properties, each playing crucial roles in drug discovery:

\subsection{Biological Activity Properties}
\begin{itemize}
    \item \textbf{DRD2} (Dopamine D2 receptor): A key target in antipsychotic drug development, with values ranging from 0 to 1 indicating binding probability. Our dataset captures substantial changes in DRD2 activity, from minor adjustments ($\pm$0.050) to major shifts ($\pm$0.951), where positive values indicate decreased binding and negative values indicate increased binding.
    
    \item \textbf{GSK3$\beta$} (Glycogen synthase kinase-3 beta): An important target in treating neurological disorders, with values from 0 to 1 representing inhibition probability. The dataset includes modifications ranging from $\pm$0.050 to $\pm$0.750.
    
    \item \textbf{JNK3} (c-Jun N-terminal kinase 3): A target for neurodegenerative diseases, with values from 0 to 1 indicating inhibition probability. Property changes range from subtle ($\pm$0.030) to significant ($\pm$0.690).
\end{itemize}

\begin{table*}[htbp]
\centering
\caption{Statistical information for each single property editing task.}
\label{tab:property_stats}
\resizebox{\textwidth}{!}{
\begin{tabular}{lccccc}
\toprule
\textbf{Property} & \textbf{Direction} & \textbf{Pairs} & \textbf{$\Delta$ Range} & \textbf{Source Range} & \textbf{Target Range} \\
\midrule
\multirow{2}{*}{DRD2} & $\uparrow$ & 80,627 & [-0.951, -0.050] & [0.000, 0.944] & [0.050, 1.000] \\
 & $\downarrow$ & 80,627 & [0.050, 0.951] & [0.050, 1.000] & [0.000, 0.944] \\
\midrule
\multirow{2}{*}{GSK3$\beta$} & $\uparrow$ & 98,310 & [-0.750, -0.050] & [0.000, 0.940] & [0.052, 0.990] \\
 & $\downarrow$ & 98,310 & [0.050, 0.750] & [0.052, 0.990] & [0.000, 0.940] \\
\midrule
\multirow{2}{*}{JNK3} & $\uparrow$ & 94,131 & [-0.690, -0.030] & [0.000, 0.880] & [0.040, 0.990] \\
 & $\downarrow$ & 94,131 & [0.030, 0.690] & [0.040, 0.990] & [0.000, 0.880] \\
\midrule
\multirow{2}{*}{QED} & $\uparrow$ & 97,750 & [-0.794, -0.380] & [0.041, 0.564] & [0.438, 0.948] \\
 & $\downarrow$ & 98,249 & [0.380, 0.794] & [0.438, 0.948] & [0.050, 0.565] \\
\midrule
\multirow{2}{*}{SA} & $\uparrow$ & 90,192 & [-6.563, -0.700] & [1.059, 7.268] & [2.189, 7.999] \\
 & $\downarrow$ & 87,453 & [0.700, 6.104] & [2.277, 7.996] & [1.397, 7.268] \\
\midrule
\multirow{2}{*}{LogP} & $\uparrow$ & 89,088 & [-6.132, -2.625] & [-17.073, 2.369] & [-13.745, 5.000] \\
 & $\downarrow$ & 90,489 & [2.625, 6.132] & [-13.745, 5.000] & [-17.073, 2.372] \\
\midrule
\multirow{2}{*}{MW} & $\uparrow$ & 80,647 & [-195.744, -99.031] & [218.106, 399.216] & [336.084, 499.999] \\
 & $\downarrow$ & 79,712 & [99.031, 195.744] & [336.073, 499.994] & [218.094, 400.241] \\
\midrule
\multirow{2}{*}{HAccept} & $\uparrow$ & 98,562 & [-7.000, -2.000] & [0.000, 8.000] & [2.000, 10.000] \\
 & $\downarrow$ & 98,562 & [2.000, 7.000] & [2.000, 10.000] & [0.000, 8.000] \\
\midrule
\multirow{2}{*}{HDonors} & $\uparrow$ & 104,468 & [-5.000, -2.000] & [0.000, 3.000] & [2.000, 5.000] \\
 & $\downarrow$ & 104,468 & [2.000, 5.000] & [2.000, 5.000] & [0.000, 3.000] \\
\midrule
\multirow{2}{*}{RotBonds} & $\uparrow$ & 66,369 & [-9.000, -3.000] & [0.000, 7.000] & [3.000, 10.000] \\
 & $\downarrow$ & 65,806 & [3.000, 9.000] & [3.000, 10.000] & [0.000, 7.000] \\
\bottomrule
\end{tabular}
}
\end{table*}

\subsection{Physicochemical Properties}
\begin{itemize}
    \item \textbf{QED} (Quantitative Estimate of Drug-likeness): Ranges from 0 to 1, where higher values indicate better drug-likeness. Our dataset covers modifications from $\pm$0.380 to $\pm$0.794.
    
    \item \textbf{SA} (Synthetic Accessibility): Ranges from 1 to 10, where lower values indicate easier synthesis. The dataset includes substantial changes from $\pm$0.700 to $\pm$6.563.
    
    \item \textbf{MW} (Molecular Weight): A fundamental property ranging from 218 to 500 Da in our dataset, with modifications spanning $\pm$99.031 to $\pm$195.744 Da.
    
    \item \textbf{LogP} (Octanol-water partition coefficient): Measures lipophilicity, ranging from $-$17 to 5 in our dataset, with changes from $\pm$2.625 to $\pm$6.132.
    
    \item \textbf{HDONORS} (Hydrogen Bond Donors): Ranges from 0 to 5, with modifications of $\pm$2 to $\pm$5 donors.
    
    \item \textbf{HACCEPT} (Hydrogen Bond Acceptors): Ranges from 0 to 10, with changes of $\pm$2 to $\pm$7 acceptors.
    
    \item \textbf{ROTBONDS} (Rotatable Bonds): Ranges from 0 to 10, with modifications of $\pm$3 to $\pm$9 bonds, affecting molecular flexibility.
\end{itemize}

For each property, table~\ref{tab:property_stats} shows the number of molecular pairs, the range of property changes ($\Delta$ Range), and the value distributions in both source and target molecules. The $\pm$ notation indicates that changes occur in both directions -- positive values for property reduction and negative values for property increase.

\subsection{Natural Language Prompts}
Table~\ref{tab:editing_prompts} presents the natural language prompts designed for our single property editing tasks. For each of the ten molecular properties, we crafted two complementary prompts corresponding to property value increase and decrease. The prompts are purposefully designed to be clear and concise while maintaining chemical accuracy and relevance. For biological activity properties (DRD2, GSK3$\beta$, JNK3), the prompts emphasize binding affinity and inhibitory activity. For physicochemical properties, the prompts use specific chemical terminology (e.g., "hydrogen bond acceptors," "rotatable bonds") while remaining accessible. Some prompts, such as those for LogP, include additional context about the property's practical implications (e.g., "enhance its fat solubility" or "improve its water solubility"). Each prompt contains a [SMILE] placeholder that is replaced with the actual SMILES string of the molecule to be modified during the editing process.

\begin{table*}[htbp]
\centering
\caption{Natural language prompts for single property editing tasks.}
\label{tab:editing_prompts}
\resizebox{\textwidth}{!}{
\begin{tabular}{lll}
\toprule
\textbf{Property} & \textbf{Direction} & \textbf{Prompt} \\
\midrule
\multirow{2}{*}{DRD2} 
& $\uparrow$ & Optimize this molecule [SMILE] to increase its DRD2 binding affinity. \\
& $\downarrow$ & Help me reduce the DRD2 binding activity of molecule [SMILE]. \\
\midrule
\multirow{2}{*}{GSK3$\beta$} 
& $\uparrow$ & Help me optimize this molecule [SMILE] to improve its GSK3$\beta$ inhibitory activity. \\
& $\downarrow$ & Reduce the GSK3$\beta$ inhibition potential of this molecule [SMILE]. \\
\midrule
\multirow{2}{*}{JNK3} 
& $\uparrow$ & Enhance the JNK3 binding properties of molecule [SMILE]. \\
& $\downarrow$ & Make changes to lower the JNK3 binding affinity of molecule [SMILE]. \\
\midrule
\multirow{2}{*}{QED} 
& $\uparrow$ & Optimize the QED score of molecule [SMILE] to make it more drug-like. \\
& $\downarrow$ & Decrease the QED value of this molecule [SMILE]. \\
\midrule
\multirow{2}{*}{SA} 
& $\uparrow$ & Make this molecule [SMILE] harder to synthesize. \\
& $\downarrow$ & Make this molecule [SMILE] easier to synthesize. \\
\midrule
\multirow{2}{*}{LogP} 
& $\uparrow$ & Help me increase the LogP value of molecule [SMILE] to enhance its fat solubility. \\
& $\downarrow$ & Help me decrease the LogP value of molecule [SMILE] to improve its water solubility. \\
\midrule
\multirow{2}{*}{MW} 
& $\uparrow$ & Help me increase the molecular weight of this molecule [SMILE]. \\
& $\downarrow$ & Help me reduce the molecular weight of this molecule [SMILE]. \\
\midrule
\multirow{2}{*}{HAccept} 
& $\uparrow$ & Add more hydrogen bond acceptors to this molecule [SMILE]. \\
& $\downarrow$ & Reduce the number of hydrogen bond acceptors in molecule [SMILE]. \\
\midrule
\multirow{2}{*}{HDonors} 
& $\uparrow$ & Help me increase the number of H-bond donors in [SMILE]. \\
& $\downarrow$ & Help me decrease the H-bond donor count in this molecule [SMILE]. \\
\midrule
\multirow{2}{*}{RotBonds} 
& $\uparrow$ & Add more rotatable bonds to this molecule [SMILE]. \\
& $\downarrow$ & Reduce the number of rotatable bonds in molecule [SMILE]. \\
\bottomrule
\end{tabular}
}
\end{table*}

\section{Limitations and Future Work}~\label{limitation}

MolEditRL demonstrates strong and consistent performance in structure-preserving editing across a wide range of chemical properties, particularly on small to medium-sized molecules. While our current experiments focus on this regime, the underlying framework is designed to generalize and is expected to extend effectively to larger biomolecules, such as proteins or complex natural products, with minor adaptations.
The reinforcement learning component leverages property oracles (e.g., from RDKit and TDC) to guide optimization. These oracles validate MolEditRL’s effectiveness on widely studied molecular properties. For less-characterized or emerging attributes, task-specific predictors can be trained and integrated, enabling flexible extension of the framework to new property domains.
Looking ahead, we plan to explore interactive, dialogue-based molecular editing, enabling users to iteratively refine molecules via multi-turn natural language instructions. This direction could support more intuitive and human-centric workflows for molecular design and lead optimization.

\section{Generalization to Unseen Properties}~\label{sec:unseen_properties}

To evaluate the generalization capability of MolEditRL, we test it on three molecular properties—BBBP (blood-brain barrier permeability), HIA (human intestinal absorption), and hERG inhibition (cardiotoxicity risk)—that are entirely excluded from the pretraining dataset. These properties are not present in MolEdit-Instruct and are commonly used to assess pharmacokinetics and safety in early-stage drug design.
In this setting, the pretrained MolEditRL model is fine-tuned using reinforcement learning with property-specific oracles for each task, but without any prior exposure to these attributes during diffusion pretraining. This simulates a realistic scenario where new optimization objectives emerge post-deployment, and the model must adapt with minimal additional data.
As shown in Table~\ref{tab:unseen_properties}, MolEditRL significantly outperforms baseline methods on all tasks. It achieves the highest editing accuracy across both strict ($\tau=0.65$) and relaxed ($\tau=0.15$) structural similarity thresholds, while maintaining excellent chemical validity and the lowest FCD scores. These results demonstrate the model's ability to generalize to novel property objectives with strong structure-preserving performance.
This highlights one of the key advantages of MolEditRL: thanks to its structure-aware pretraining and RL-based fine-tuning design, it can be extended to new, previously unseen properties by plugging in task-specific reward oracles—without retraining from scratch.

\begin{table}[htbp]
\centering
\caption{Generalization to unseen properties. Results on editing three held-out molecular properties (BBBP, HIA, hERG) that are excluded from the pretraining dataset. }
\label{tab:unseen_properties}
\resizebox{\textwidth}{!}{%
\begin{tabular}{ccccccccccccccc}
\hline
Model             & Task                                                                                                              & Validity & \begin{tabular}[c]{@{}c@{}}\(\text{Acc}_{\text{all}}\)\\ (0.65)\end{tabular} & \begin{tabular}[c]{@{}c@{}}\(\text{Acc}_{\text{valid}}\)\\ (0.65)\end{tabular} & \begin{tabular}[c]{@{}c@{}}\(\text{Acc}_{\text{all}}\)\\ (0.15)\end{tabular} & \begin{tabular}[c]{@{}c@{}}\(\text{Acc}_{\text{valid}}\)\\ (0.15)\end{tabular} & FCD              & Task                                                                                                              & Validity & \begin{tabular}[c]{@{}c@{}}\(\text{Acc}_{\text{all}}\)\\ (0.65)\end{tabular} & \begin{tabular}[c]{@{}c@{}}\(\text{Acc}_{\text{valid}}\)\\ (0.65)\end{tabular} & \begin{tabular}[c]{@{}c@{}}\(\text{Acc}_{\text{all}}\)\\ (0.15)\end{tabular} & \begin{tabular}[c]{@{}c@{}}\(\text{Acc}_{\text{valid}}\)\\ (0.15)\end{tabular} & FCD              \\ \hline
BioT5           & \multirow{5}{*}{\textbf{BBBP${\downarrow}$}} & 1.0               & 0.0                          & 0.0                          & 0.276                        & 0.276                        & 25.7031         & \multirow{5}{*}{\textbf{BBBP${\uparrow}$}} & 1.0               & 0.0                          & 0.0                          & 0.452                        & 0.452                        & 21.4869         \\
DrugAssist      &                                      & 0.9719            & 0.3066                       & 0.3155                       & 0.3727                       & 0.3835                       & 7.4848          &                                    & 0.9879            & 0.3952                       & 0.4                          & 0.5423                       & 0.549                        & 7.9316          \\
GeLLM$^3$O\_M   &                                      & 0.9               & 0.17                         & 0.1889                       & 0.33                         & 0.3667                       & 7.8632          &                                    & 0.908             & 0.06                         & 0.0661                       & 0.356                        & 0.3921                       & 9.4042          \\
GeLLM$^3$O\_L     &                                      & 0.92              & 0.116                        & 0.1261                       & 0.292                        & 0.3174                       & 7.8038          &                                    & 0.91              & 0.254                        & 0.2791                       & 0.716                        & 0.7868                       & 7.2142          \\
MolEditRL             &                                      & 0.944             & \textbf{0.326}               & \textbf{0.337}               & \textbf{0.516}               & \textbf{0.5347}              & \textbf{7.0901} &                                    & 0.954             & \textbf{0.409}               & \textbf{0.418}               & \textbf{0.782}               & \textbf{0.8075}              & \textbf{6.7043} \\ \hline
BioT5           & \multirow{5}{*}{\textbf{HIA${\downarrow}$}}  & 1.0               & 0.0                          & 0.0                          & 0.426                        & 0.426                        & 15.246          & \multirow{5}{*}{\textbf{HIA${\uparrow}$}}  & 1.0               & 0.0                          & 0.0                          & 0.356                        & 0.356                        & 15.7829         \\
DrugAssist      &                                      & 0.982             & 0.344                        & 0.3503                       & 0.408                        & 0.4155                       & 7.0462          &                                    & 0.976             & 0.2725                       & 0.2793                       & 0.4068                       & 0.4168                       & 9.7314          \\
GeLLM$^3$O\_M   &                                      & 0.904             & 0.124                        & 0.1372                       & 0.286                        & 0.3164                       & 8.2303          &                                    & 0.904             & 0.348                        & 0.385                        & 0.662                        & 0.7323                       & 7.2866          \\
GeLLM$^3$O\_L     &                                      & 0.894             & 0.134                        & 0.1499                       & 0.382                        & 0.4273                       & 8.1844          &                                    & 0.922             & 0.222                        & 0.2408                       & 0.554                        & 0.6009                       & 7.7484          \\
MolEditRL             &                                      & 0.98              & \textbf{0.374}               & \textbf{0.3816}              & \textbf{0.628}               & \textbf{0.6396}              & \textbf{6.8726} &                                    & 0.986             & \textbf{0.446}               & \textbf{0.4523}              & \textbf{0.738}               & \textbf{0.7459}              & \textbf{6.7928} \\ \hline
BioT5           & \multirow{5}{*}{\textbf{hERG${\downarrow}$}} & 1.0               & 0.0                          & 0.0                          & 0.396                        & 0.396                        & 16.3127         & \multirow{5}{*}{\textbf{hERG${\uparrow}$}} & 1.0               & 0.0                          & 0.0                          & 0.368                        & 0.368                        & 15.0537         \\
DrugAssist      &                                      & 0.9659            & 0.2992                       & 0.3098                       & 0.4438                       & 0.4595                       & 7.815           &                                    & 0.9839            & 0.2294                       & 0.2331                       & 0.4064                       & 0.4131                       & 9.9241          \\
GeLLM$^3$O\_M   &                                      & 0.91              & 0.162                        & 0.178                        & 0.308                        & 0.3385                       & 7.9945          &                                    & 0.89              & 0.194                        & 0.218                        & 0.554                        & 0.6225                       & 7.5913          \\
GeLLM$^3$O\_L     &                                      & 0.914             & 0.144                        & 0.1575                       & 0.424                        & 0.4639                       & 8.0866          &                                    & 0.922             & 0.062                        & 0.0672                       & 0.364                        & 0.3948                       & 11.7363         \\
MolEditRL             &                                      & 0.986             & \textbf{0.474}               & \textbf{0.4807}              & \textbf{0.694}               & \textbf{0.7039}              & \textbf{6.0764} &                                    & 0.972             & \textbf{0.31}                & \textbf{0.3189}              & \textbf{0.59}                & \textbf{0.6147}              & \textbf{6.8304} \\ \hline
\end{tabular}}
\end{table}

\section{Extended Single-Property Results}~\label{sec:extended_singleprop}

Table~\ref{tab:singleprop_full} reports extended quantitative results for 10 representative single-property molecular editing tasks from the MolEdit-Instruct benchmark. 
MolEditRL consistently achieves the highest accuracy across both similarity thresholds, while maintaining high chemical validity and the lowest FCD scores across most tasks. This indicates strong structural fidelity and superior alignment with target property distributions. In contrast, baselines such as BioT5 and MolGen often generate valid molecules but fail to satisfy property and similarity constraints. REINVENT4 and DrugAssist perform moderately well but fall short in structural preservation and distributional realism. These detailed results further confirm the robustness and effectiveness of MolEditRL in single-property editing scenarios.

\begin{table}[t]
\centering
\caption{Extended results on single-property molecular editing tasks. Bold indicates best performance. Arrows ($\uparrow$, $\downarrow$) denote desired property increase or decrease.}
\label{tab:singleprop_full}
\resizebox{\textwidth}{!}{%
\begin{tabular}{ccccccccccccccc}
\hline
Model             & Task                                                                                                              & Validity & \begin{tabular}[c]{@{}c@{}}\(\text{Acc}_{\text{all}}\)\\ (0.65)\end{tabular} & \begin{tabular}[c]{@{}c@{}}\(\text{Acc}_{\text{valid}}\)\\ (0.65)\end{tabular} & \begin{tabular}[c]{@{}c@{}}\(\text{Acc}_{\text{all}}\)\\ (0.15)\end{tabular} & \begin{tabular}[c]{@{}c@{}}\(\text{Acc}_{\text{valid}}\)\\ (0.15)\end{tabular} & FCD              & Task                                                                                                              & Validity & \begin{tabular}[c]{@{}c@{}}\(\text{Acc}_{\text{all}}\)\\ (0.65)\end{tabular} & \begin{tabular}[c]{@{}c@{}}\(\text{Acc}_{\text{valid}}\)\\ (0.65)\end{tabular} & \begin{tabular}[c]{@{}c@{}}\(\text{Acc}_{\text{all}}\)\\ (0.15)\end{tabular} & \begin{tabular}[c]{@{}c@{}}\(\text{Acc}_{\text{valid}}\)\\ (0.15)\end{tabular} & FCD              \\ \hline
REINVENT4      & \multirow{7}{*}{\textbf{HACCEPT${\uparrow}$}}  & 0.524             & 0.19                           & 0.3626                         & 0.4                            & 0.7634                         & 11.0566         & \multirow{7}{*}{\textbf{HACCEPT${\downarrow}$}} & 0.704             & 0.19                           & 0.2699                         & 0.442                          & 0.6278                         & 11.7456         \\
MolGen         &                                        & 1.0               & 0.022                          & 0.022                          & 0.256                          & 0.256                          & 14.6826         &                                         & 1.0               & 0.004                          & 0.004                          & 0.404                          & 0.404                          & 14.943          \\
BioT5          &                                        & 1.0               & 0.0                            & 0.0                            & 0.148                          & 0.148                          & 30.3159         &                                         & 1.0               & 0.0                            & 0.0                            & 0.472                          & 0.472                          & 15.1916         \\
DrugAssist     &                                        & 0.9439            & 0.3467                         & 0.3673                         & 0.4429                         & 0.4692                         & 8.7609          &                                         & 0.9819            & 0.161                          & 0.1639                         & 0.3421                         & 0.3484                         & 11.8052         \\
Gellmo\_M      &                                        & 0.904             & 0.064                          & 0.0708                         & 0.15                           & 0.1659                         & 14.259          &                                         & 0.89              & 0.298                          & 0.3348                         & 0.524                          & 0.5888                         & 9.2035          \\
Gellmo\_L      &                                        & 0.89              & 0.07                           & 0.0787                         & 0.162                          & 0.182                          & 12.7839         &                                         & 0.914             & 0.178                          & 0.1947                         & 0.508                          & 0.5558                         & 8.8893          \\
MolEditRL      &                                        & 0.968             & \textbf{0.484}                 & \textbf{0.5}                   & \textbf{0.826}                 & \textbf{0.8533}                & \textbf{7.3163} &                                         & 0.974             & \textbf{0.388}                 & \textbf{0.3984}                & \textbf{0.712}                 & \textbf{0.731}                 & \textbf{9.0711} \\ \hline
REINVENT4      & \multirow{7}{*}{\textbf{SA${\uparrow}$}}       & 0.568             & 0.268                          & 0.4718                         & 0.548                          & 0.7648                         & 9.966           & \multirow{7}{*}{\textbf{MW${\downarrow}$}}      & 0.7581            & 0.3841                         & 0.5067                         & 0.6585                         & 0.8686                         & 7.7976          \\
MolGen         &                                        & 1.0               & 0.038                          & 0.038                          & 0.418                          & 0.418                          & 9.8619          &                                         & 1.0               & 0.016                          & 0.016                          & 0.432                          & 0.432                          & 14.4007         \\
BioT5          &                                        & 1.0               & 0.0                            & 0.0                            & 0.36                           & 0.36                           & 16.5037         &                                         & 1.0               & 0.0                            & 0.0                            & 0.348                          & 0.348                          & 17.2348         \\
DrugAssist     &                                        & 0.988             & 0.216                          & 0.2186                         & 0.294                          & 0.2976                         & 10.14           &                                         & 0.98              & 0.5391                         & 0.5501                         & 0.5872                         & 0.5992                         & 9.2859          \\
Gellmo\_M      &                                        & 0.91              & 0.12                           & 0.1319                         & 0.288                          & 0.3165                         & 8.9319          &                                         & 0.898             & 0.29                           & 0.3229                         & 0.564                          & 0.6281                         & 6.6778          \\
Gellmo\_L      &                                        & 0.912             & 0.104                          & 0.114                          & 0.28                           & 0.307                          & 8.9652          &                                         & 0.906             & 0.26                           & 0.287                          & 0.684                          & 0.755                          & 6.6965          \\
MolEditRL      &                                        & 0.95              & \textbf{0.49}                  & \textbf{0.5158}                & \textbf{0.776}                 & \textbf{0.8168}                & \textbf{7.5281} &                                         & 0.984             & \textbf{0.632}                 & \textbf{0.6423}                & \textbf{0.952}                 & \textbf{0.9675}                & \textbf{6.3935} \\ \hline
REINVENT4      & \multirow{7}{*}{\textbf{HDONORS${\uparrow}$}}  & 0.678             & 0.268                          & 0.3953                         & 0.458                          & 0.6755                         & 11.739          & \multirow{7}{*}{\textbf{DRD2${\downarrow}$}}    & 0.7               & 0.286                          & 0.4086                         & 0.418                          & 0.5971                         & 9.2662          \\
MolGen         &                                        & 1.0               & 0.022                          & 0.022                          & 0.243                          & 0.243                          & 13.4729         &                                         & 1.0               & 0.042                          & 0.042                          & 0.418                          & 0.418                          & 11.0476         \\
BioT5          &                                        & 1.0               & 0.0                            & 0.0                            & 0.144                          & 0.144                          & 23.7964         &                                         & 1.0               & 0.0                            & 0.0                            & 0.272                          & 0.272                          & 16.231          \\
DrugAssist     &                                        & 0.9319            & 0.3267                         & 0.3505                         & 0.4449                         & 0.4774                         & 8.4057          &                                         & 0.984             & 0.524                          & 0.5325                         & 0.57                           & 0.5793                         & 7.5935          \\
Gellmo\_M      &                                        & 0.898             & 0.044                          & 0.049                          & 0.092                          & 0.1024                         & 12.5726         &                                         & 0.922             & 0.136                          & 0.1475                         & 0.274                          & 0.2972                         & 9.3582          \\
Gellmo\_L      &                                        & 0.896             & 0.04                           & 0.0446                         & 0.1                            & 0.1116                         & 14.7698         &                                         & 0.916             & 0.132                          & 0.1441                         & 0.336                          & 0.3668                         & 9.8566          \\
MolEditRL      &                                        & 0.942             & \textbf{0.582}                 & \textbf{0.6178}                & \textbf{0.842}                 & \textbf{0.8938}                & \textbf{8.0114} &                                         & 0.986             & \textbf{0.656}                 & \textbf{0.6639}                & \textbf{0.72}                  & \textbf{0.7302}                & \textbf{6.549}  \\ \hline
REINVENT4      & \multirow{7}{*}{\textbf{LOGP${\uparrow}$}}     & 0.61              & 0.114                          & 0.1869                         & 0.36                           & 0.5902                         & 13.997          & \multirow{7}{*}{\textbf{LOGP${\downarrow}$}}    & 0.508             & 0.268                          & 0.5276                         & 0.424                          & 0.8346                         & 7.5582          \\
MolGen         &                                        & 1.0               & 0.094                          & 0.094                          & 0.474                          & 0.474                          & 10.5549         &                                         & 1.0               & 0.11                           & 0.11                           & 0.474                          & 0.474                          & 13.7158         \\
BioT5          &                                        & 1.0               & 0.0                            & 0.0                            & 0.492                          & 0.492                          & 27.5634         &                                         & 1.0               & 0.0                            & 0.0                            & 0.202                          & 0.202                          & 34.7724         \\
DrugAssist     &                                        & 0.964             & 0.382                          & 0.3963                         & 0.442                          & 0.4585                         & 11.3282         &                                         & 0.966             & 0.548                          & 0.5673                         & 0.604                          & 0.6253                         & 6.3703          \\
Gellmo\_M      &                                        & 0.89              & 0.374                          & 0.4202                         & 0.724                          & 0.8135                         & 6.878           &                                         & 0.906             & 0.004                          & 0.0044                         & 0.224                          & 0.2472                         & 12.9582         \\
Gellmo\_L      &                                        & 0.91              & 0.268                          & 0.2945                         & 0.59                           & 0.6484                         & 6.7566          &                                         & 0.918             & 0.15                           & 0.1634                         & 0.444                          & 0.4837                         & 7.7802          \\
MolEditRL      &                                        & 0.964             & \textbf{0.578}                 & \textbf{0.5996}                & \textbf{0.91}                  & \textbf{0.944}                 & \textbf{6.0118} &                                         & 0.972             & \textbf{0.71}                  & \textbf{0.7305}                & \textbf{0.94}                  & \textbf{0.9671}                & \textbf{5.1015} \\ \hline
REINVENT4      & \multirow{7}{*}{\textbf{ROTBONDS${\uparrow}$}} & 0.61              & 0.112                          & 0.1836                         & 0.384                          & 0.6295                         & 11.671          & \multirow{7}{*}{\textbf{QED${\downarrow}$}}     & 0.652             & 0.15                           & 0.2301                         & 0.312                          & 0.4785                         & 11.1066         \\
MolGen         &                                        & 1.0               & 0.084                          & 0.084                          & 0.356                          & 0.356                          & 11.3428         &                                         & 1.0               & 0.024                          & 0.024                          & 0.421                          & 0.421                          & 10.9996         \\
BioT5          &                                        & 1.0               & 0.0                            & 0.0                            & 0.306                          & 0.306                          & 16.85           &                                         & 1.0               & 0.0                            & 0.0                            & 0.374                          & 0.374                          & 15.7723         \\
DrugAssist     &                                        & 0.9537            & 0.1469                         & 0.154                          & 0.2716                         & 0.2848                         & 10.7588         &                                         & 0.9859            & 0.1044                         & 0.1059                         & 0.247                          & 0.2505                         & 10.8724         \\
Gellmo\_M      &                                        & 0.888             & 0.072                          & 0.0811                         & 0.16                           & 0.1802                         & 12.1059         &                                         & 0.924             & 0.012                          & 0.013                          & 0.15                           & 0.1623                         & 15.4165         \\
Gellmo\_L      &                                        & 0.888             & 0.098                          & 0.1104                         & 0.218                          & 0.2455                         & 10.0684         &                                         & 0.904             & 0.088                          & 0.0973                         & 0.218                          & 0.2412                         & 10.3736         \\
MolEditRL      &                                        & 0.934             & \textbf{0.392}                 & \textbf{0.4197}                & \textbf{0.764}                 & \textbf{0.818}                 & \textbf{7.2532} &                                         & 0.948             & \textbf{0.612}                 & \textbf{0.6456}                & \textbf{0.894}                 & \textbf{0.943}                 & \textbf{6.9314} \\ \hline
\end{tabular}}
\end{table}

\section{Extended Multi-Property Results}~\label{sec:extended_multitask}

Table~\ref{tab:multitask_full} presents detailed evaluation results on multi-property molecular editing tasks from the MolEdit-Instruct benchmark. Each task involves optimizing 2 to 4 chemical properties simultaneously, reflecting practical constraints encountered in real-world molecular design.
MolEditRL consistently achieves strong performance across all multi-property tasks, demonstrating its ability to balance complex property requirements while preserving molecular validity and structural similarity. The results confirm its robustness under increasingly constrained and realistic editing scenarios.
The property combinations in these tasks are carefully selected to reflect common design goals in medicinal chemistry. For example, tasks like (HACCEPT${\downarrow}$, HDONORS${\downarrow}$) aim to reduce molecular polarity, which is essential for improving membrane permeability and bioavailability. (LOGP${\downarrow}$, ROTBONDS${\downarrow}$) targets molecules with lower lipophilicity and rigidity, which improves metabolic stability and reduces off-target binding. On the other hand, combinations such as (MW${\uparrow}$, QED${\downarrow}$) simulate early-stage exploration of larger, less drug-like molecules, often relevant in hit expansion or macrocycle design. Biologically motivated combinations like (DRD2${\downarrow}$, GSK3$\beta$${\uparrow}$) reflect efforts to reduce off-target dopamine receptor activity while enhancing GSK3$\beta$ inhibition, a common challenge in polypharmacology. Furthermore, high-complexity tasks such as (GSK3$\beta$${\uparrow}$, HDONORS${\uparrow}$, QED${\downarrow}$, SA${\uparrow}$) require optimizing target activity while managing solubility, drug-likeness, and synthetic complexity—mirroring real trade-offs in lead optimization pipelines.
These results collectively showcase MolEditRL’s effectiveness not only in individual property edits but also in realistic, multi-objective optimization scenarios critical for practical drug development.

\begin{table}[t]
\centering
\label{tab:multitask_full}
\caption{Extended results on multi-property molecular editing tasks. Bold indicates best performance. Arrows ($\uparrow$, $\downarrow$) denote desired property increase or decrease.}
\label{tab:multitask_full}
\resizebox{\textwidth}{!}{%
\begin{tabular}{ccccccccccccccc}
\hline
Model             & Task                                                                                                              & Validity & \begin{tabular}[c]{@{}c@{}}\(\text{Acc}_{\text{all}}\)\\ (0.65)\end{tabular} & \begin{tabular}[c]{@{}c@{}}\(\text{Acc}_{\text{valid}}\)\\ (0.65)\end{tabular} & \begin{tabular}[c]{@{}c@{}}\(\text{Acc}_{\text{all}}\)\\ (0.15)\end{tabular} & \begin{tabular}[c]{@{}c@{}}\(\text{Acc}_{\text{valid}}\)\\ (0.15)\end{tabular} & FCD              & Task                                                                                                              & Validity & \begin{tabular}[c]{@{}c@{}}\(\text{Acc}_{\text{all}}\)\\ (0.65)\end{tabular} & \begin{tabular}[c]{@{}c@{}}\(\text{Acc}_{\text{valid}}\)\\ (0.65)\end{tabular} & \begin{tabular}[c]{@{}c@{}}\(\text{Acc}_{\text{all}}\)\\ (0.15)\end{tabular} & \begin{tabular}[c]{@{}c@{}}\(\text{Acc}_{\text{valid}}\)\\ (0.15)\end{tabular} & FCD              \\ \hline
BioT5          & \multirow{5}{*}{\textbf{\begin{tabular}[c]{@{}c@{}}HACCEPT${\downarrow}$\\ HDONORS${\downarrow}$\end{tabular}}}                   & 1.0               & 0.0                          & 0.0                          & 0.352                        & 0.352                        & 17.731           & \multirow{5}{*}{\textbf{\begin{tabular}[c]{@{}c@{}}JNK3${\downarrow}$\\ QED${\uparrow}$\end{tabular}}}                          & 1.0               & 0.0                          & 0.0                          & 0.19                         & 0.19                         & 19.8292          \\
DrugAssist     &                                                                                                               & 0.9819            & 0.2711                       & 0.2761                       & 0.3574                       & 0.364                        & 12.987           &                                                                                                             & 0.98              & 0.292                        & 0.298                        & 0.336                        & 0.3429                       & 11.1755          \\
GeLLM$^3$O\_M  &                                                                                                               & 0.89              & 0.108                        & 0.1213                       & 0.264                        & 0.2966                       & 14.7575          &                                                                                                             & 0.914             & 0.148                        & 0.1619                       & 0.326                        & 0.3567                       & 10.123           \\
GeLLM$^3$O\_L    &                                                                                                               & 0.9               & 0.146                        & 0.1622                       & 0.36                         & 0.4                          & 12.0622          &                                                                                                             & 0.9               & 0.098                        & 0.1089                       & 0.352                        & 0.3911                       & 10.866           \\
MolEditRL            &                                                                                                               & 0.972             & \textbf{0.358}               & \textbf{0.3739}              & \textbf{0.612}               & \textbf{0.6497}              & \textbf{11.7393} &                                                                                                             & 0.976             & \textbf{0.33}                & \textbf{0.3381}              & \textbf{0.416}               & \textbf{0.4262}              & \textbf{9.6139}  \\ \hline
BioT5          & \multirow{5}{*}{\textbf{\begin{tabular}[c]{@{}c@{}}HACCEPT${\uparrow}$\\ SA${\uparrow}$\end{tabular}}}                            & 1.0               & 0.0                          & 0.0                          & 0.098                        & 0.098                        & 24.7313          & \multirow{5}{*}{\textbf{\begin{tabular}[c]{@{}c@{}}DRD2${\downarrow}$\\ GSK3B${\uparrow}$\end{tabular}}}                        & 1.0               & 0.0                          & 0.0                          & 0.088                        & 0.088                        & 24.19            \\
DrugAssist     &                                                                                                               & 0.954             & 0.226                        & 0.2369                       & 0.284                        & 0.2977                       & 11.5424          &                                                                                                             & 0.992             & 0.104                        & 0.1048                       & 0.126                        & 0.127                        & 12.3998          \\
GeLLM$^3$O\_M  &                                                                                                               & 0.918             & 0.012                        & 0.0131                       & 0.07                         & 0.0763                       & 23.0712          &                                                                                                             & 0.942             & 0.036                        & 0.0382                       & 0.064                        & 0.0679                       & 15.6116          \\
GeLLM$^3$O\_L    &                                                                                                               & 0.904             & 0.026                        & 0.0288                       & 0.048                        & 0.0531                       & 14.8785          &                                                                                                             & 0.9               & 0.05                         & 0.0556                       & 0.098                        & 0.1089                       & 12.7831          \\
MolEditRL            &                                                                                                               & 0.962             & \textbf{0.316}               & \textbf{0.3583}              & \textbf{0.58}                & \textbf{0.6576}              & \textbf{11.2492} &                                                                                                             & 0.97              & \textbf{0.186}               & \textbf{0.1918}              & \textbf{0.228}               & \textbf{0.2351}              & \textbf{11.4433} \\ \hline
BioT5          & \multirow{5}{*}{\textbf{\begin{tabular}[c]{@{}c@{}}LOGP${\downarrow}$\\ ROTBONDS${\downarrow}$\end{tabular}}}                     & 1.0               & 0.0                          & 0.0                          & 0.104                        & 0.104                        & 27.651           & \multirow{5}{*}{\textbf{\begin{tabular}[c]{@{}c@{}}DRD2${\uparrow}$\\ SA${\uparrow}$\end{tabular}}}                             & 1.0               & 0.0                          & 0.0                          & 0.25                         & 0.25                         & 28.7563          \\
DrugAssist     &                                                                                                               & 0.98              & 0.346                        & 0.3531                       & 0.384                        & 0.3918                       & 8.4341           &                                                                                                             & 0.976             & 0.212                        & 0.2172                       & 0.254                        & 0.2602                       & 11.4352          \\
GeLLM$^3$O\_M  &                                                                                                               & 0.892             & 0.032                        & 0.0359                       & 0.128                        & 0.1435                       & 16.9039          &                                                                                                             & 0.898             & 0.102                        & 0.1136                       & 0.232                        & 0.2584                       & 11.5436          \\
GeLLM$^3$O\_L    &                                                                                                               & 0.908             & 0.09                         & 0.0991                       & 0.3                          & 0.3304                       & 10.9156          &                                                                                                             & 0.924             & 0.07                         & 0.0758                       & 0.222                        & 0.2403                       & 12.0196          \\
MolEditRL            &                                                                                                               & 0.97              & \textbf{0.454}               & \textbf{0.468}               & \textbf{0.686}               & \textbf{0.7072}              & \textbf{6.2095}  &                                                                                                             & 0.912             & \textbf{0.23}                & \textbf{0.2522}              & \textbf{0.398}               & \textbf{0.4364}              & \textbf{10.8934} \\ \hline
BioT5          & \multirow{5}{*}{\textbf{\begin{tabular}[c]{@{}c@{}}LOGP${\downarrow}$\\ ROTBONDS${\uparrow}$\end{tabular}}}                       & 1.0               & 0.0                          & 0.0                          & 0.072                        & 0.072                        & 31.688           & \multirow{5}{*}{\textbf{\begin{tabular}[c]{@{}c@{}}QED${\downarrow}$\\ ROTBONDS${\uparrow}$\end{tabular}}}                      & 1.0               & 0.0                          & 0.0                          & 0.216                        & 0.216                        & 19.1627          \\
DrugAssist     &                                                                                                               & 0.96              & 0.09                         & 0.0938                       & 0.1                          & 0.1042                       & 19.404           &                                                                                                             & 0.984             & 0.24                         & 0.2439                       & 0.276                        & 0.2805                       & 11.0221          \\
GeLLM$^3$O\_M  &                                                                                                               & 0.86              & 0.014                        & 0.0163                       & 0.034                        & 0.0395                       & 30.3704          &                                                                                                             & 0.878             & 0.014                        & 0.0159                       & 0.096                        & 0.1093                       & 21.1825          \\
GeLLM$^3$O\_L    &                                                                                                               & 0.906             & 0.022                        & 0.0243                       & 0.06                         & 0.0662                       & 16.6925          &                                                                                                             & 0.902             & 0.064                        & 0.071                        & 0.166                        & 0.184                        & 11.7206          \\
MolEditRL            &                                                                                                               & 0.954             & \textbf{0.344}               & \textbf{0.3891}              & \textbf{0.634}               & \textbf{0.7172}              & \textbf{12.0673} &                                                                                                             & 0.943             & \textbf{0.422}               & \textbf{0.4742}              & \textbf{0.83}                & \textbf{0.9326}              & \textbf{7.564}   \\ \hline
BioT5          & \multirow{5}{*}{\textbf{\begin{tabular}[c]{@{}c@{}}MW${\uparrow}$\\ QED${\downarrow}$\end{tabular}}}                              & 1.0               & 0.0                          & 0.0                          & 0.27                         & 0.27                         & 17.3349          & \multirow{5}{*}{\textbf{\begin{tabular}[c]{@{}c@{}}QED${\downarrow}$\\ SA${\uparrow}$\end{tabular}}}                            & 1.0               & 0.0                          & 0.0                          & 0.196                        & 0.196                        & 20.749           \\
DrugAssist     &                                                                                                               & 0.98              & 0.298                        & 0.3041                       & 0.354                        & 0.3612                       & 9.5465           &                                                                                                             & 0.978             & 0.2325                       & 0.2377                       & 0.2766                       & 0.2828                       & 11.0132          \\
GeLLM$^3$O\_M  &                                                                                                               & 0.926             & 0.072                        & 0.0778                       & 0.238                        & 0.257                        & 12.8711          &                                                                                                             & 0.906             & 0.086                        & 0.0949                       & 0.184                        & 0.2031                       & 10.3846          \\
GeLLM$^3$O\_L    &                                                                                                               & 0.882             & 0.158                        & 0.1791                       & 0.316                        & 0.3583                       & 7.8458           &                                                                                                             & 0.894             & 0.078                        & 0.0872                       & 0.172                        & 0.1924                       & 10.3566          \\
MolEditRL            &                                                                                                               & 0.944             & \textbf{0.35}                & \textbf{0.4147}              & \textbf{0.79}                & \textbf{0.936}               & \textbf{7.0482}  &                                                                                                             & 0.938             & \textbf{0.592}               & \textbf{0.6311}              & \textbf{0.878}               & \textbf{0.936}               & \textbf{7.3882}  \\ \hline
BioT5          & \multirow{5}{*}{\textbf{\begin{tabular}[c]{@{}c@{}}DRD2${\downarrow}$\\ HACCEPT${\uparrow}$\\ MW${\downarrow}$\end{tabular}}}               & 1.0               & 0.0                          & 0.0                          & 0.016                        & 0.016                        & 49.8934          & \multirow{5}{*}{\textbf{\begin{tabular}[c]{@{}c@{}}DRD2${\uparrow}$\\ HACCEPT${\uparrow}$\\ SA${\uparrow}$\end{tabular}}}                 & 1.0               & 0.0                          & 0.0                          & 0.07                         & 0.07                         & 36.9063          \\
DrugAssist     &                                                                                                               & 0.95              & 0.062                        & 0.0653                       & 0.082                        & 0.0863                       & 14.6988          &                                                                                                             & 0.956             & 0.142                        & 0.1485                       & 0.192                        & 0.2008                       & 13.9149          \\
GeLLM$^3$O\_M  &                                                                                                               & 0.91              & 0.018                        & 0.0198                       & 0.034                        & 0.0374                       & 16.9543          &                                                                                                             & 0.904             & 0.004                        & 0.0044                       & 0.064                        & 0.0708                       & 28.5396          \\
GeLLM$^3$O\_L    &                                                                                                               & 0.916             & 0.01                         & 0.0109                       & 0.02                         & 0.0218                       & 24.1798          &                                                                                                             & 0.92              & 0.016                        & 0.0174                       & 0.05                         & 0.0543                       & 18.2579          \\
MolEditRL            &                                                                                                               & 0.962             & \textbf{0.1}                 & \textbf{0.104}               & \textbf{0.264}               & \textbf{0.2744}              & \textbf{12.5672} &                                                                                                             & 0.966             & \textbf{0.192}               & \textbf{0.1988}              & \textbf{0.288}               & \textbf{0.2981}              & \textbf{13.7681} \\ \hline
BioT5          & \multirow{5}{*}{\textbf{\begin{tabular}[c]{@{}c@{}}DRD2${\uparrow}$\\ HACCEPT${\uparrow}$\\ JNK3${\uparrow}$\end{tabular}}}                 & 1.0               & 0.0                          & 0.0                          & 0.086                        & 0.086                        & 28.2977          & \multirow{5}{*}{\textbf{\begin{tabular}[c]{@{}c@{}}DRD2${\uparrow}$\\ JNK3${\uparrow}$\\ QED${\downarrow}$\end{tabular}}}                 & 1.0               & 0.0                          & 0.0                          & 0.116                        & 0.116                        & 23.0032          \\
DrugAssist     &                                                                                                               & 0.972             & 0.09                         & 0.0926                       & 0.14                         & 0.144                        & 15.8384          &                                                                                                             & 0.99              & 0.114                        & 0.1152                       & 0.158                        & 0.1596                       & 13.438           \\
GeLLM$^3$O\_M  &                                                                                                               & 0.904             & 0.03                         & 0.0332                       & 0.082                        & 0.0907                       & 23.1175          &                                                                                                             & 0.91              & 0.062                        & 0.0681                       & 0.13                         & 0.1429                       & 15.6618          \\
GeLLM$^3$O\_L    &                                                                                                               & 0.91              & 0.026                        & 0.0286                       & 0.054                        & 0.0593                       & 18.5348          &                                                                                                             & 0.912             & 0.034                        & 0.0373                       & 0.082                        & 0.0899                       & 18.1148          \\
MolEditRL            &                                                                                                               & 0.94              & \textbf{0.22}                & \textbf{0.234}               & \textbf{0.294}               & \textbf{0.3128}              & \textbf{13.1873} &                                                                                                             & 0.938             & \textbf{0.258}               & \textbf{0.2751}              & \textbf{0.438}               & \textbf{0.467}               & \textbf{9.2013}  \\ \hline
BioT5          & \multirow{5}{*}{\textbf{\begin{tabular}[c]{@{}c@{}}GSK3B${\uparrow}$\\ HDONORS${\uparrow}$\\ QED${\downarrow}$\\ SA${\uparrow}$\end{tabular}}}        & 1.0               & 0.0                          & 0.0                          & 0.066                        & 0.066                        & 32.3437          & \multirow{5}{*}{\textbf{\begin{tabular}[c]{@{}c@{}}DRD2${\downarrow}$\\ GSK3B${\uparrow}$\\ HDONORS${\uparrow}$\\ LOGP${\downarrow}$\end{tabular}}} & 1.0               & 0.0                          & 0.0                          & 0.024                        & 0.024                        & 43.4139          \\
DrugAssist     &                                                                                                               & 0.948             & 0.056                        & 0.0591                       & 0.062                        & 0.0654                       & 21.3896          &                                                                                                             & 0.954             & 0.02                         & 0.021                        & 0.028                        & 0.0294                       & 19.8179          \\
GeLLM$^3$O\_M  &                                                                                                               & 0.902             & 0.002                        & 0.0022                       & 0.004                        & 0.0044                       & 24.7065          &                                                                                                             & 0.884             & 0.0                          & 0.0                          & 0.006                        & 0.0068                       & 60.738           \\
GeLLM$^3$O\_L    &                                                                                                               & 0.914             & 0.012                        & 0.0131                       & 0.036                        & 0.0394                       & 18.4036          &                                                                                                             & 0.898             & 0.002                        & 0.0022                       & 0.008                        & 0.0089                       & 19.9507          \\
MolEditRL            &                                                                                                               & 0.954             & \textbf{0.206}               & \textbf{0.2159}              & \textbf{0.416}               & \textbf{0.4361}              & \textbf{14.599}  &                                                                                                             & 0.962             & \textbf{0.174}               & \textbf{0.1809}              & \textbf{0.232}               & \textbf{0.2412}              & \textbf{11.4978} \\ \hline
BioT5          & \multirow{5}{*}{\textbf{\begin{tabular}[c]{@{}c@{}}DRD2${\downarrow}$\\ GSK3B${\downarrow}$\\ HACCEPT${\downarrow}$\\ SA${\downarrow}$\end{tabular}}} & 1.0               & 0.0                          & 0.0                          & 0.088                        & 0.088                        & 30.4482          & \multirow{5}{*}{\textbf{\begin{tabular}[c]{@{}c@{}}GSK3B${\downarrow}$\\ HDONORS${\downarrow}$\\ LOGP${\uparrow}$\\ MW${\downarrow}$\end{tabular}}} & 1.0               & 0.0                          & 0.0                          & 0.1                          & 0.1                          & 27.6262          \\
DrugAssist     &                                                                                                               & 0.988             & 0.082                        & 0.083                        & 0.092                        & 0.0931                       & 21.3253          &                                                                                                             & 0.992             & 0.09                         & 0.0907                       & 0.098                        & 0.0988                       & 24.7748          \\
GeLLM$^3$O\_M  &                                                                                                               & 0.91              & 0.042                        & 0.0462                       & 0.094                        & 0.1033                       & 18.8187          &                                                                                                             & 0.906             & 0.05                         & 0.0552                       & 0.108                        & 0.1192                       & 18.0398          \\
GeLLM$^3$O\_L    &                                                                                                               & 0.918             & 0.042                        & 0.0458                       & 0.12                         & 0.1307                       & 20.2584          &                                                                                                             & 0.91              & 0.034                        & 0.0374                       & 0.11                         & 0.1209                       & 20.1217          \\
MolEditRL            &                                                                                                               & 0.986             & \textbf{0.122}               & \textbf{0.1237}              & \textbf{0.21}                & \textbf{0.213}               & \textbf{14.735}  &                                                                                                             & 0.966             & \textbf{0.146}               & \textbf{0.1511}              & \textbf{0.212}               & \textbf{0.2195}              & \textbf{18.0029} \\ \hline
\end{tabular}}
\end{table}

\section{More Visualization of Molecular Editing}~\label{sec:visualization}

\begin{figure}[htbp]
    \centering
    \begin{subfigure}[b]{0.13\textwidth}
     \centering
     \includegraphics[width=\linewidth,height=5cm,keepaspectratio]{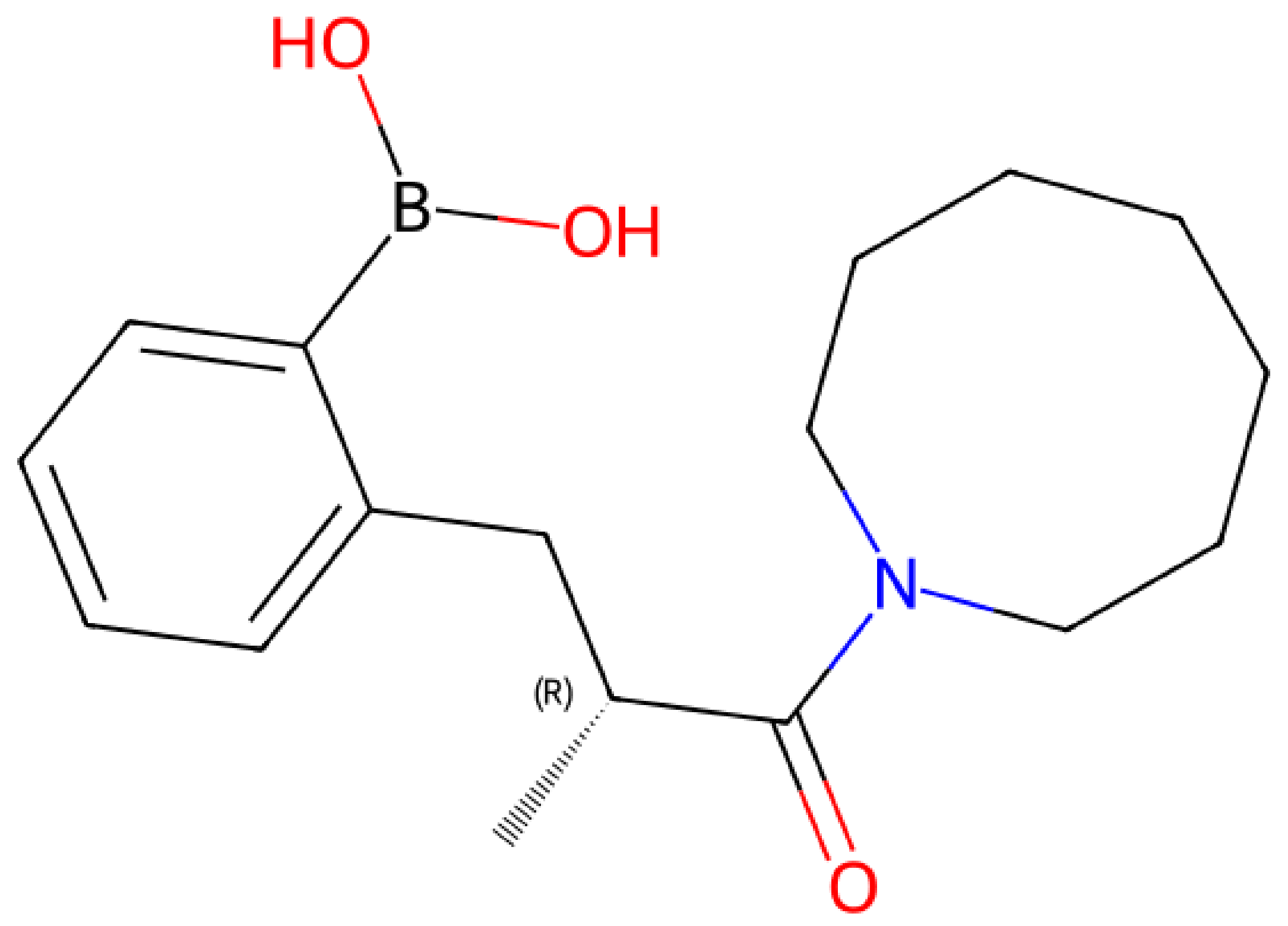}
     \caption{Source}
    \end{subfigure}%
    \hfill
    \begin{subfigure}[b]{0.35\textwidth}
     \centering
     \includegraphics[width=\linewidth,height=5cm,keepaspectratio]{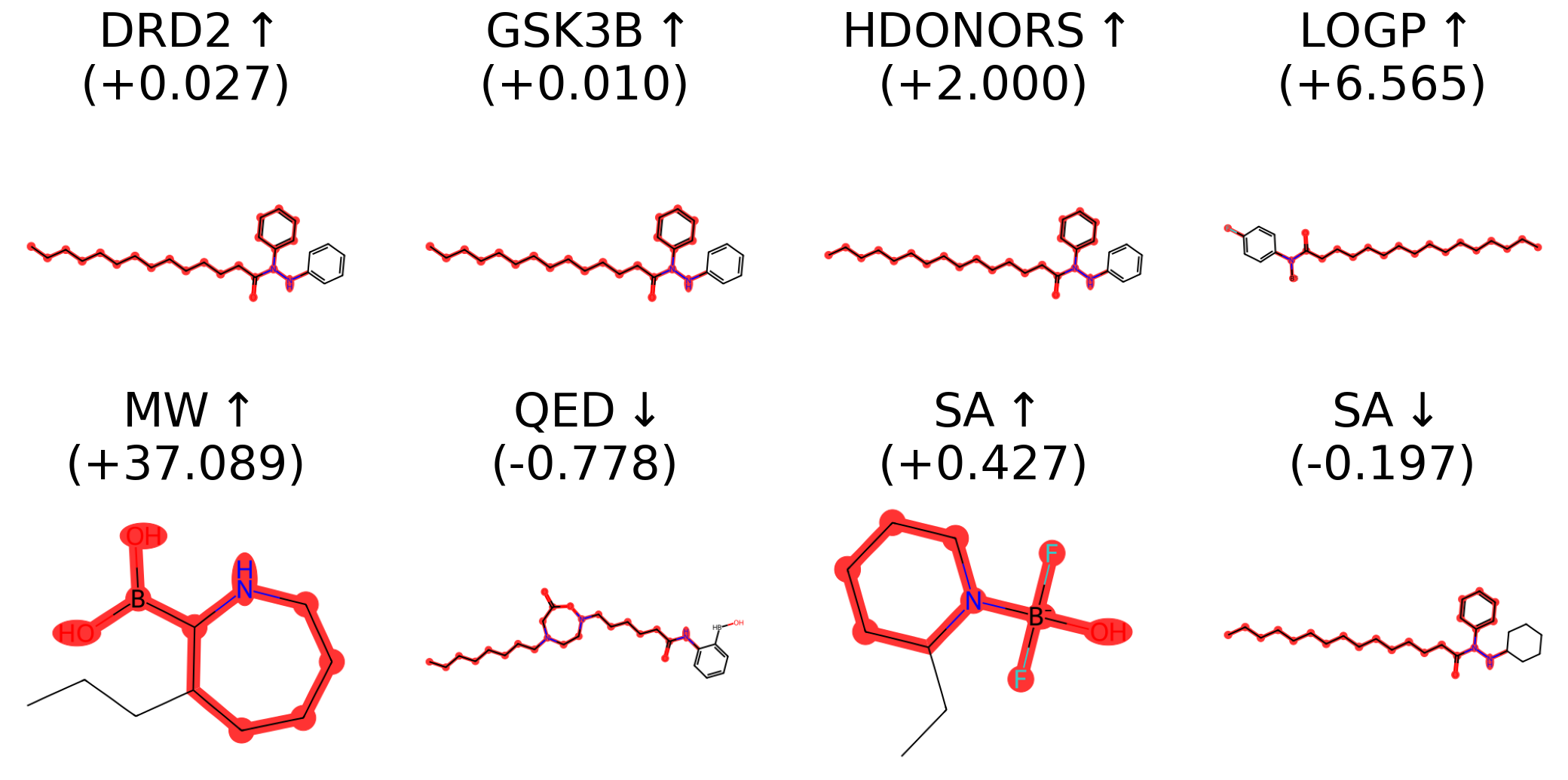}
     \caption{BioT5 (8/20 successful edits)}
    \end{subfigure}%
    \hfill
    \begin{subfigure}[b]{0.47\textwidth}
     \centering
     \includegraphics[width=\linewidth,height=5cm,keepaspectratio]{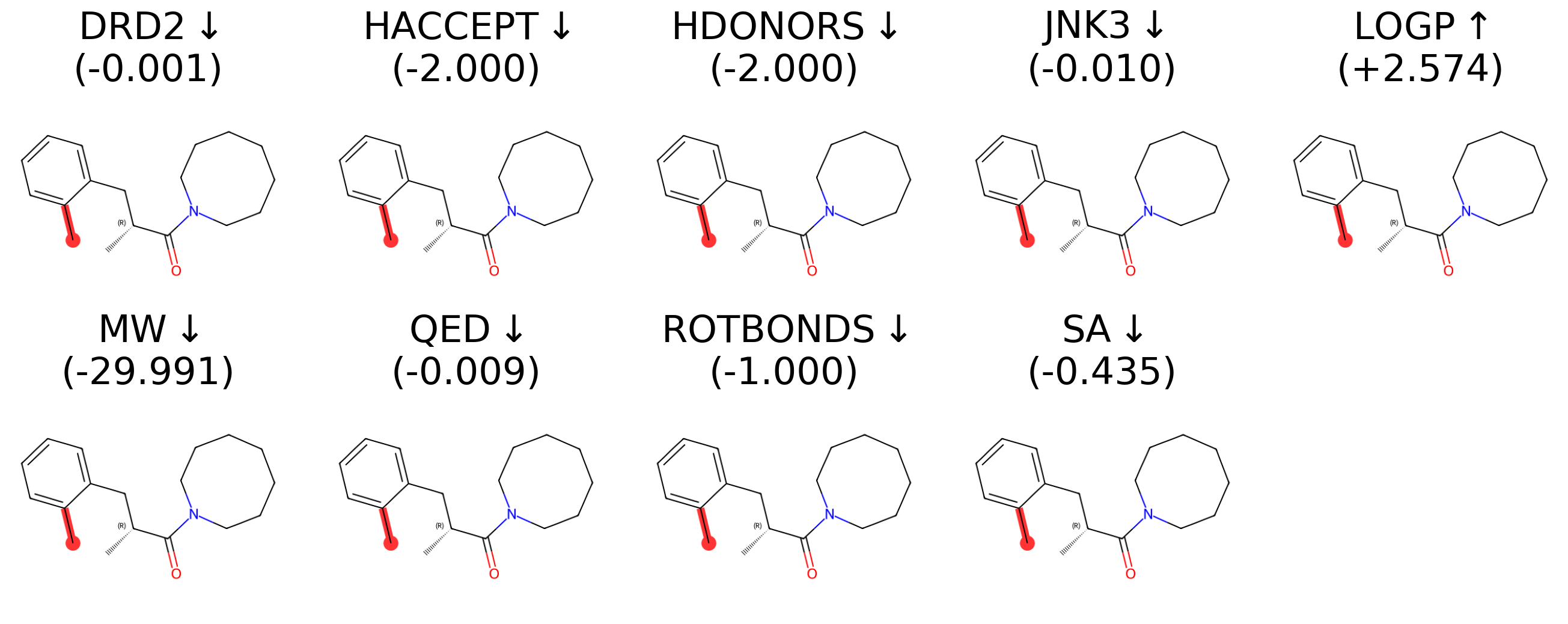}
     \caption{DrugAssist (9/20 successful edits)}
    \end{subfigure}
    
    \vspace{1em}
    
    \begin{subfigure}[b]{0.22\textwidth}
     \centering
     \includegraphics[width=\linewidth,height=5cm,keepaspectratio]{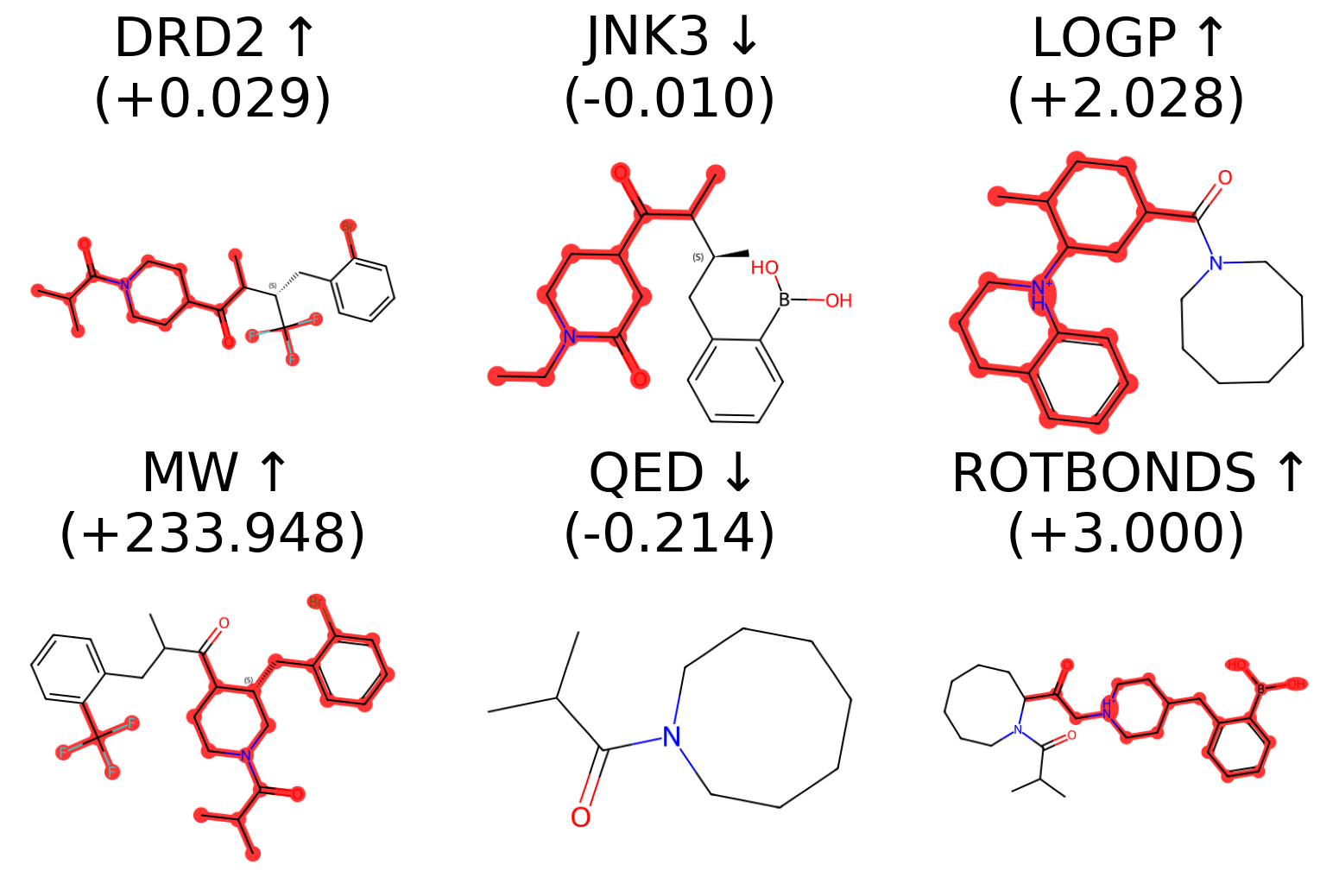}
     \caption{GeLLMO\_L (6/20 successful edits)}
    \end{subfigure}%
    \hfill
    \begin{subfigure}[b]{0.73\textwidth}
     \centering
     \includegraphics[width=\linewidth,height=5cm,keepaspectratio]{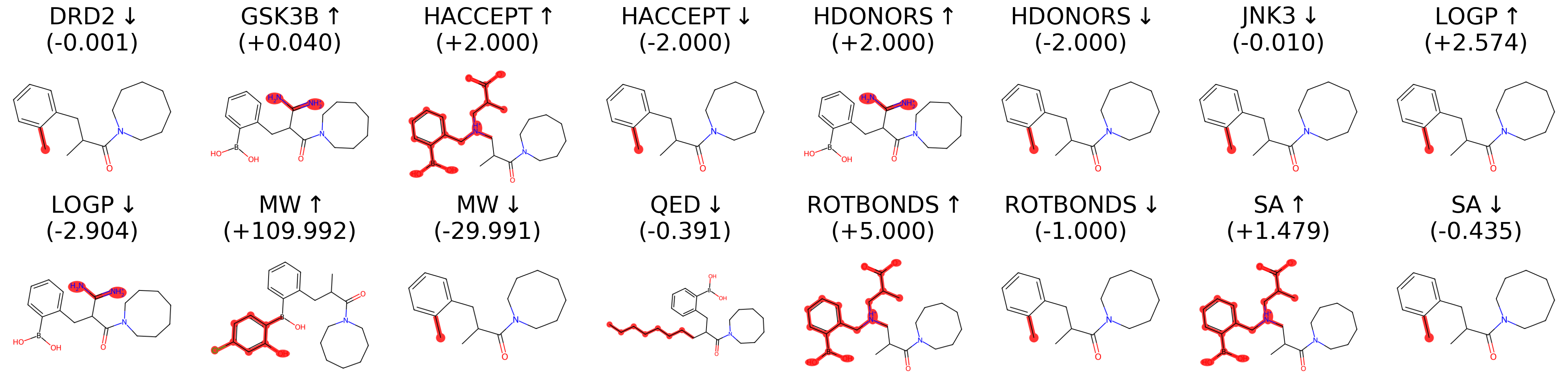}
     \caption{MolEditRL (16/20 successful edits)}
    \end{subfigure}

    \vspace{1em}
    
    \begin{subfigure}[b]{0.13\textwidth}
        \centering
        \includegraphics[width=\linewidth,height=5cm,keepaspectratio]{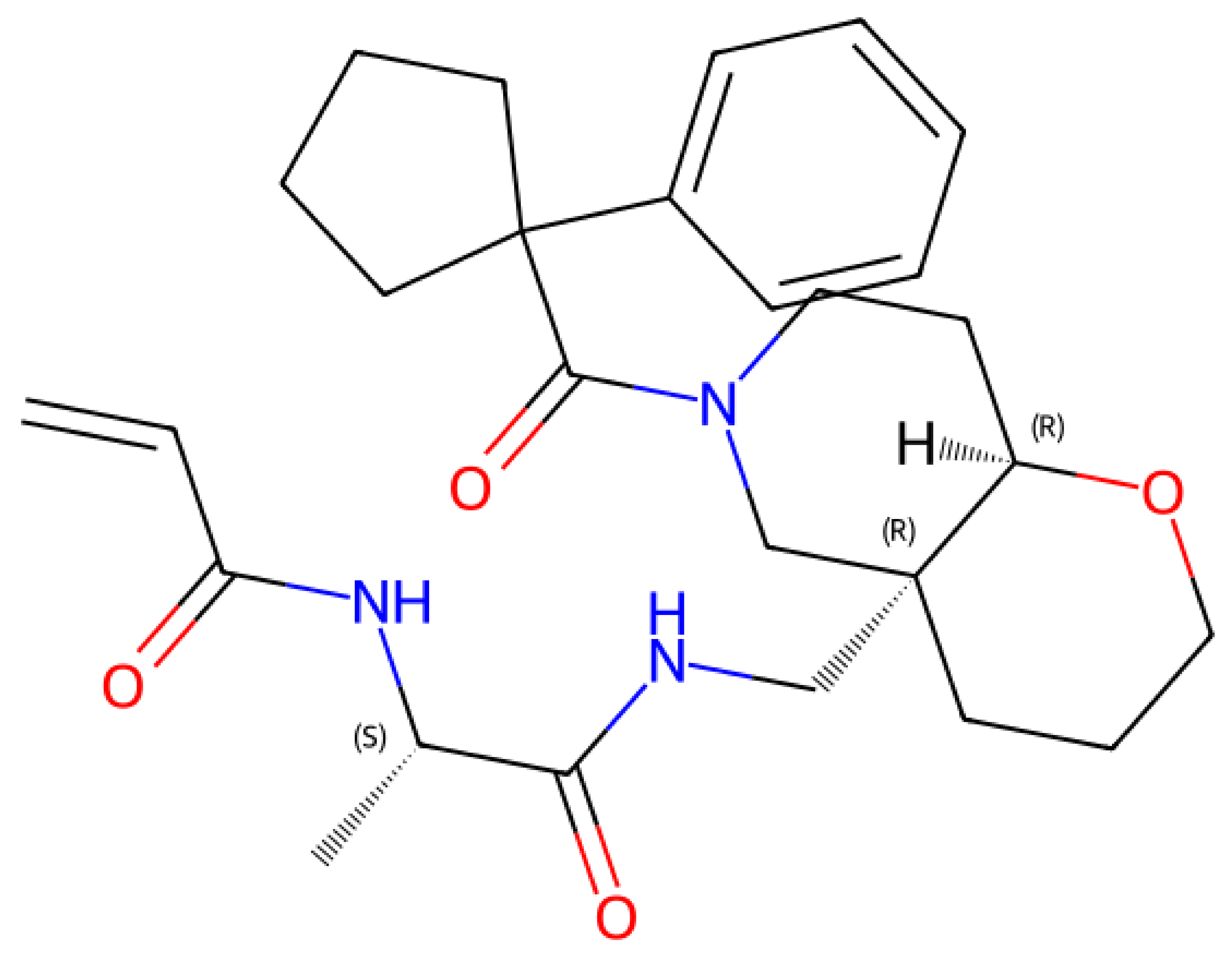}
        \caption{Source}
       \end{subfigure}%
       \hfill
       \begin{subfigure}[b]{0.41\textwidth}
        \centering
        \includegraphics[width=\linewidth,height=5cm,keepaspectratio]{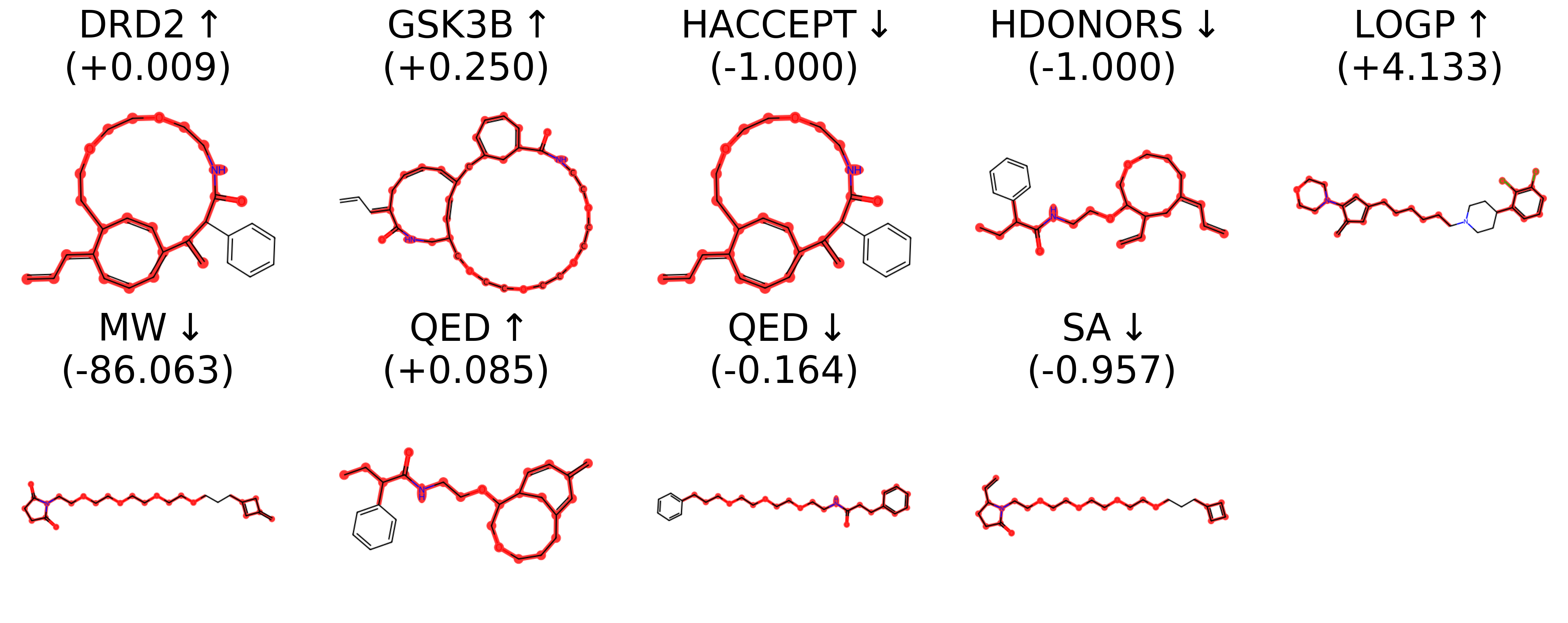}
        \caption{BioT5 (9/20 successful edits)}
       \end{subfigure}%
       \hfill
       \begin{subfigure}[b]{0.42\textwidth}
        \centering
        \includegraphics[width=\linewidth,height=5cm,keepaspectratio]{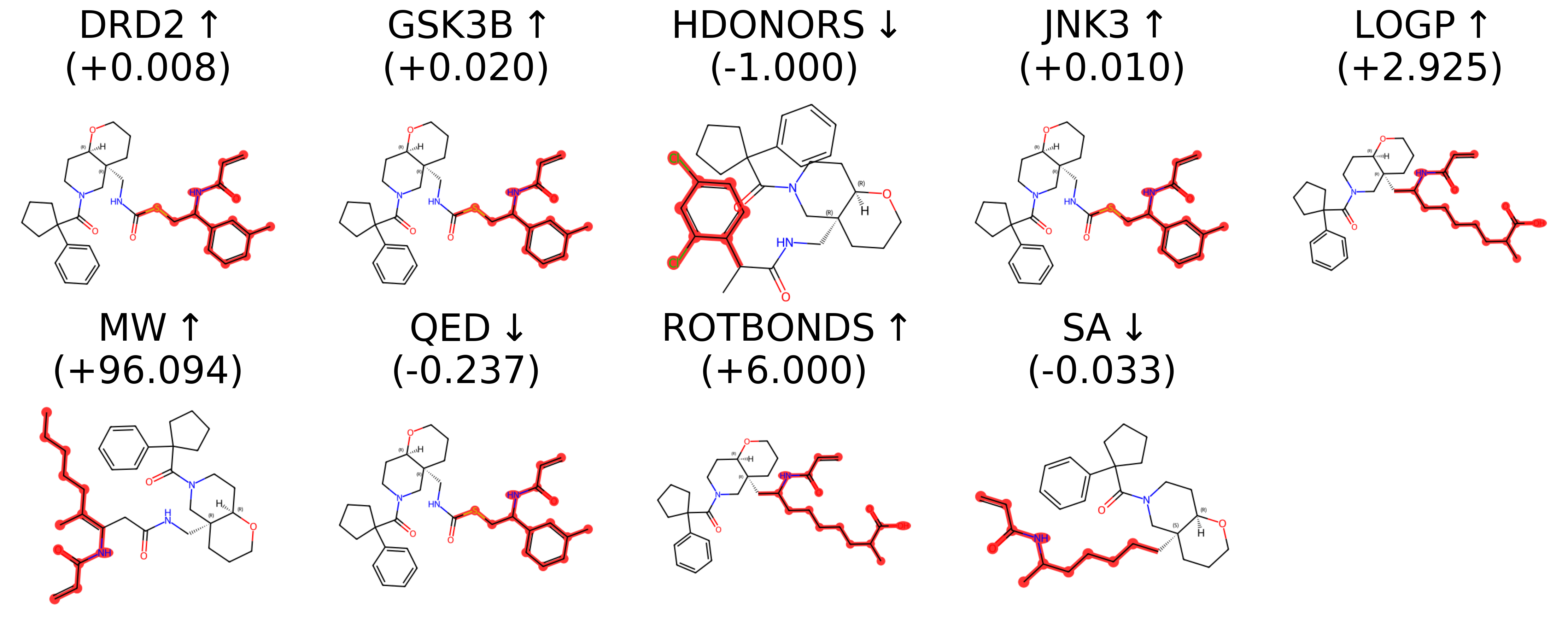}
        \caption{DrugAssist (9/20 successful edits)}
       \end{subfigure}
       
       \vspace{1em}
       
       \begin{subfigure}[b]{0.34\textwidth}
        \centering
        \includegraphics[width=\linewidth,height=5cm,keepaspectratio]{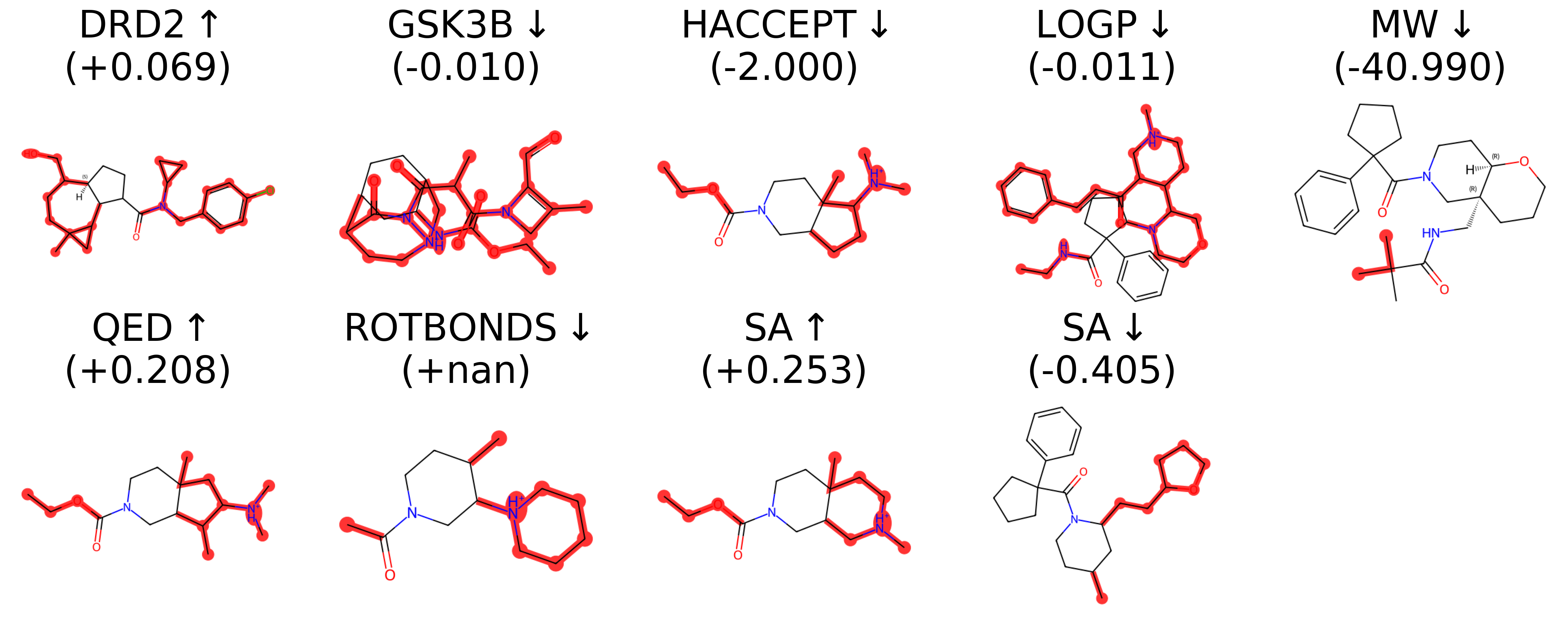}
        \caption{GeLLMO\_L (9/20 successful edits)}
       \end{subfigure}%
       \hfill
       \begin{subfigure}[b]{0.60\textwidth}
        \centering
        \includegraphics[width=\linewidth,height=5cm,keepaspectratio]{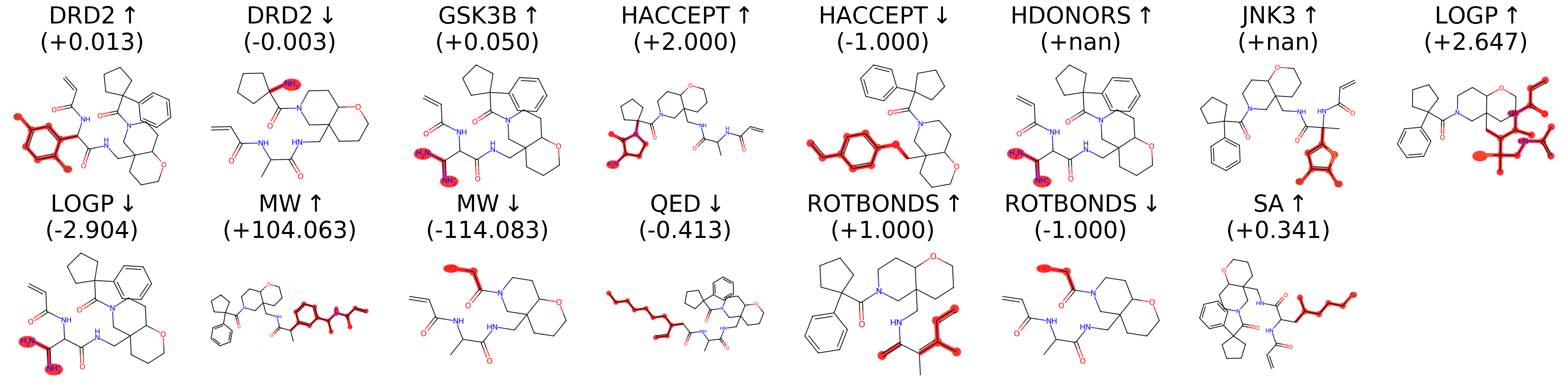}
        \caption{MolEditRL (15/20 successful edits)}
       \end{subfigure}

       \vspace{1em}

       \begin{subfigure}[b]{0.13\textwidth}
        \centering
        \includegraphics[width=\linewidth,height=5cm,keepaspectratio]{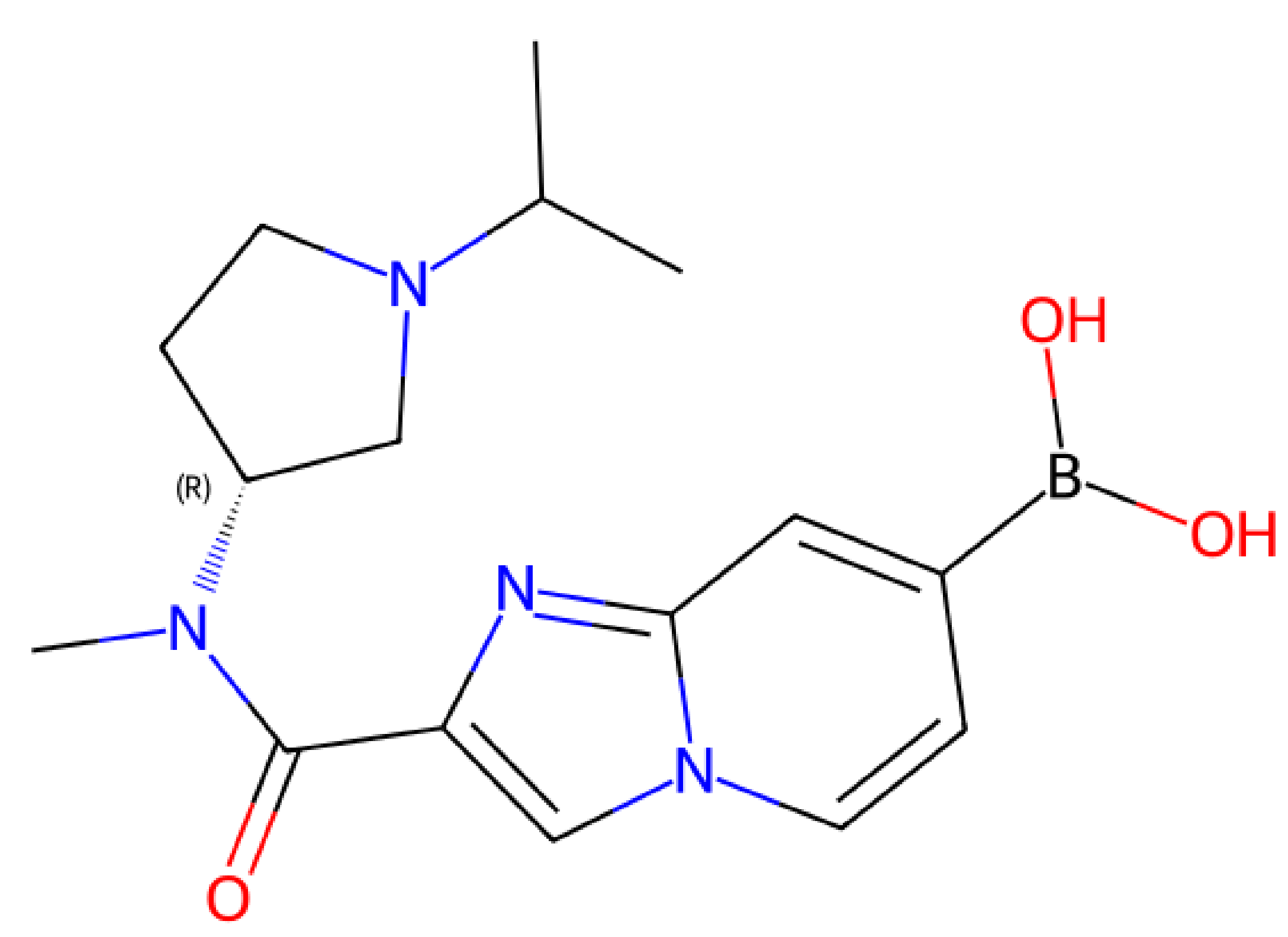}
        \caption{Source}
       \end{subfigure}%
       \hfill
       \begin{subfigure}[b]{0.48\textwidth}
        \centering
        \includegraphics[width=\linewidth,height=5cm,keepaspectratio]{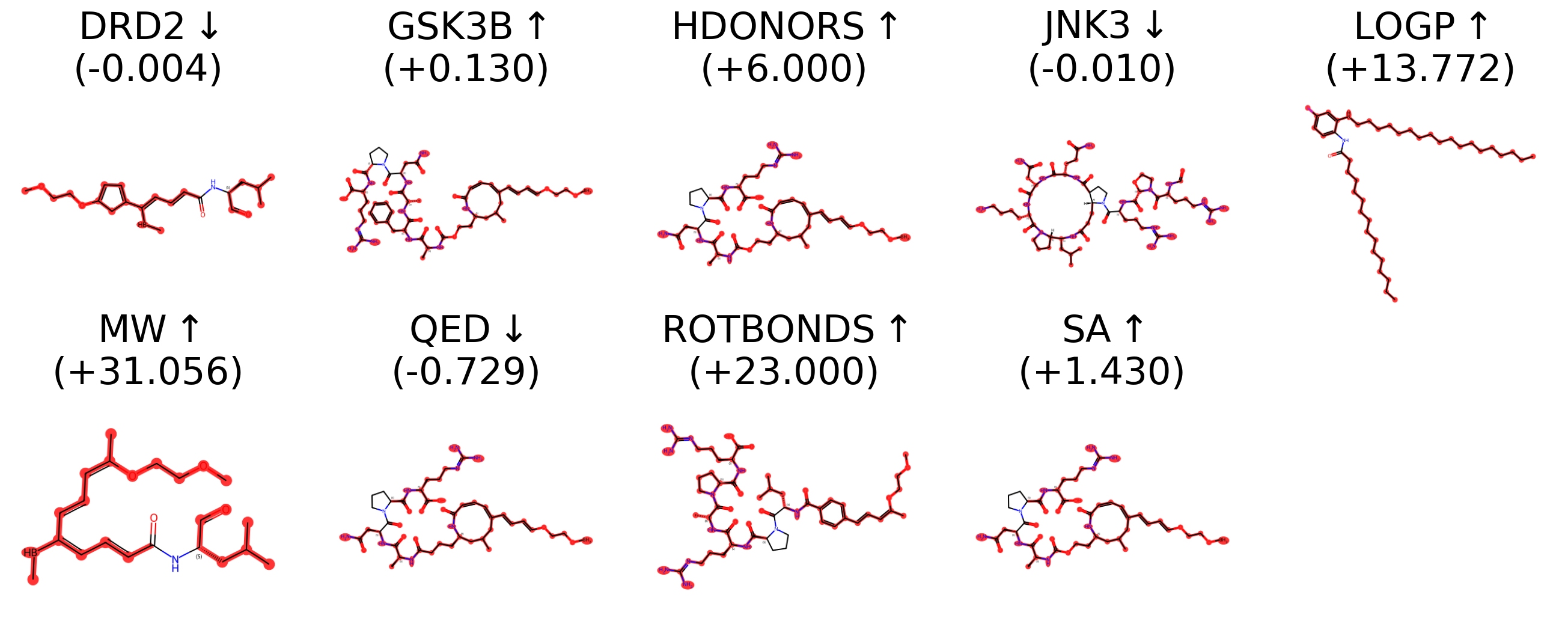}
        \caption{BioT5 (9/20 successful edits)}
       \end{subfigure}%
       \hfill
       \begin{subfigure}[b]{0.37\textwidth}
        \centering
        \includegraphics[width=\linewidth,height=5cm,keepaspectratio]{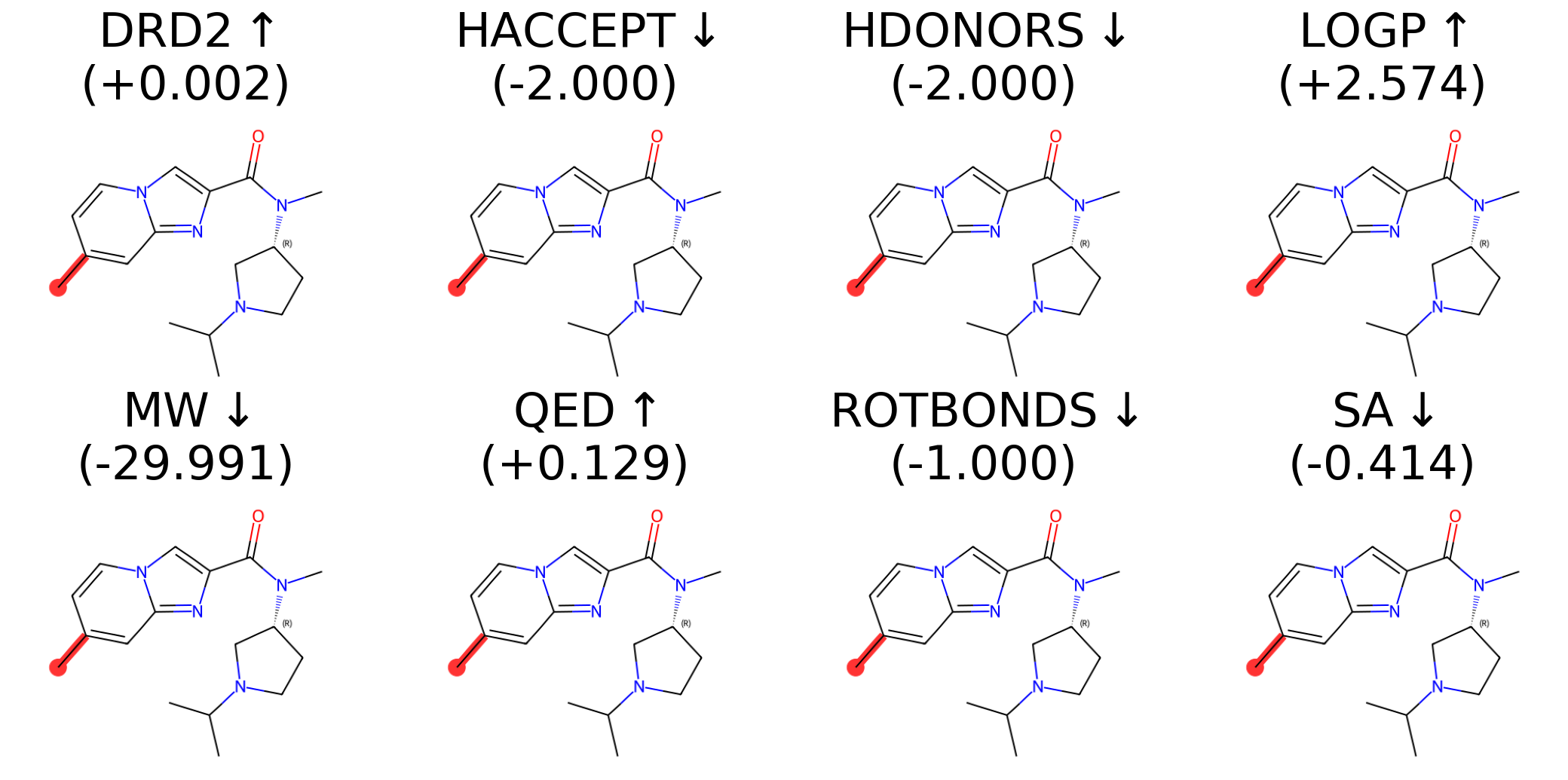}
        \caption{DrugAssist (8/20 successful edits)}
       \end{subfigure}
       
       \vspace{1em}
       
       \begin{subfigure}[b]{0.42\textwidth}
        \centering
        \includegraphics[width=\linewidth,height=5cm,keepaspectratio]{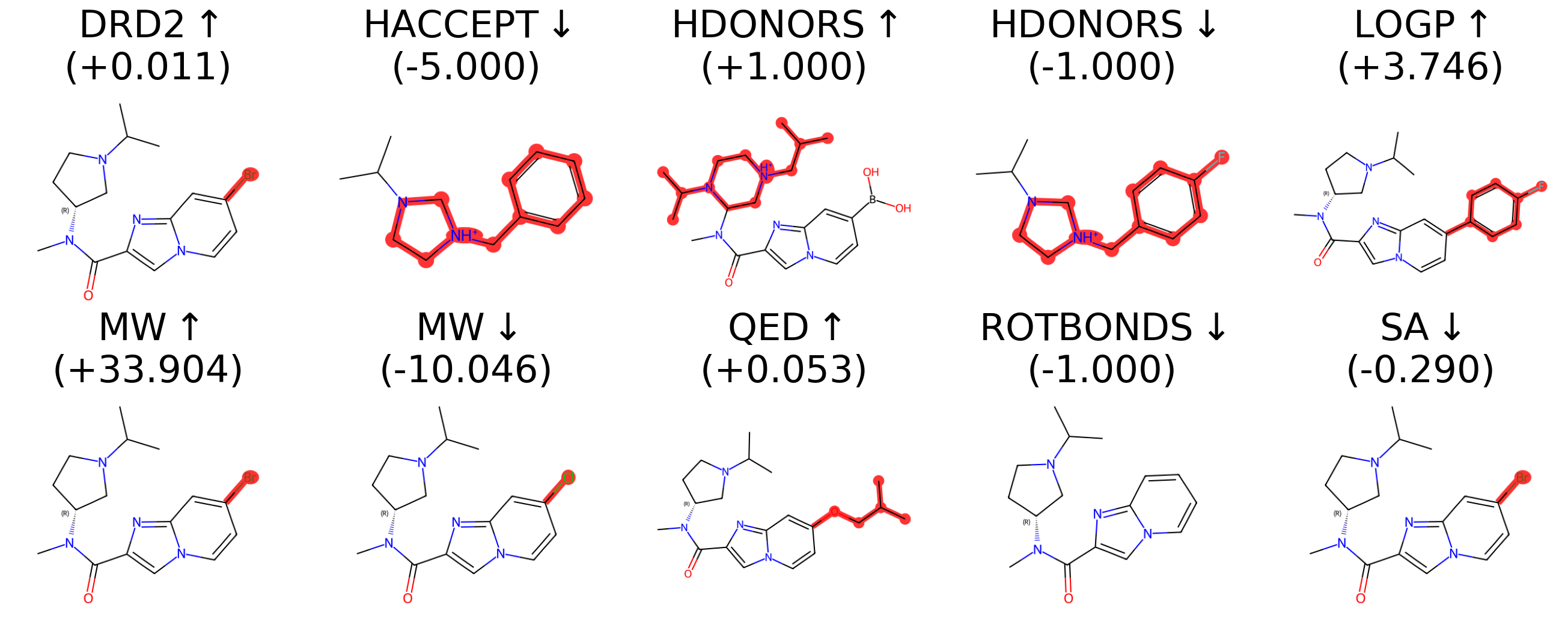}
        \caption{GeLLMO\_L (10/20 successful edits)}
       \end{subfigure}%
       \hfill
       \begin{subfigure}[b]{0.54\textwidth}
        \centering
        \includegraphics[width=\linewidth,height=5cm,keepaspectratio]{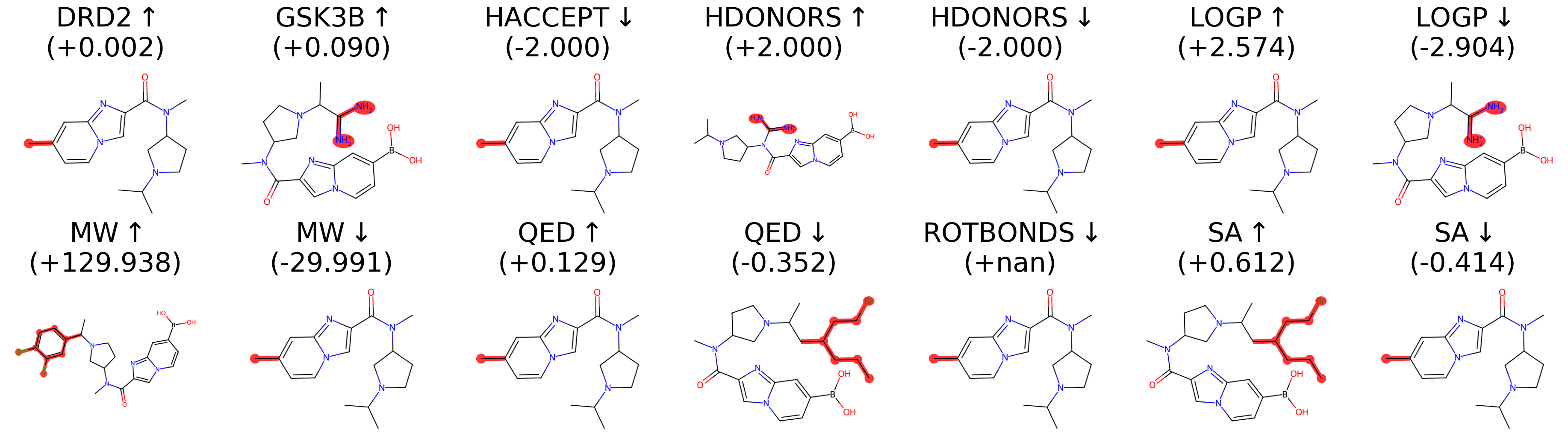}
        \caption{MolEditRL (14/20 successful edits)}
       \end{subfigure}

    \caption{More visualization of edits on 20 tasks.}
    \label{fig:visualize1}
   \end{figure}

\begin{figure}[htbp]
\centering
    \begin{subfigure}[b]{0.13\textwidth}
    \centering
    \includegraphics[width=\linewidth,height=5cm,keepaspectratio]{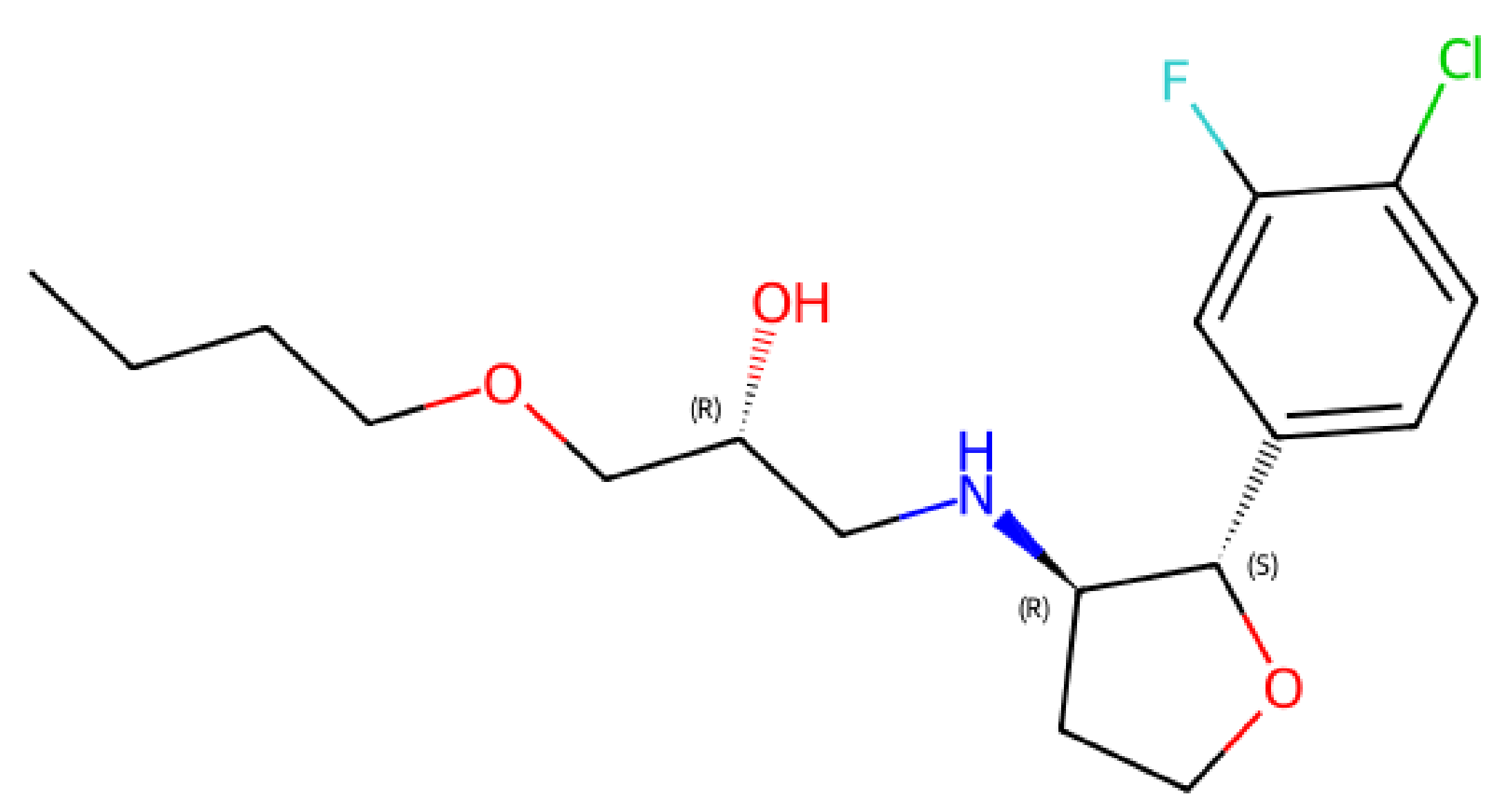}
    \caption{Source}
    \end{subfigure}%
    \hfill
    \begin{subfigure}[b]{0.39\textwidth}
    \centering
    \includegraphics[width=\linewidth,height=5cm,keepaspectratio]{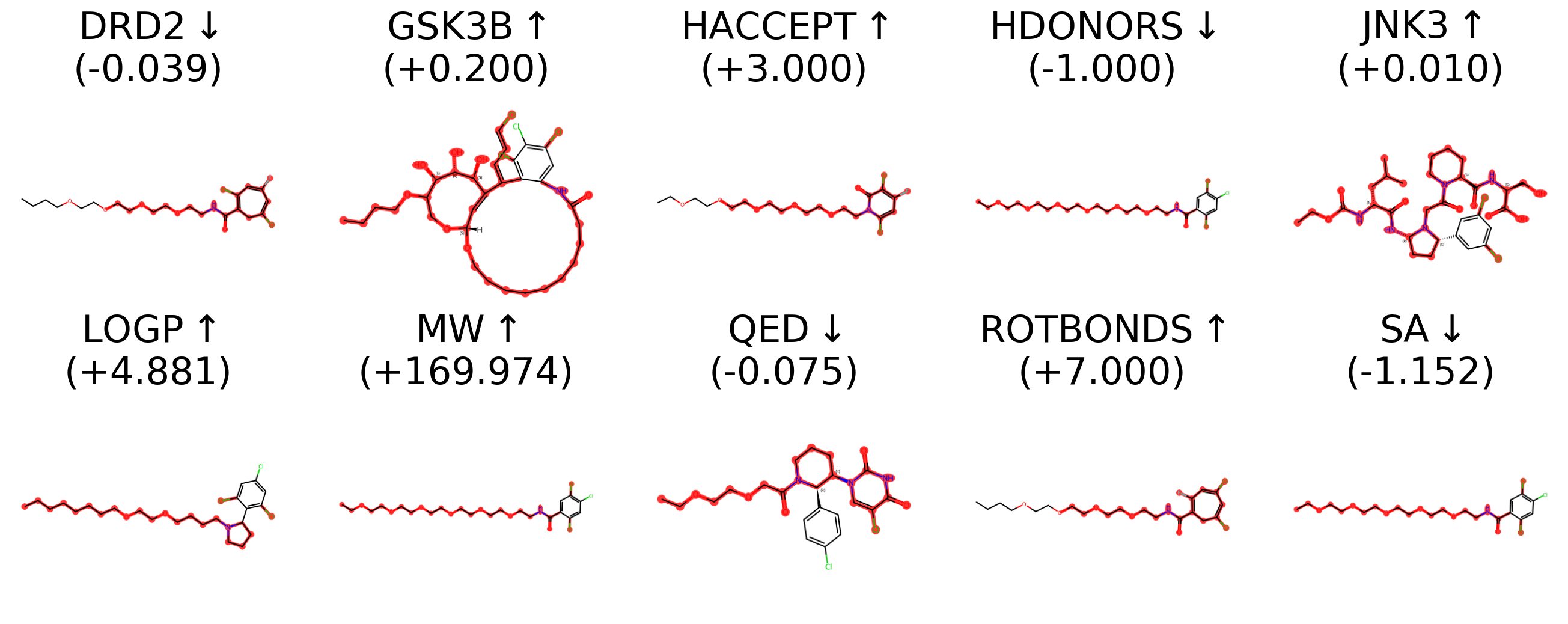}
    \caption{BioT5 (10/20 successful edits)}
    \end{subfigure}%
    \hfill
    \begin{subfigure}[b]{0.44\textwidth}
    \centering
    \includegraphics[width=\linewidth,height=5cm,keepaspectratio]{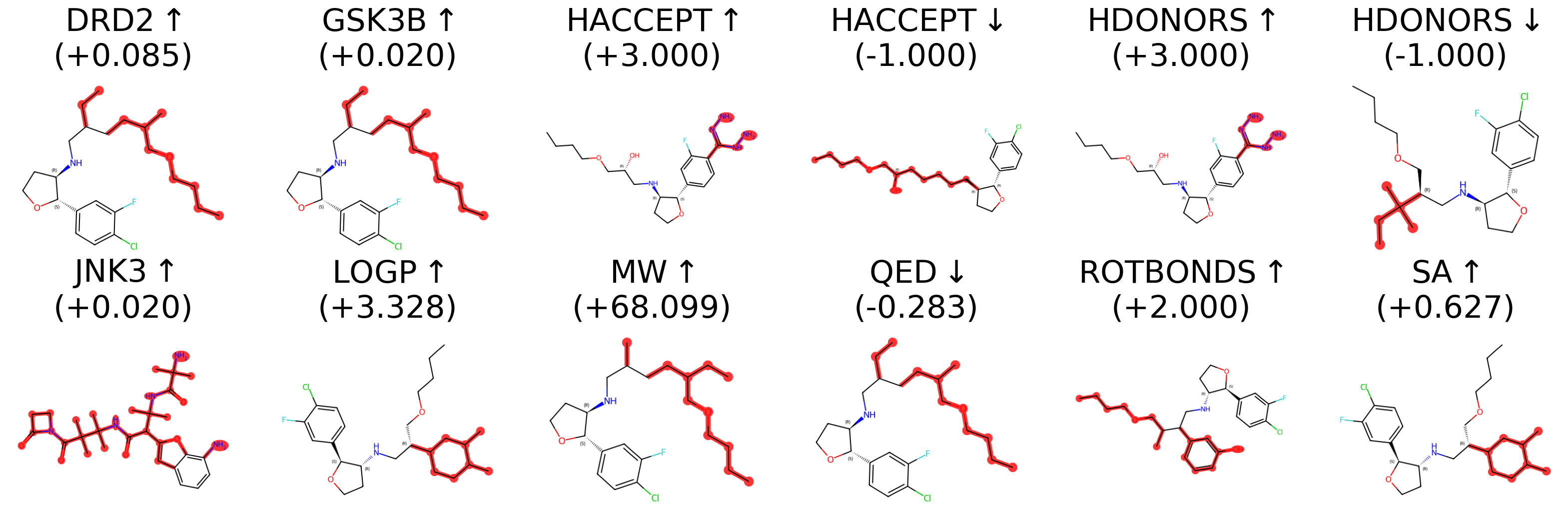}
    \caption{DrugAssist (12/20 successful edits)}
    \end{subfigure}
    
    \vspace{1em}
    
    \begin{subfigure}[b]{0.24\textwidth}
    \centering
    \includegraphics[width=\linewidth,height=5cm,keepaspectratio]{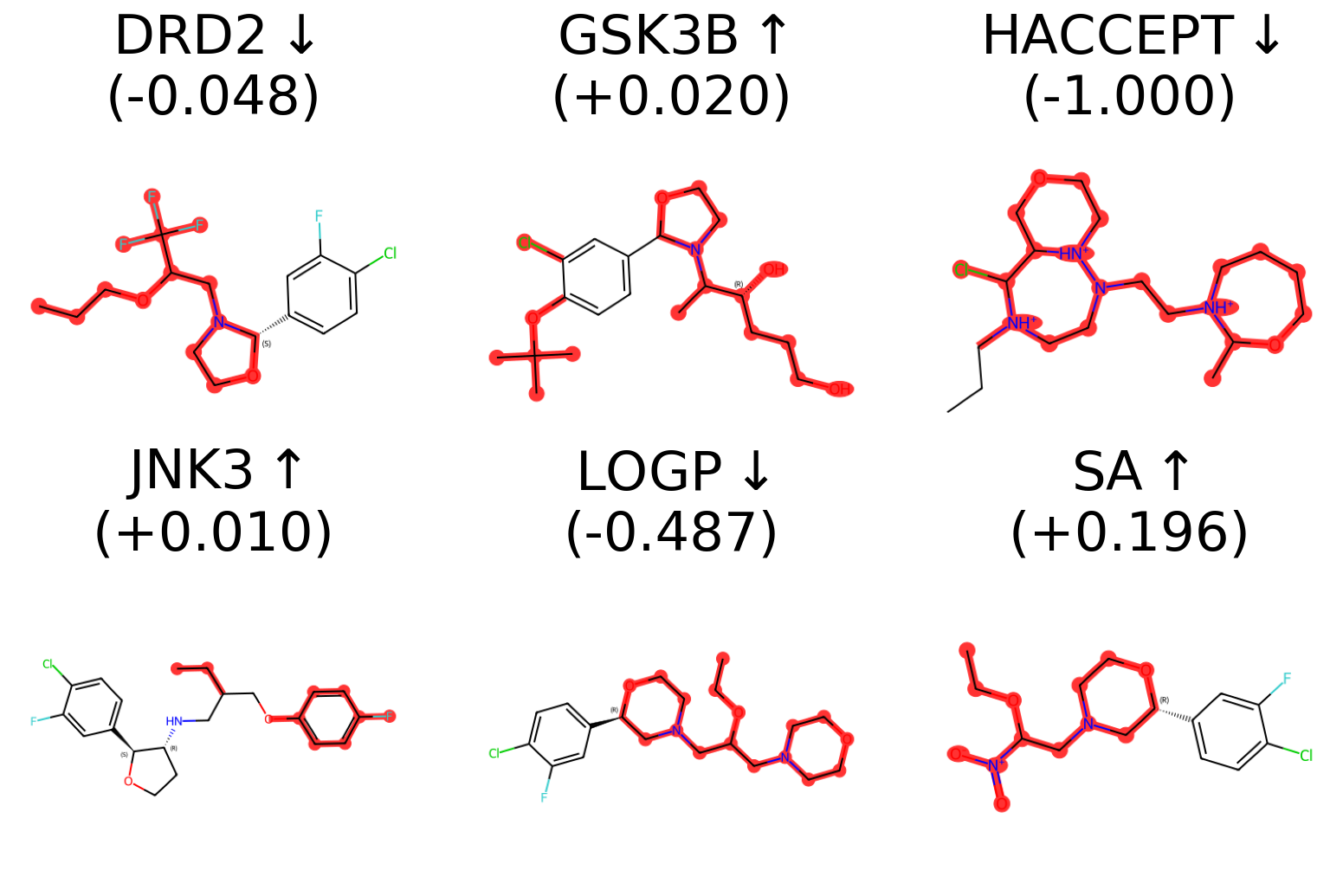}
    \caption{GeLLMO\_L (6/20 successful edits)}
    \end{subfigure}%
    \hfill
    \begin{subfigure}[b]{0.70\textwidth}
    \centering
    \includegraphics[width=\linewidth,height=5cm,keepaspectratio]{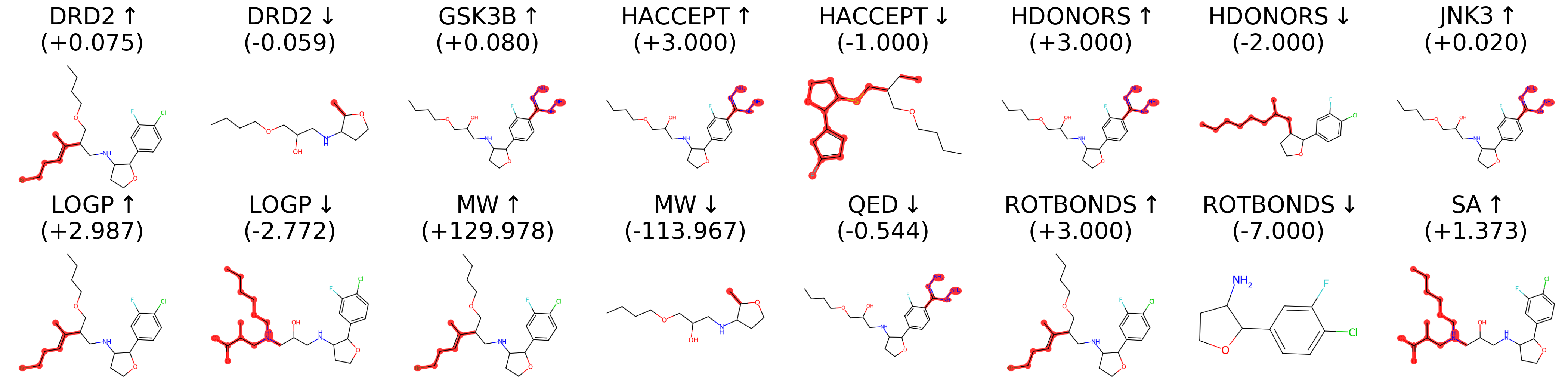}
    \caption{MolEditRL (16/20 successful edits)}
    \end{subfigure}

    \vspace{1em}

    \begin{subfigure}[b]{0.13\textwidth}
    \centering
    \includegraphics[width=\linewidth,height=5cm,keepaspectratio]{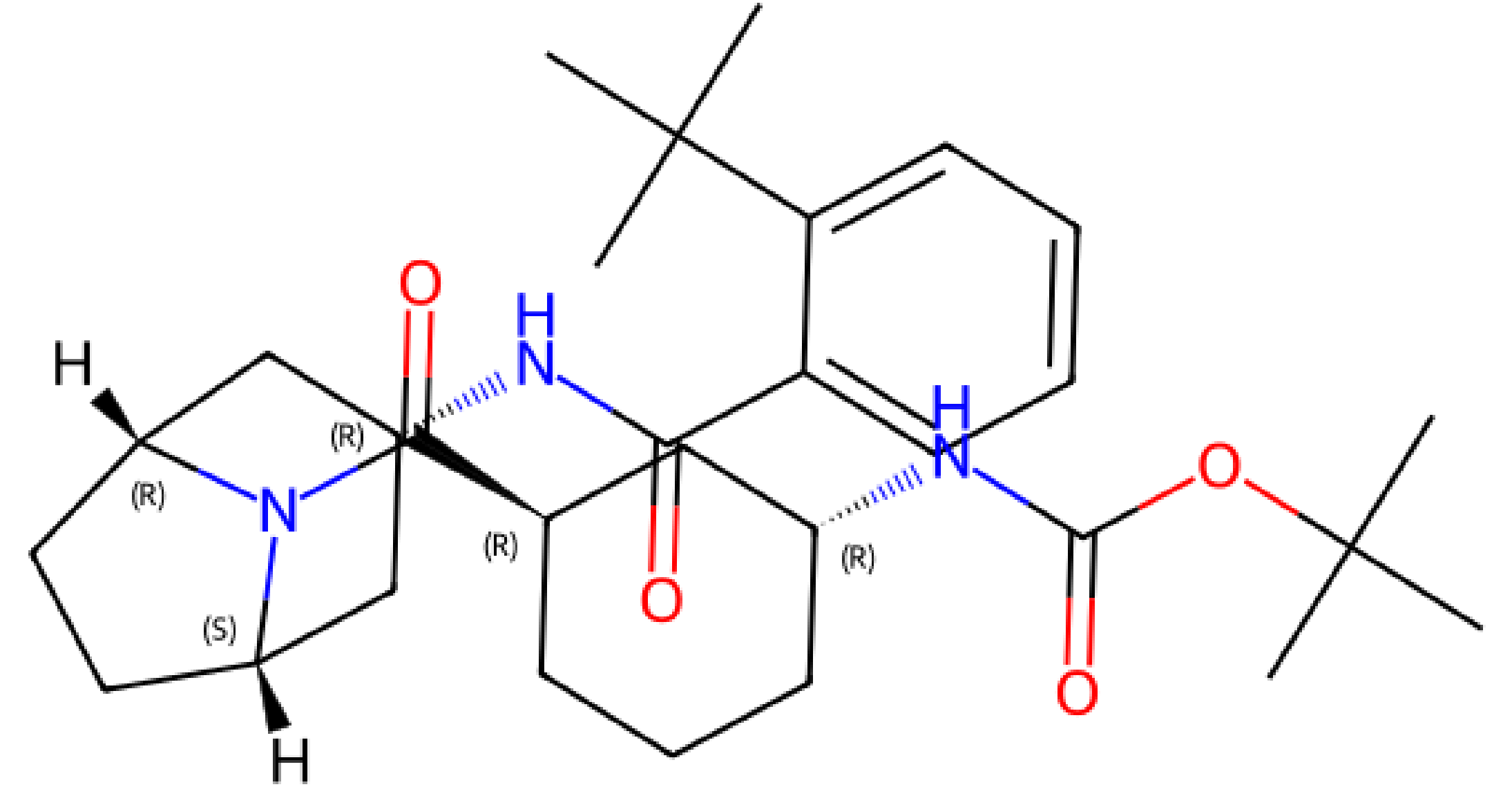}
    \caption{Source}
    \end{subfigure}%
    \hfill
    \begin{subfigure}[b]{0.4\textwidth}
    \centering
    \includegraphics[width=\linewidth,height=5cm,keepaspectratio]{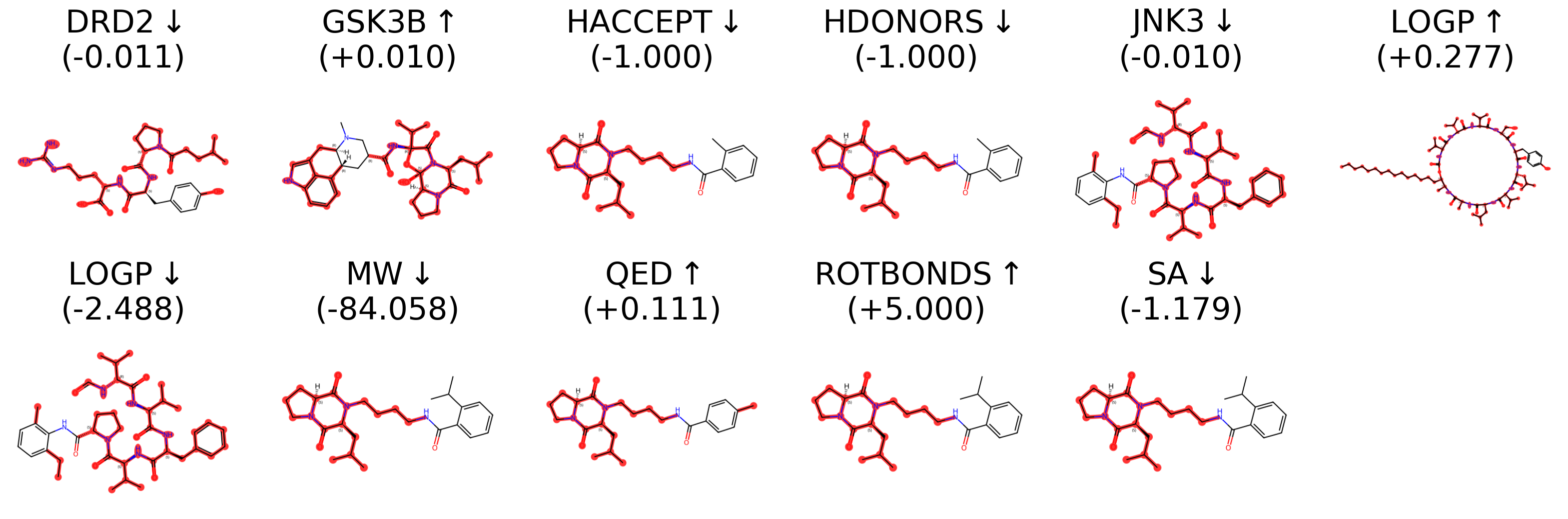}
    \caption{BioT5 (11/20 successful edits)}
    \end{subfigure}%
    \hfill
    \begin{subfigure}[b]{0.45\textwidth}
    \centering
    \includegraphics[width=\linewidth,height=5cm,keepaspectratio]{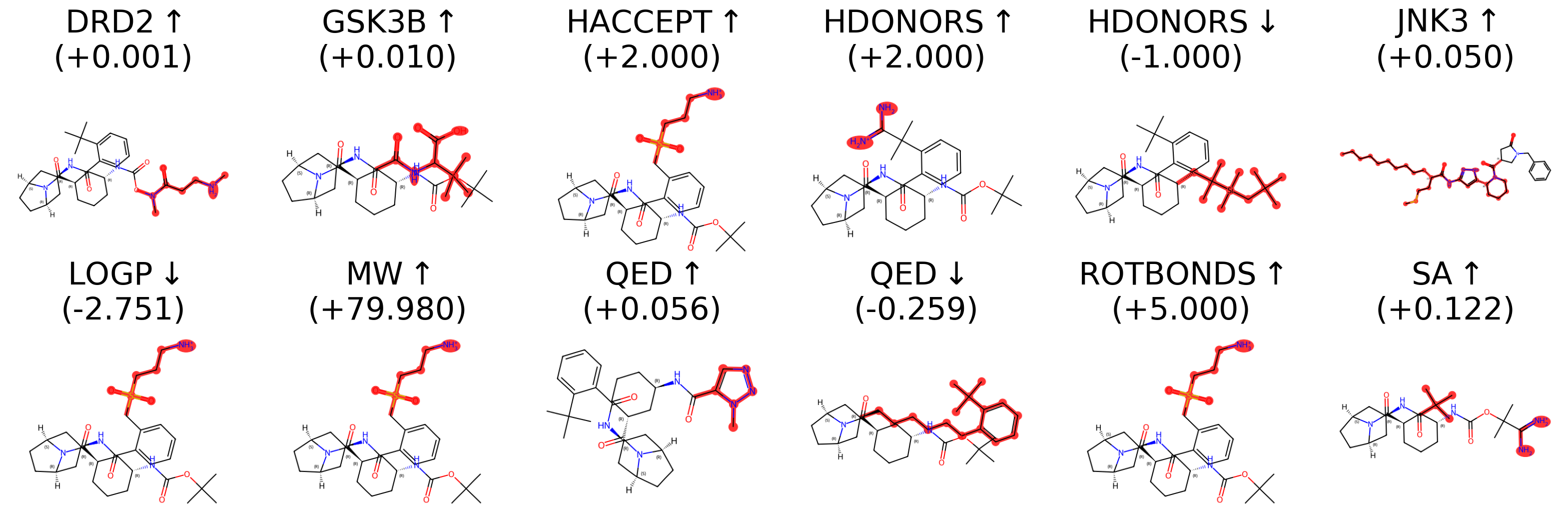}
    \caption{DrugAssist (12/20 successful edits)}
    \end{subfigure}
    
    \vspace{1em}
    
    \begin{subfigure}[b]{0.42\textwidth}
    \centering
    \includegraphics[width=\linewidth,height=5cm,keepaspectratio]{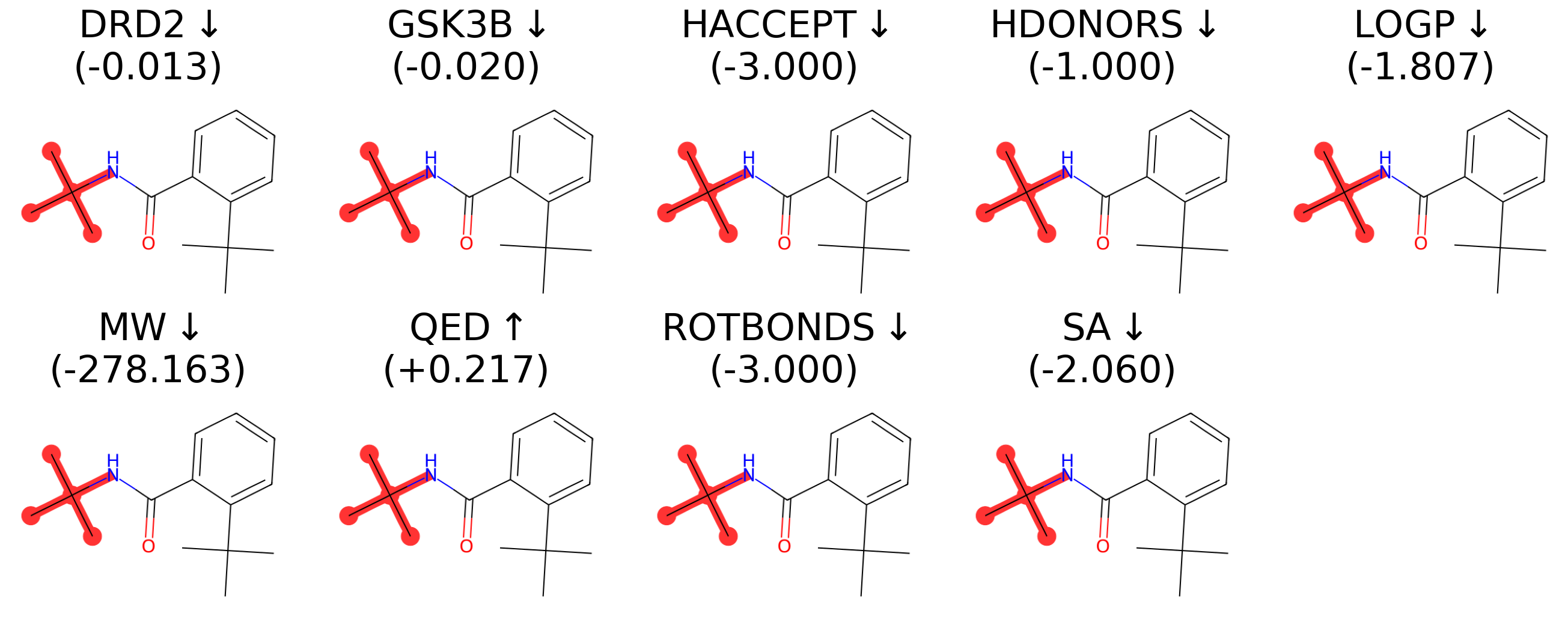}
    \caption{GeLLMO\_L (9/20 successful edits)}
    \end{subfigure}%
    \hfill
    \begin{subfigure}[b]{0.54\textwidth}
    \centering
    \includegraphics[width=\linewidth,height=5cm,keepaspectratio]{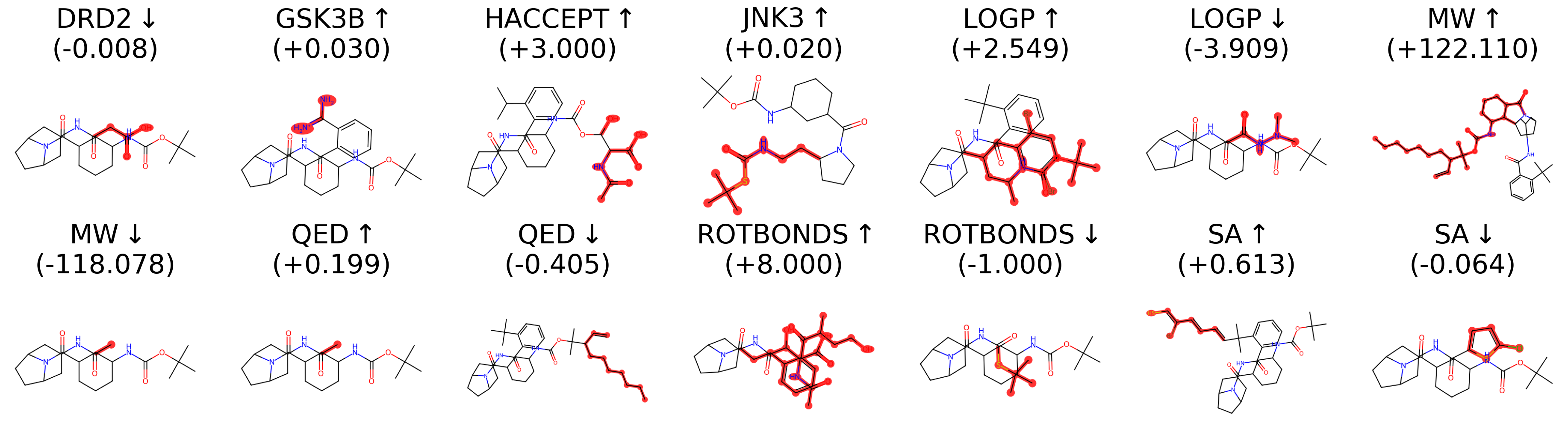}
    \caption{MolEditRL (14/20 successful edits)}
    \end{subfigure}

    \vspace{1em}

    \begin{subfigure}[b]{0.13\textwidth}
    \centering
    \includegraphics[width=\linewidth,height=5cm,keepaspectratio]{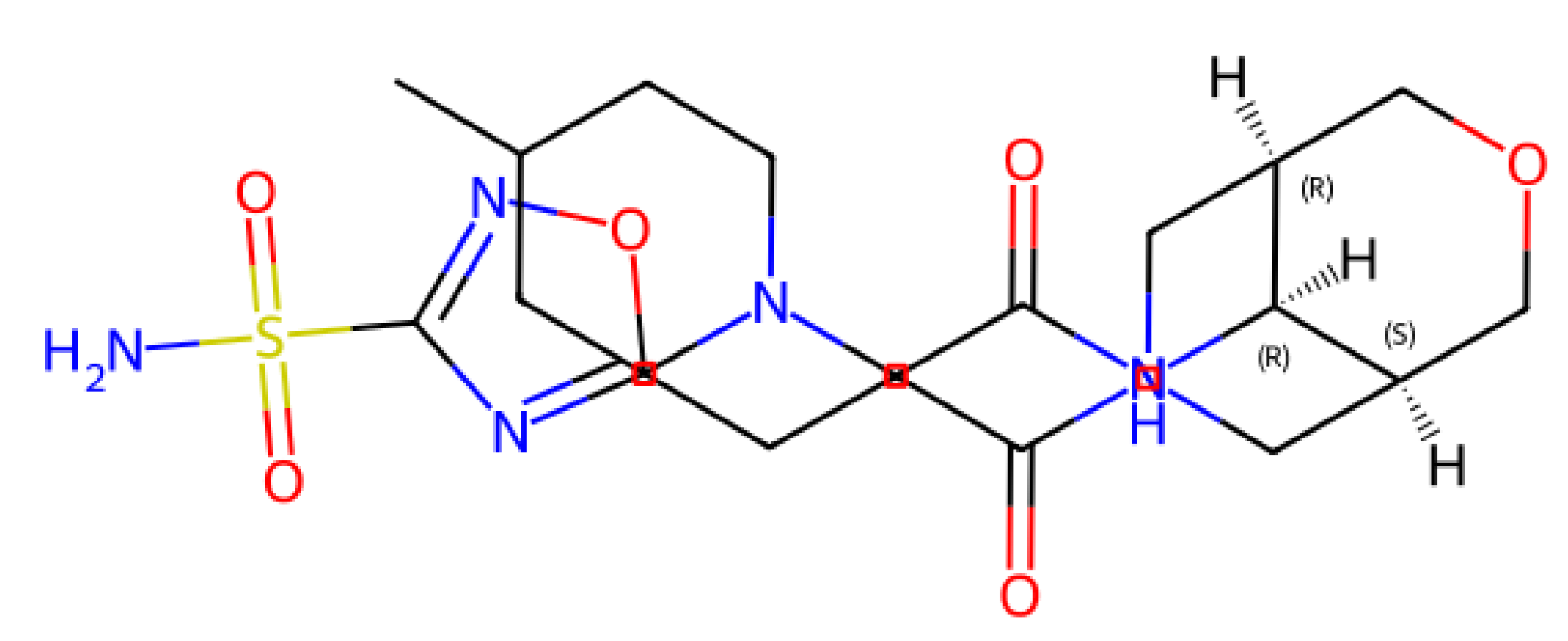}
    \caption{Source}
    \end{subfigure}%
    \hfill
    \begin{subfigure}[b]{0.41\textwidth}
    \centering
    \includegraphics[width=\linewidth,height=5cm,keepaspectratio]{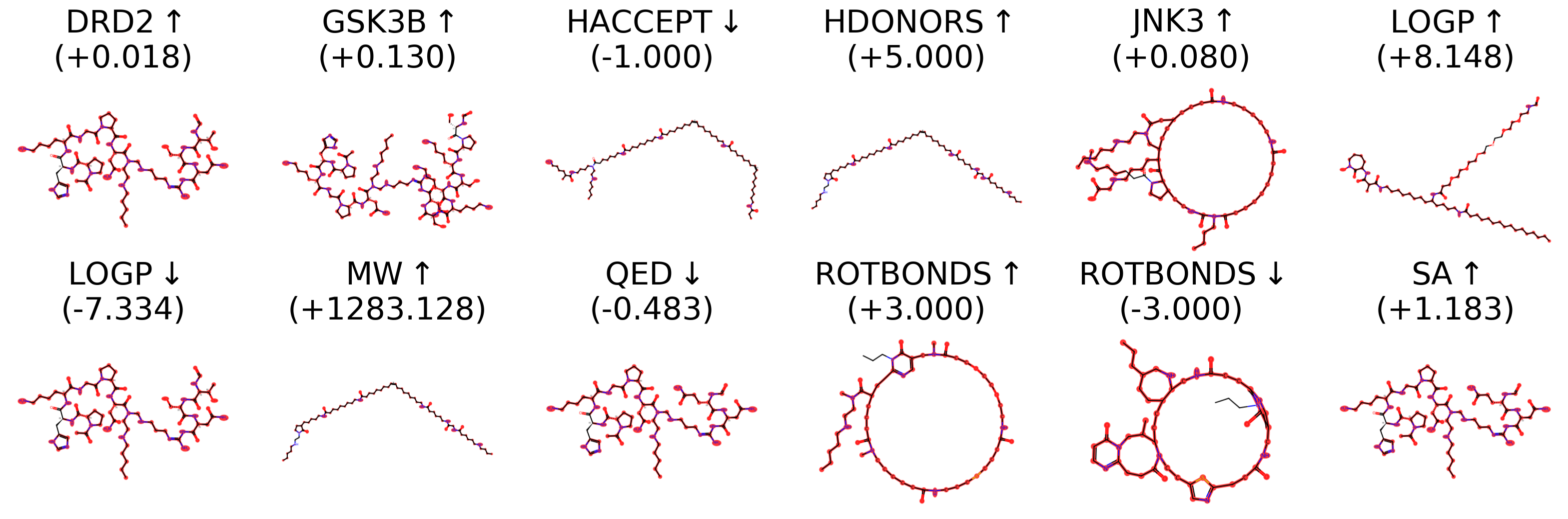}
    \caption{BioT5 (12/20 successful edits)}
    \end{subfigure}%
    \hfill
    \begin{subfigure}[b]{0.42\textwidth}
    \centering
    \includegraphics[width=\linewidth,height=5cm,keepaspectratio]{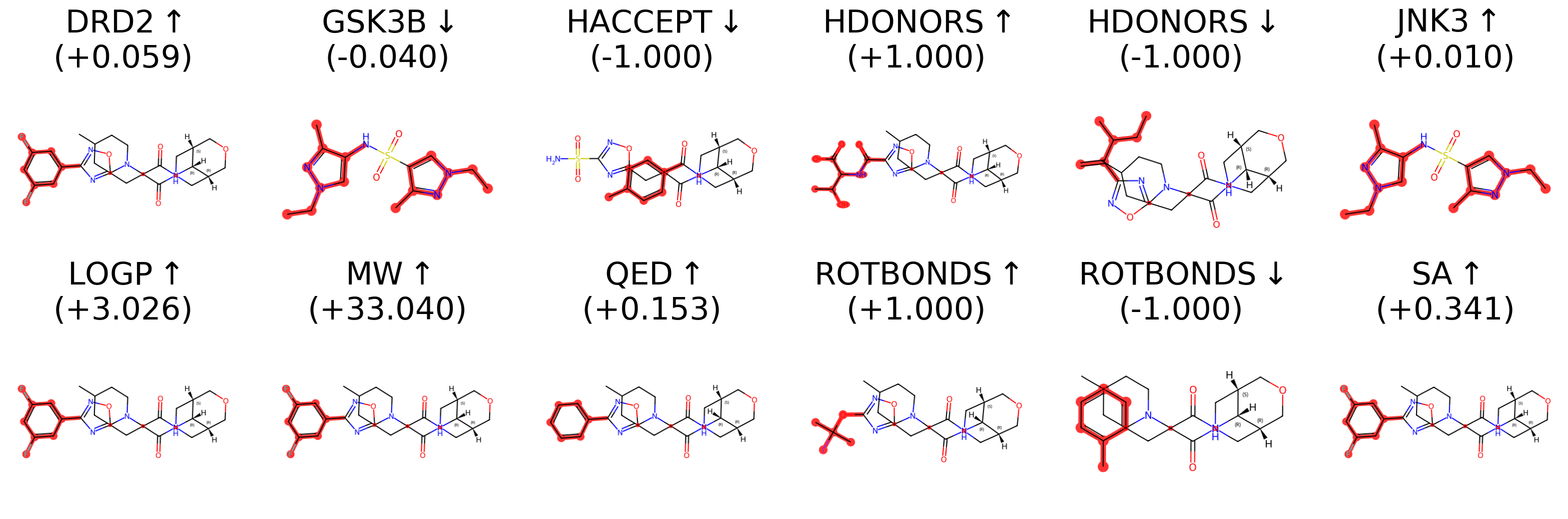}
    \caption{DrugAssist (12/20 successful edits)}
    \end{subfigure}
    
    \vspace{1em}
    
    \begin{subfigure}[b]{0.34\textwidth}
    \centering
    \includegraphics[width=\linewidth,height=5cm,keepaspectratio]{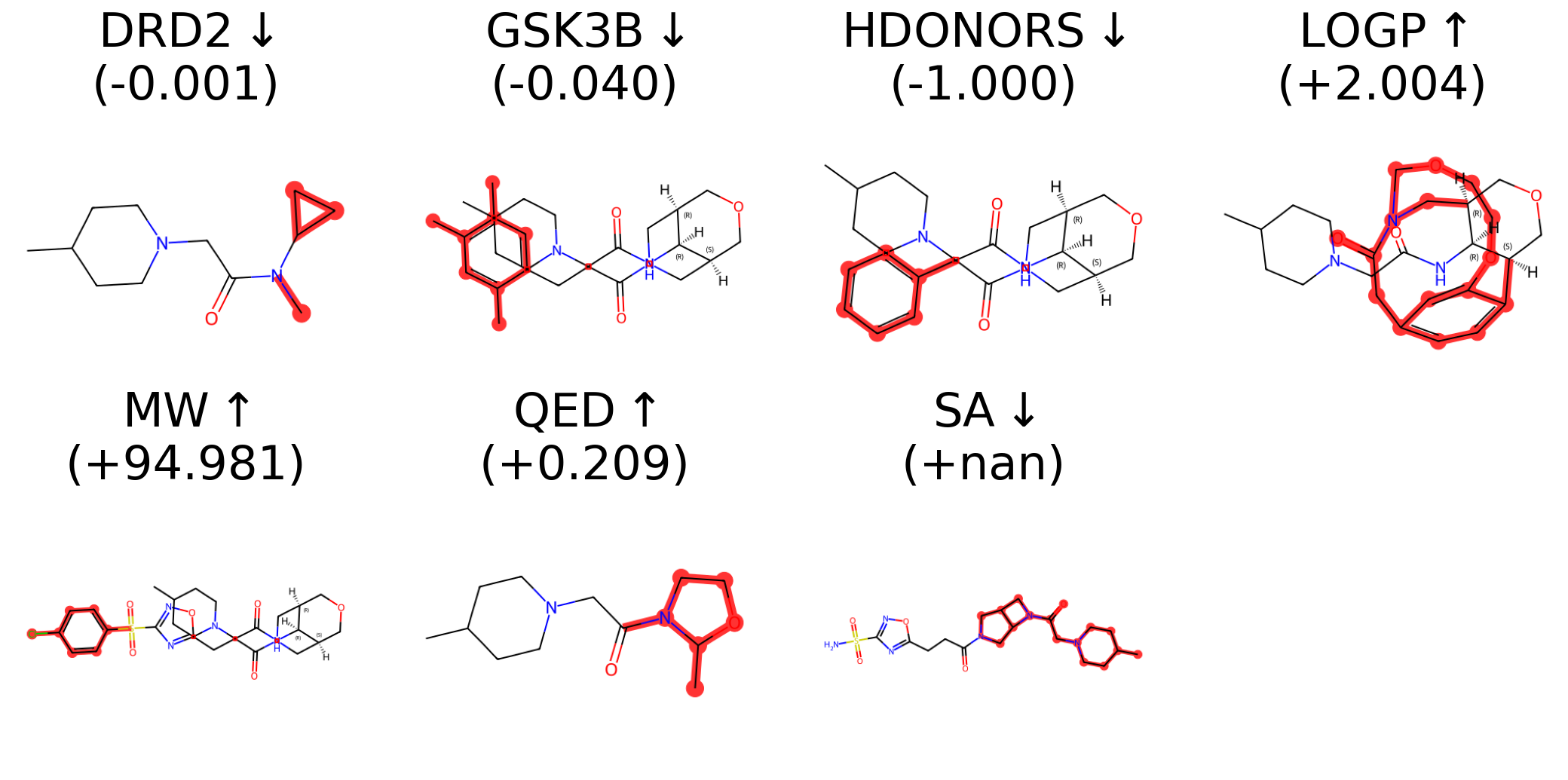}
    \caption{GeLLMO\_L (7/20 successful edits)}
    \end{subfigure}%
    \hfill
    \begin{subfigure}[b]{0.62\textwidth}
    \centering
    \includegraphics[width=\linewidth,height=5cm,keepaspectratio]{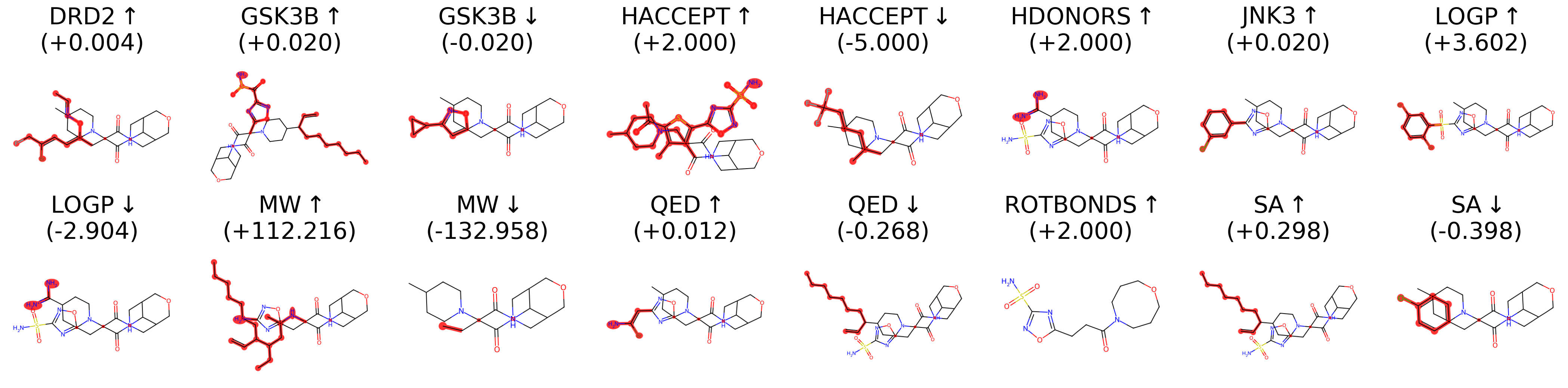}
    \caption{MolEditRL (16/20 successful edits)}
    \end{subfigure}
\caption{More visualization of edits on 20 tasks.}
\label{fig:visualize2}
\end{figure}

\begin{figure}[htbp]
    \centering
        \begin{subfigure}[b]{0.13\textwidth}
        \centering
        \includegraphics[width=\linewidth,height=5cm,keepaspectratio]{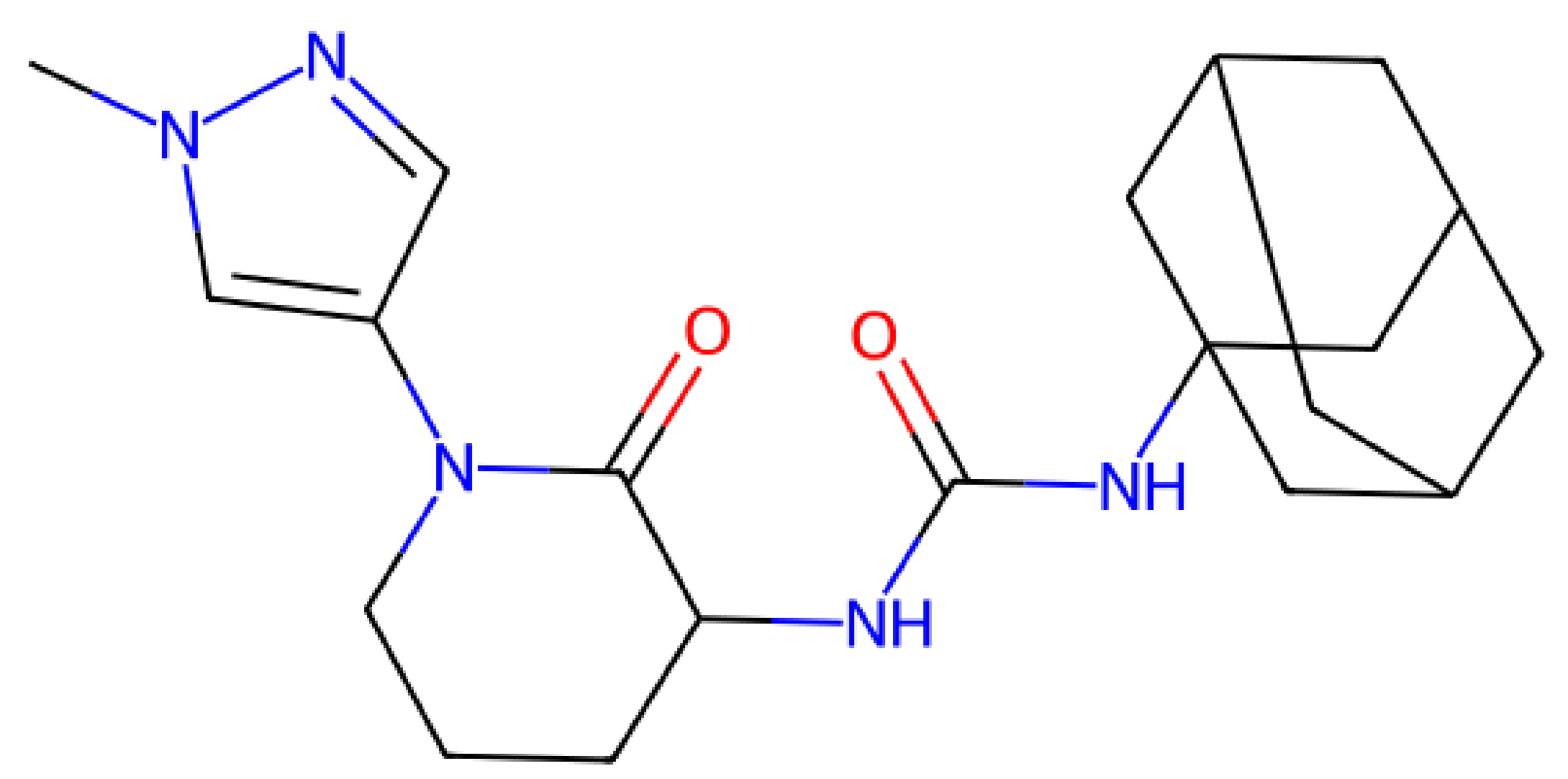}
        \caption{Source}
        \end{subfigure}%
        \hfill
        \begin{subfigure}[b]{0.31\textwidth}
        \centering
        \includegraphics[width=\linewidth,height=5cm,keepaspectratio]{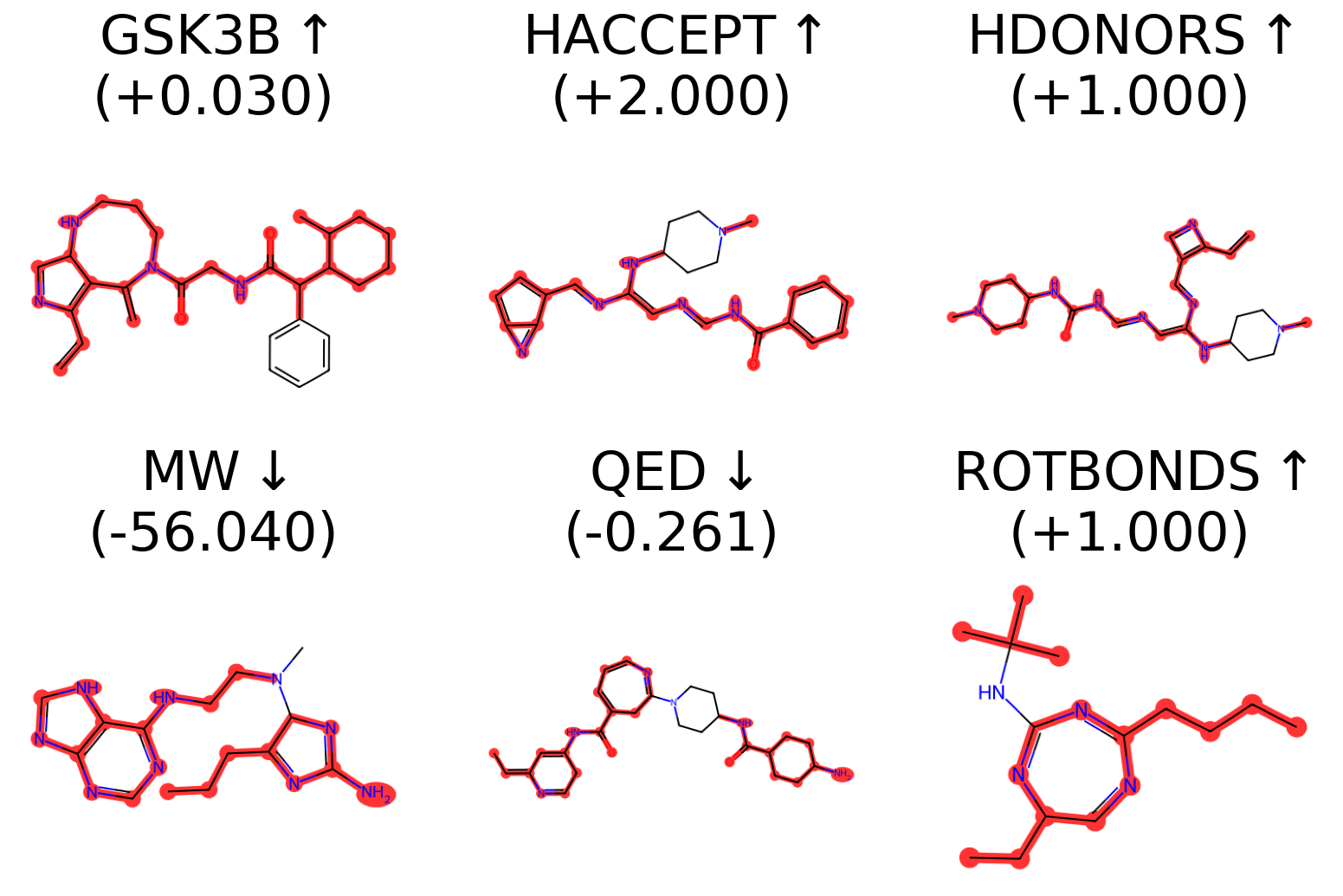}
        \caption{BioT5 (6/20 successful edits)}
        \end{subfigure}%
        \hfill
        \begin{subfigure}[b]{0.52\textwidth}
        \centering
        \includegraphics[width=\linewidth,height=5cm,keepaspectratio]{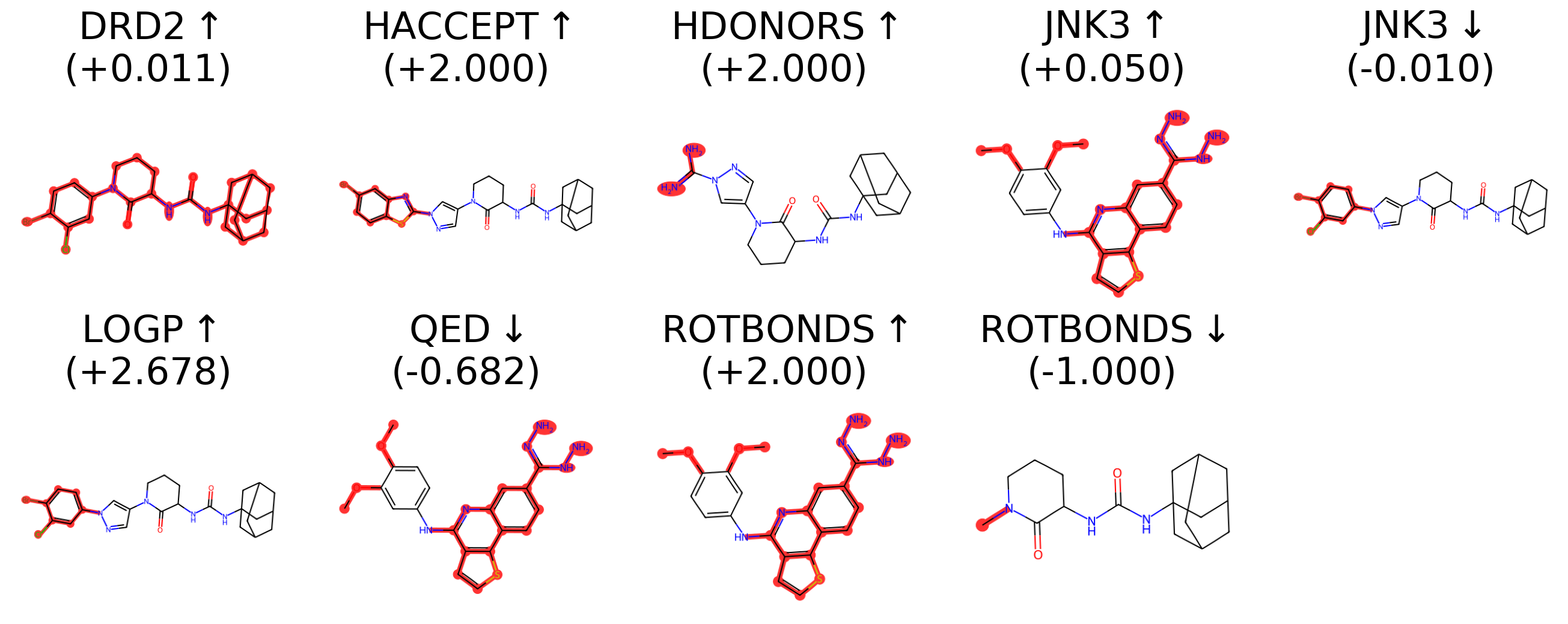}
        \caption{DrugAssist (9/20 successful edits)}
        \end{subfigure}
        
        \vspace{1em}
        
        \begin{subfigure}[b]{0.34\textwidth}
        \centering
        \includegraphics[width=\linewidth,height=5cm,keepaspectratio]{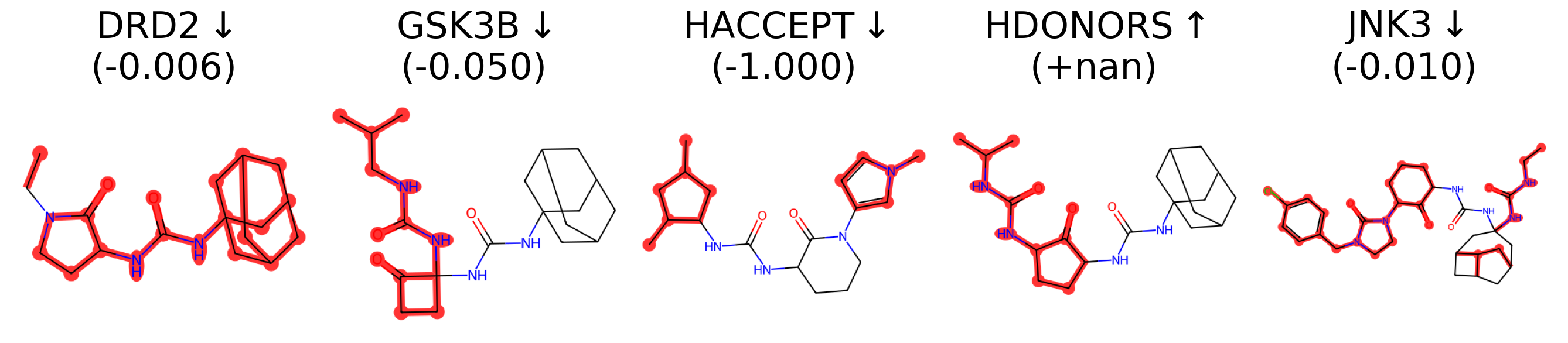}
        \caption{GeLLMO\_L (5/20 successful edits)}
        \end{subfigure}%
        \hfill
        \begin{subfigure}[b]{0.62\textwidth}
        \centering
        \includegraphics[width=\linewidth,height=5cm,keepaspectratio]{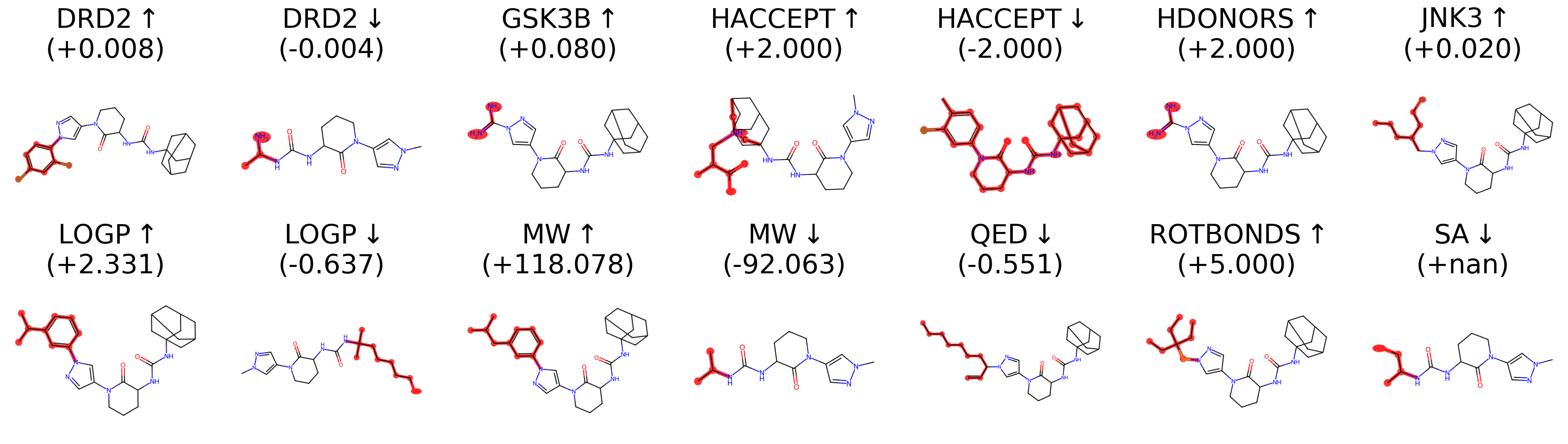}
        \caption{MolEditRL (14/20 successful edits)}
        \end{subfigure}
    
        \vspace{1em}
    
        \begin{subfigure}[b]{0.13\textwidth}
        \centering
        \includegraphics[width=\linewidth,height=5cm,keepaspectratio]{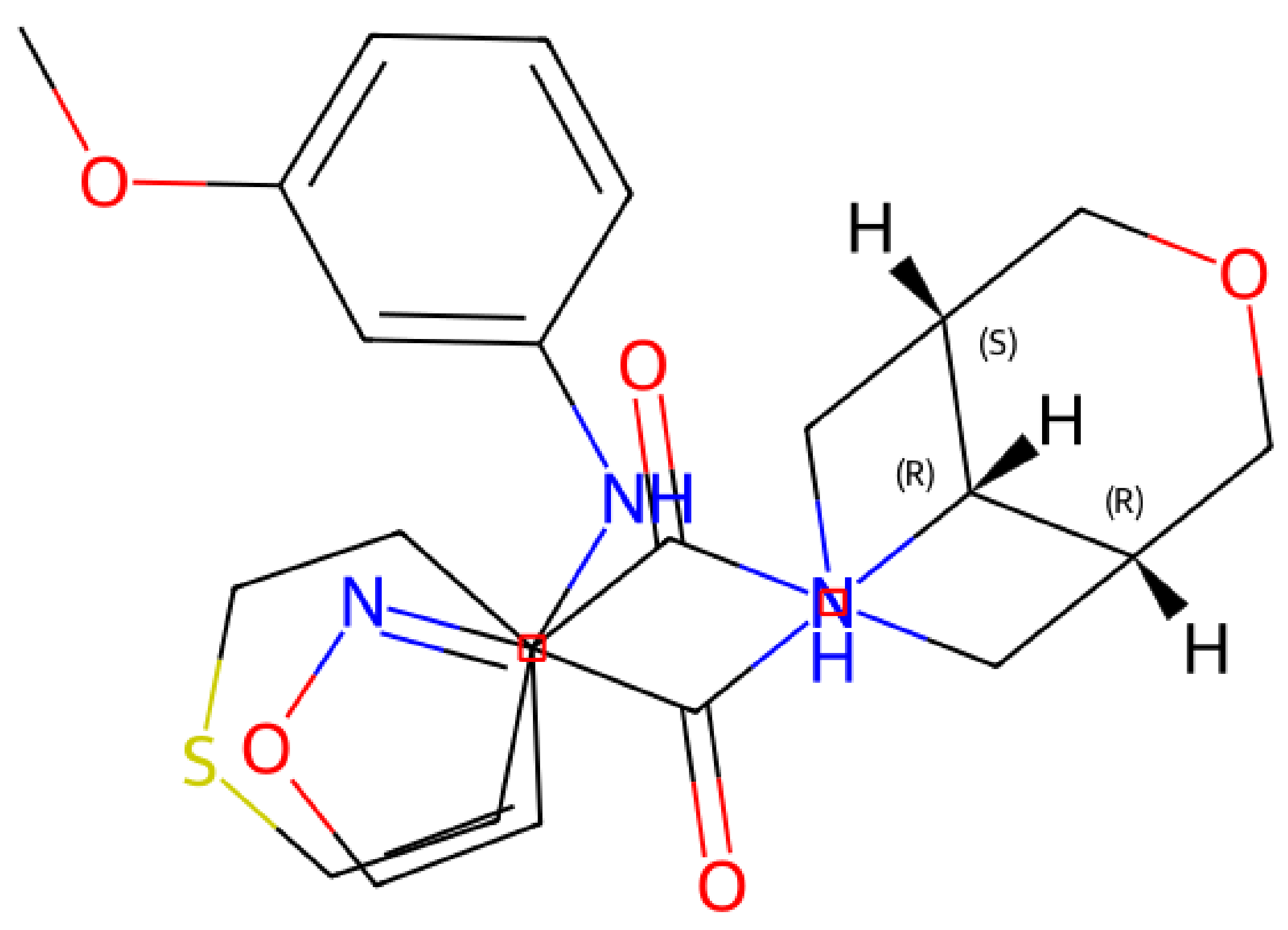}
        \caption{Source}
        \end{subfigure}%
        \hfill
        \begin{subfigure}[b]{0.41\textwidth}
        \centering
        \includegraphics[width=\linewidth,height=5cm,keepaspectratio]{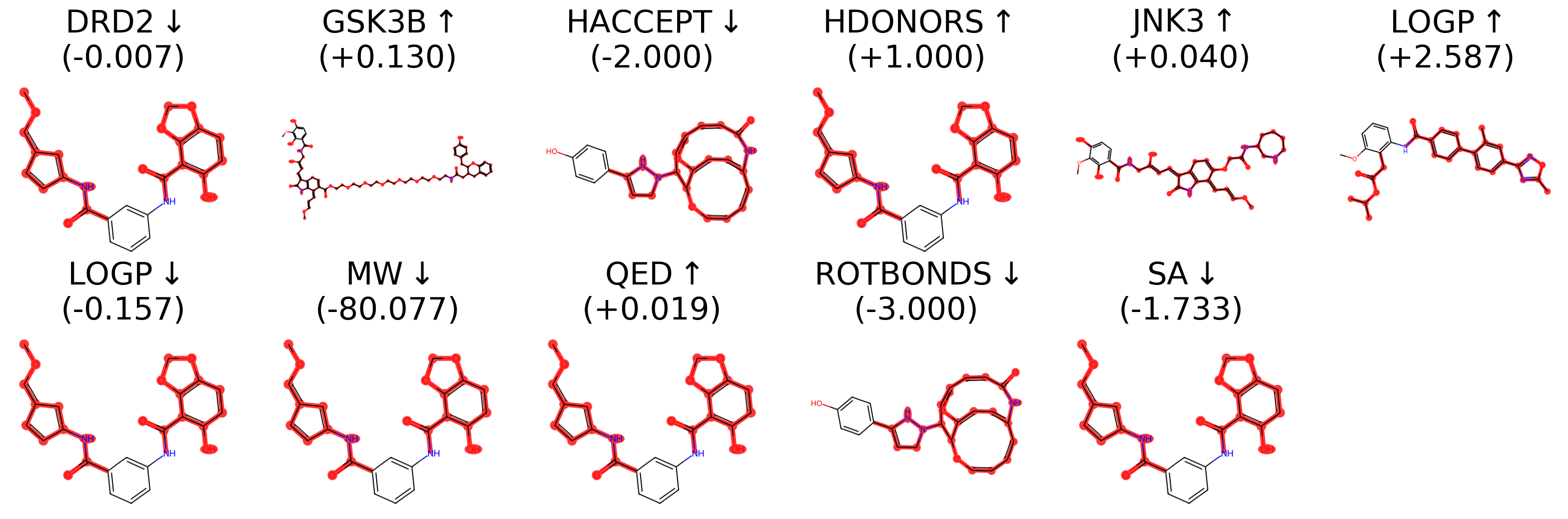}
        \caption{BioT5 (11/20 successful edits)}
        \end{subfigure}%
        \hfill
        \begin{subfigure}[b]{0.42\textwidth}
        \centering
        \includegraphics[width=\linewidth,height=5cm,keepaspectratio]{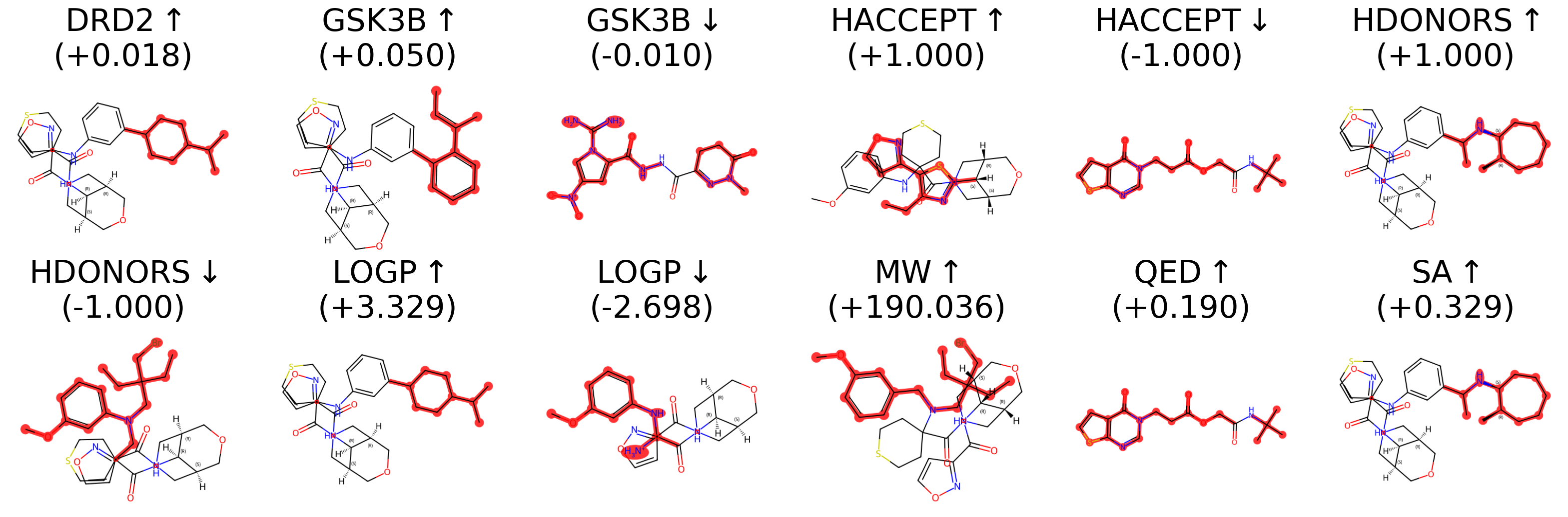}
        \caption{DrugAssist (12/20 successful edits)}
        \end{subfigure}
        
        \vspace{1em}
        
        \begin{subfigure}[b]{0.3\textwidth}
        \centering
        \includegraphics[width=\linewidth,height=5cm,keepaspectratio]{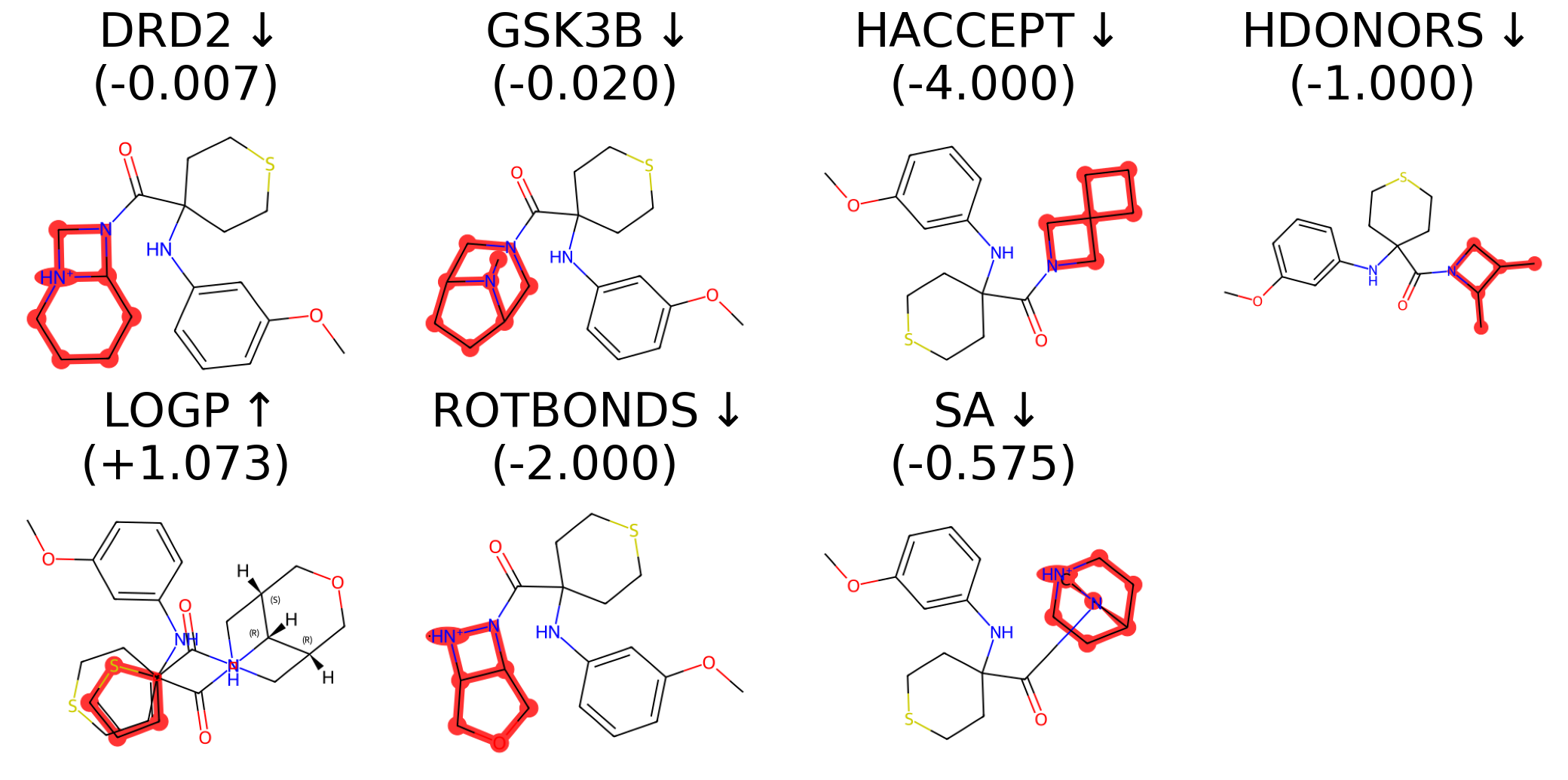}
        \caption{GeLLMO\_L (7/20 successful edits)}
        \end{subfigure}%
        \hfill
        \begin{subfigure}[b]{0.62\textwidth}
        \centering
        \includegraphics[width=\linewidth,height=5cm,keepaspectratio]{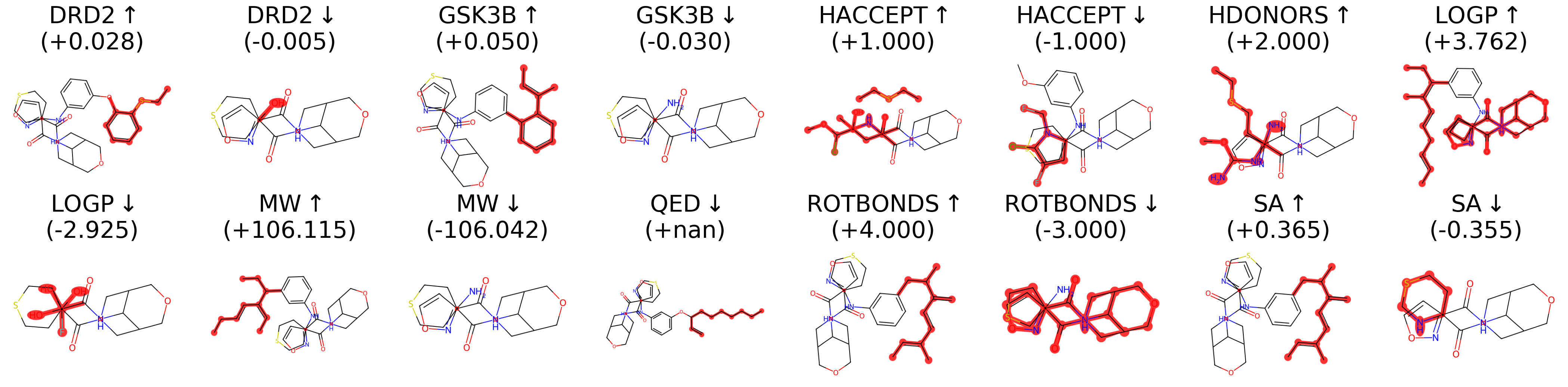}
        \caption{MolEditRL (16/20 successful edits)}
        \end{subfigure}
    
        \vspace{1em}
    
        \begin{subfigure}[b]{0.13\textwidth}
        \centering
        \includegraphics[width=\linewidth,height=5cm,keepaspectratio]{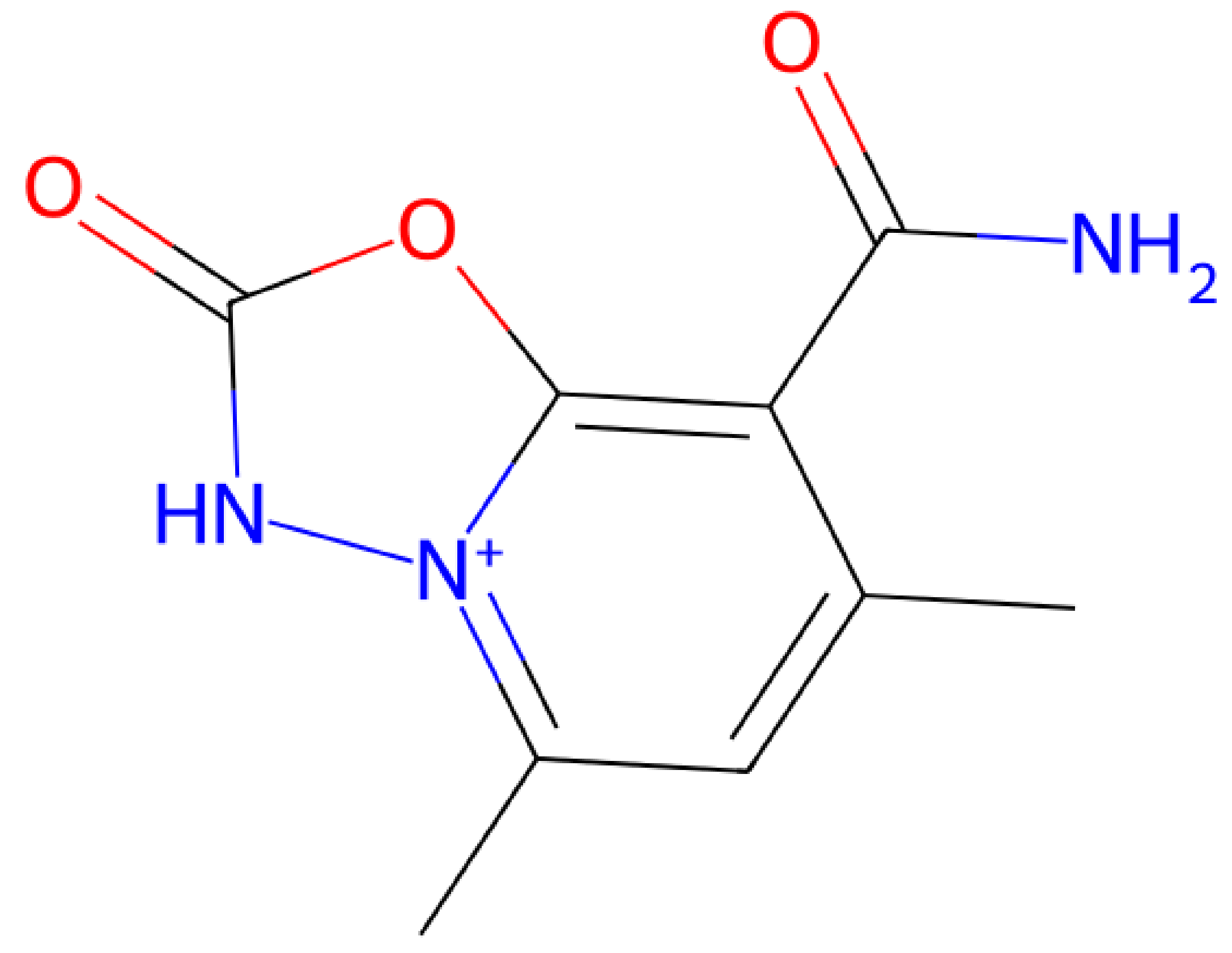}
        \caption{Source}
        \end{subfigure}%
        \hfill
        \begin{subfigure}[b]{0.41\textwidth}
        \centering
        \includegraphics[width=\linewidth,height=5cm,keepaspectratio]{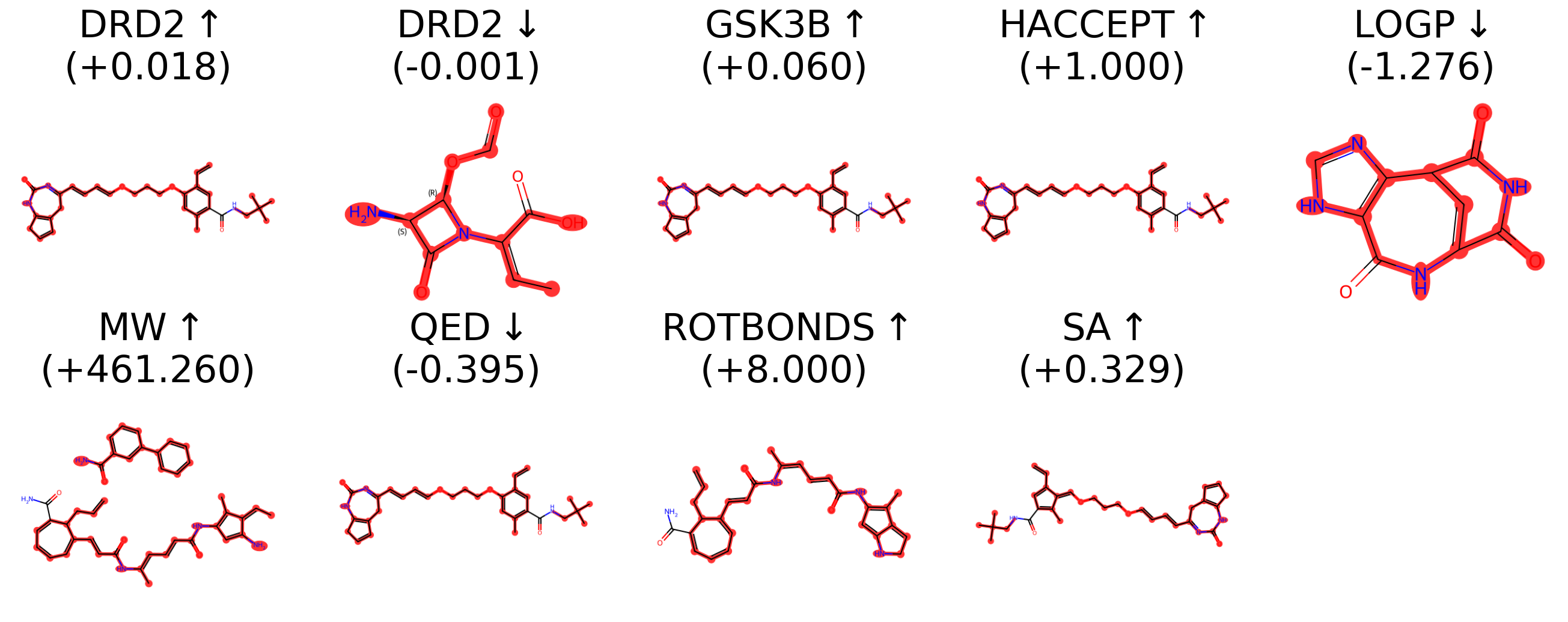}
        \caption{BioT5 (9/20 successful edits)}
        \end{subfigure}%
        \hfill
        \begin{subfigure}[b]{0.42\textwidth}
        \centering
        \includegraphics[width=\linewidth,height=5cm,keepaspectratio]{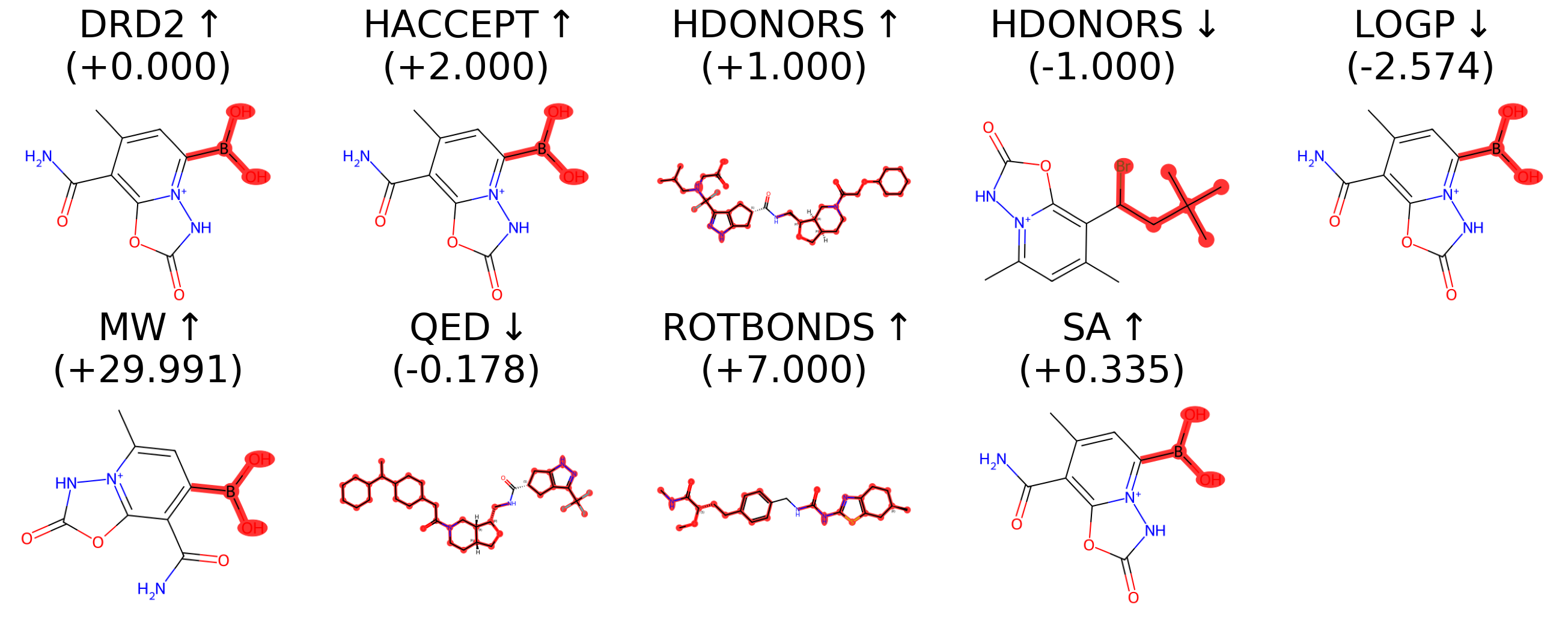}
        \caption{DrugAssist (9/20 successful edits)}
        \end{subfigure}
        
        \vspace{1em}
        
        \begin{subfigure}[b]{0.44\textwidth}
        \centering
        \includegraphics[width=\linewidth,height=5cm,keepaspectratio]{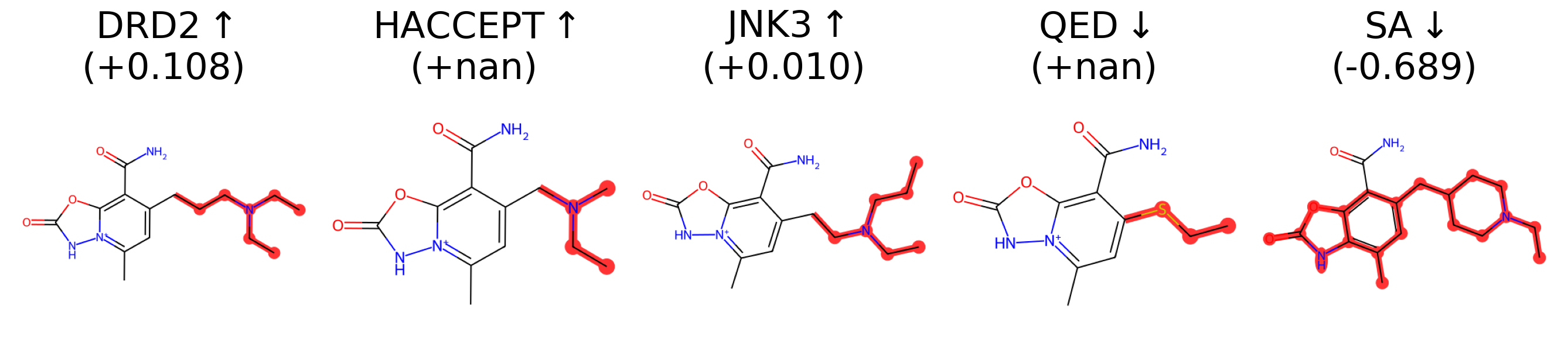}
        \caption{GeLLMO\_L (5/20 successful edits)}
        \end{subfigure}%
        \hfill
        \begin{subfigure}[b]{0.52\textwidth}
        \centering
        \includegraphics[width=\linewidth,height=5cm,keepaspectratio]{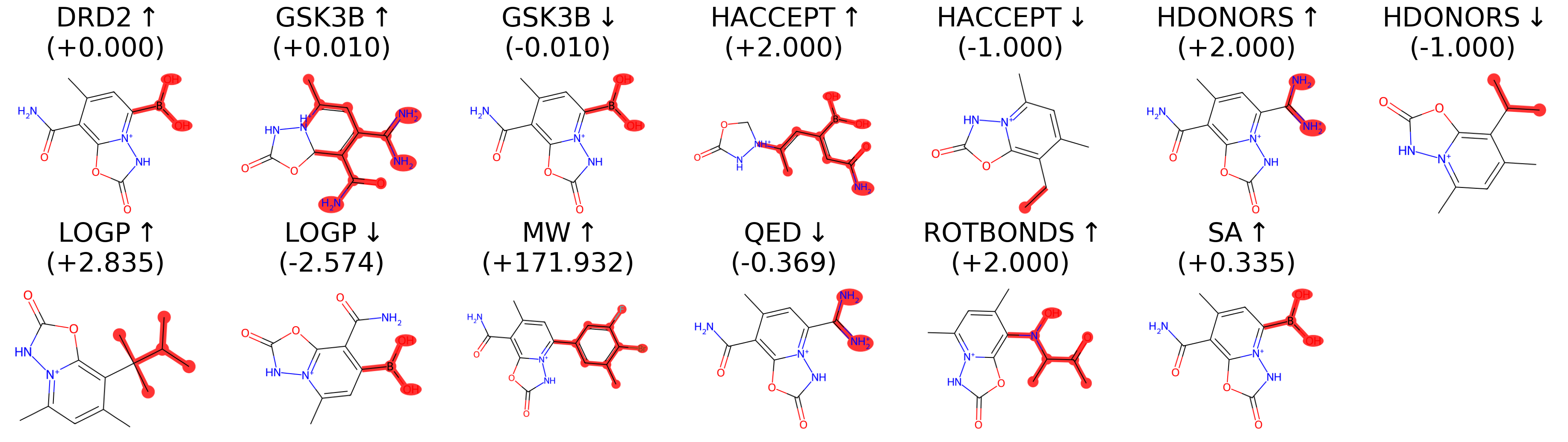}
        \caption{MolEditRL (13/20 successful edits)}
        \end{subfigure}
    \caption{More visualization of edits on 20 tasks.}
    \label{fig:visualize3}
    \end{figure}

\begin{figure}[htbp]
  \centering
  \begin{subfigure}[b]{0.85\textwidth}
    \centering
    \includegraphics[width=\textwidth]{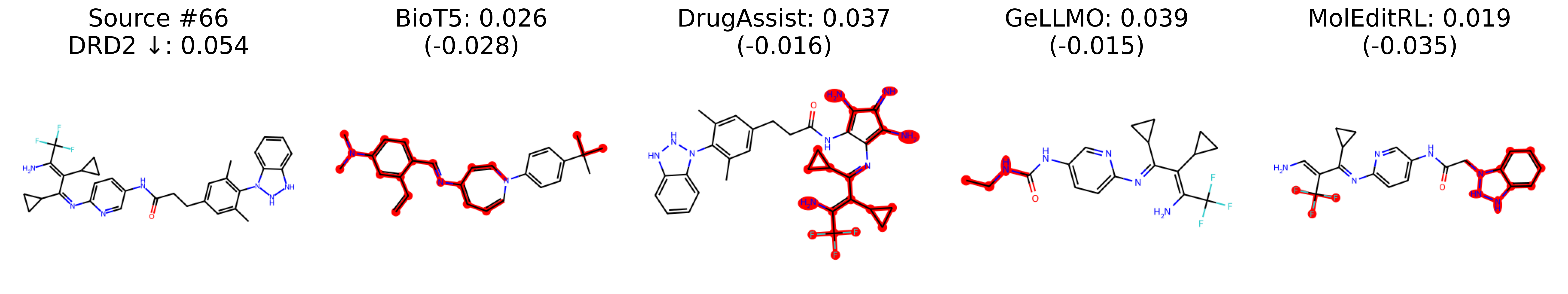}
  \end{subfigure}%

  \begin{subfigure}[b]{0.85\textwidth}
    \centering
    \includegraphics[width=\textwidth]{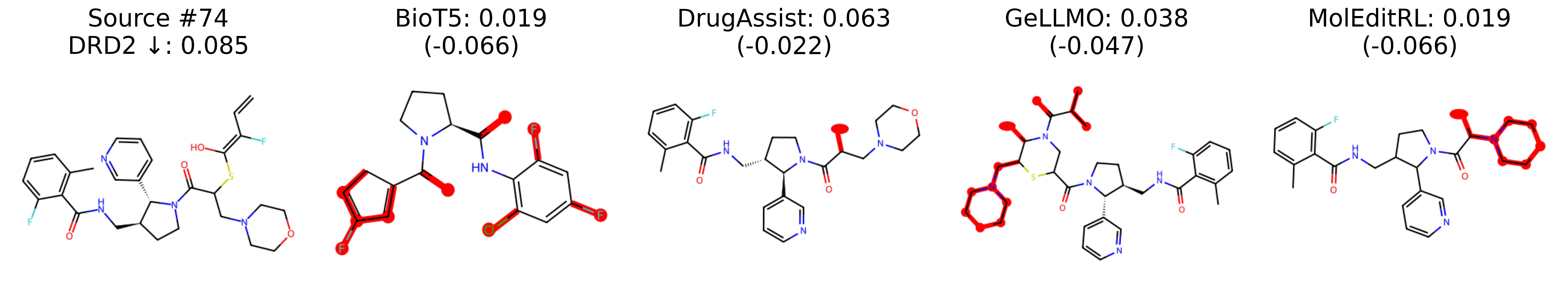}
  \end{subfigure}

  \begin{subfigure}[b]{0.85\textwidth}
    \centering
    \includegraphics[width=\textwidth]{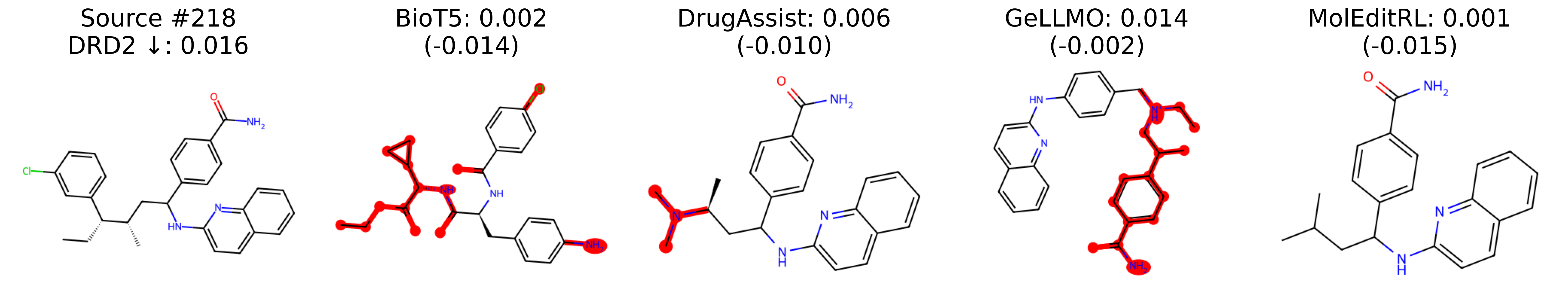}
  \end{subfigure}

  \begin{subfigure}[b]{0.85\textwidth}
    \centering
    \includegraphics[width=\textwidth]{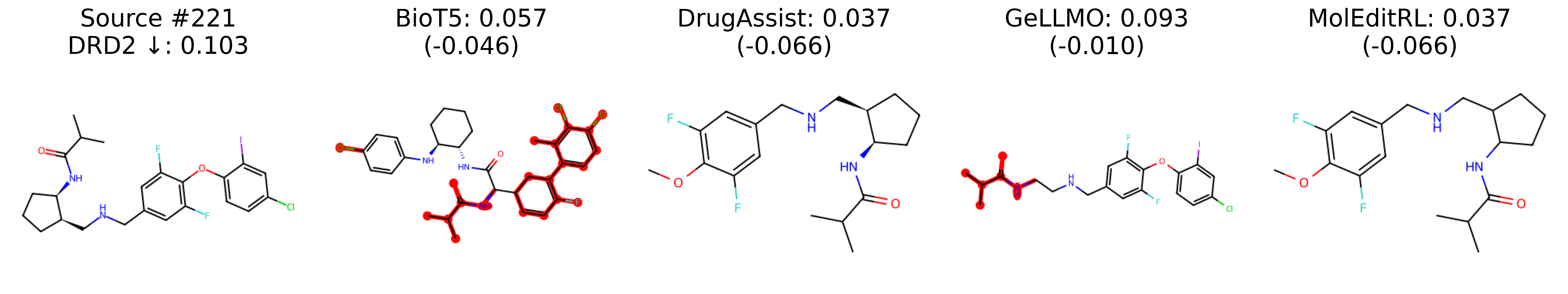}
  \end{subfigure}%

  \begin{subfigure}[b]{0.85\textwidth}
    \centering
    \includegraphics[width=\textwidth]{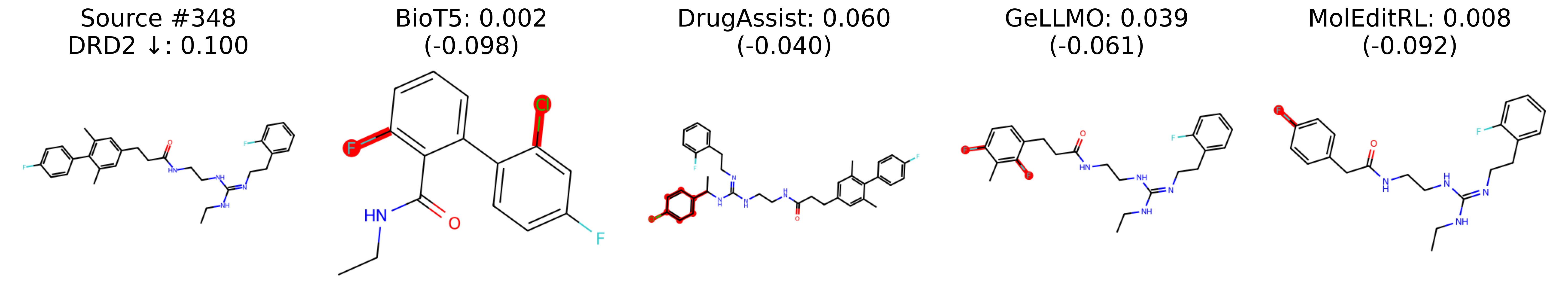}
  \end{subfigure}

  \begin{subfigure}[b]{0.85\textwidth}
    \centering
    \includegraphics[width=\textwidth]{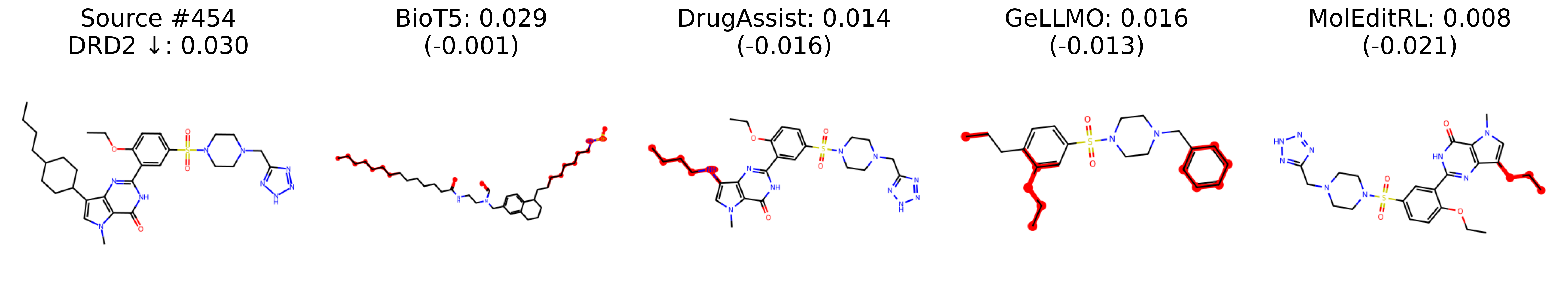}
  \end{subfigure}%

  \begin{subfigure}[b]{0.85\textwidth}
    \centering
    \includegraphics[width=\textwidth]{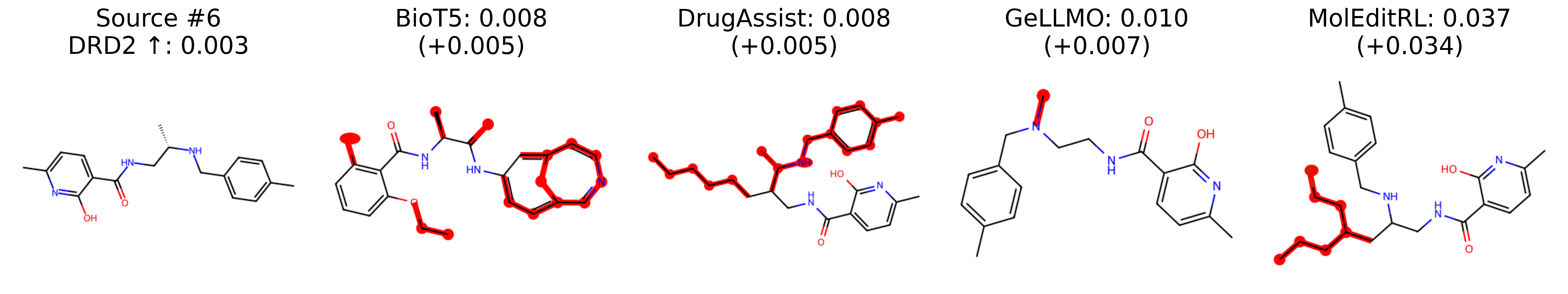}
  \end{subfigure}

  \begin{subfigure}[b]{0.85\textwidth}
    \centering
    \includegraphics[width=\textwidth]{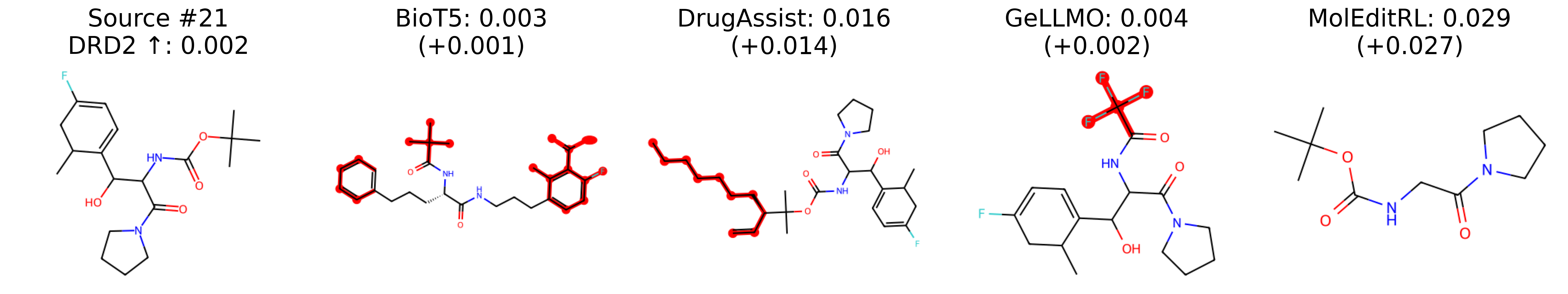}
  \end{subfigure}

  \begin{subfigure}[b]{0.85\textwidth}
    \centering
    \includegraphics[width=\textwidth]{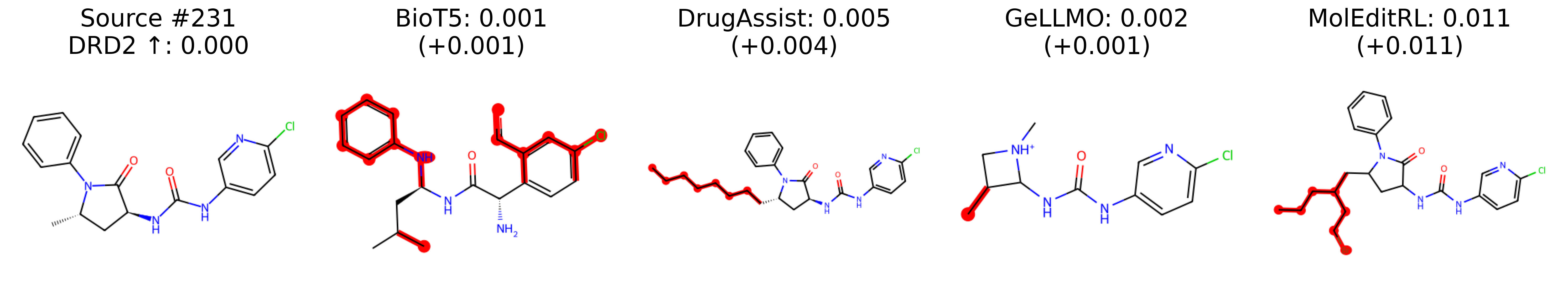}
  \end{subfigure}

  \begin{subfigure}[b]{0.85\textwidth}
    \centering
    \includegraphics[width=\textwidth]{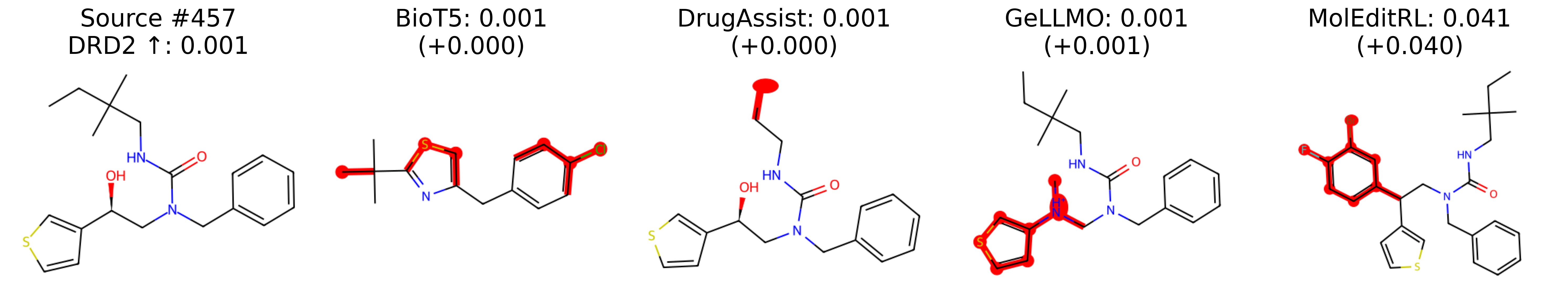}
  \end{subfigure}

  \caption{Qualitative comparison of molecular editing methods.}
  \label{fig:1}
\end{figure}

\begin{figure}[htbp]
  \centering
  \begin{subfigure}[b]{0.85\textwidth}
    \centering
    \includegraphics[width=\textwidth]{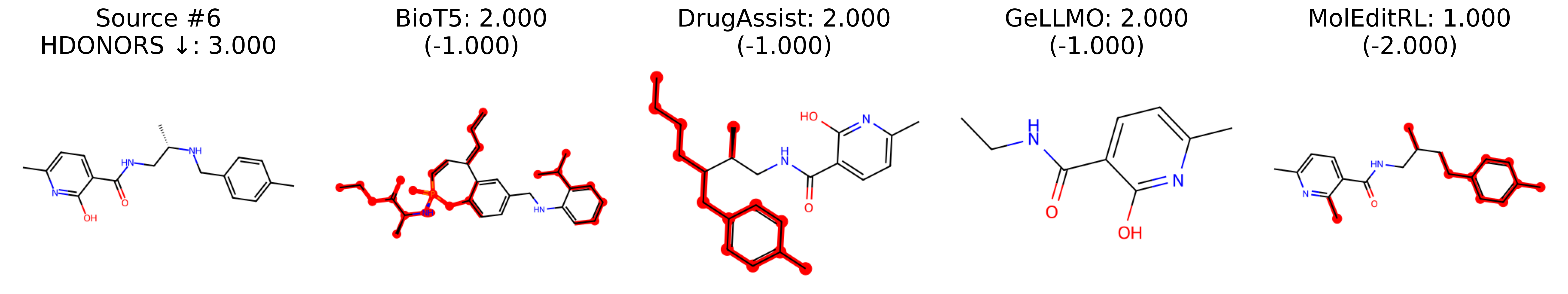}
  \end{subfigure}%

  \begin{subfigure}[b]{0.85\textwidth}
    \centering
    \includegraphics[width=\textwidth]{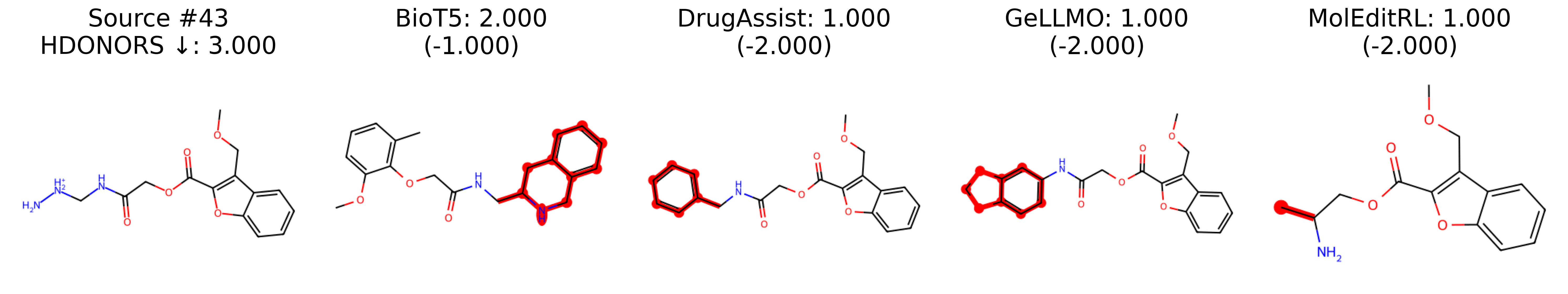}
  \end{subfigure}

  \begin{subfigure}[b]{0.85\textwidth}
    \centering
    \includegraphics[width=\textwidth]{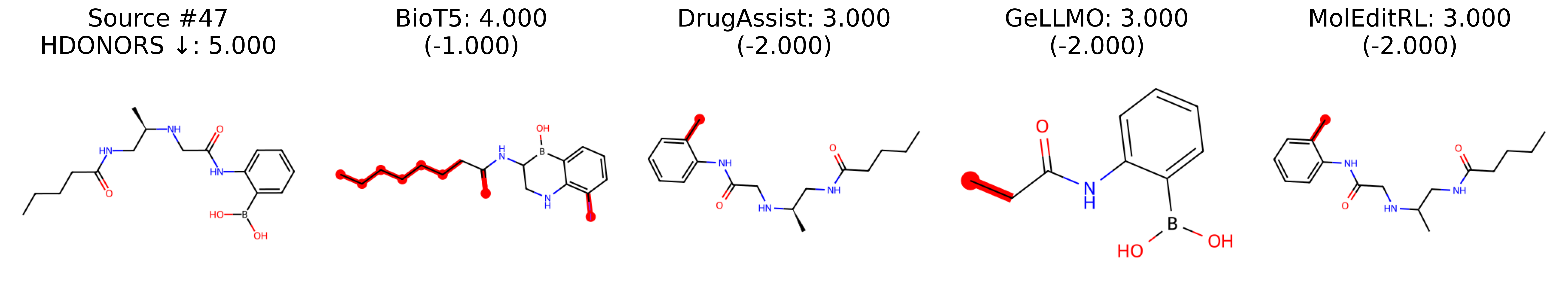}
  \end{subfigure}

  \begin{subfigure}[b]{0.85\textwidth}
    \centering
    \includegraphics[width=\textwidth]{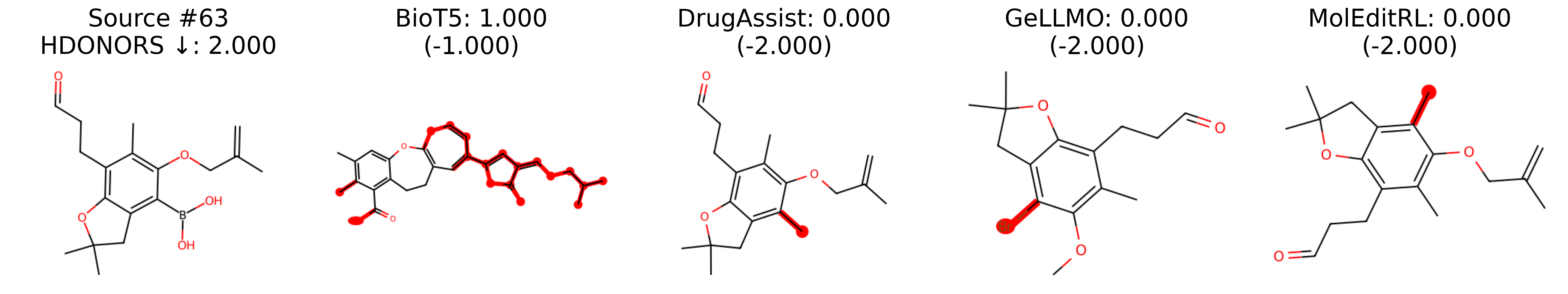}
  \end{subfigure}%

  \begin{subfigure}[b]{0.85\textwidth}
    \centering
    \includegraphics[width=\textwidth]{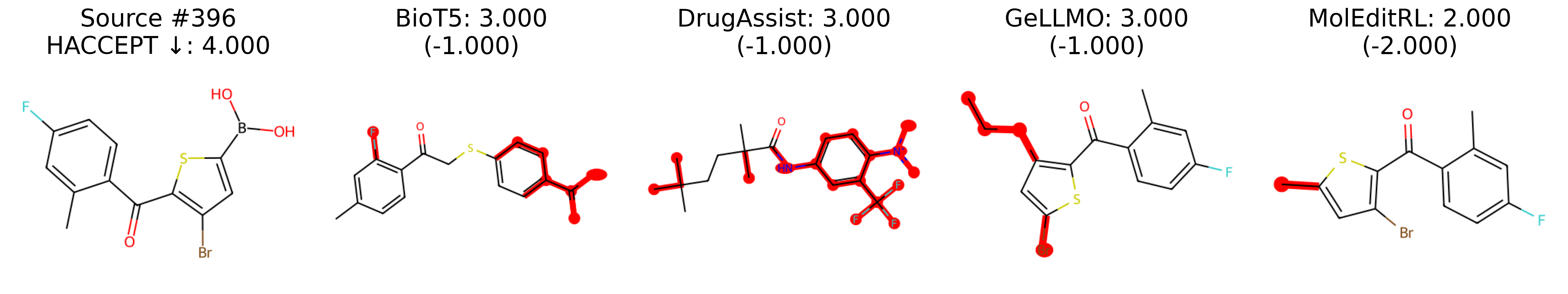}
  \end{subfigure}

  \begin{subfigure}[b]{0.85\textwidth}
    \centering
    \includegraphics[width=\textwidth]{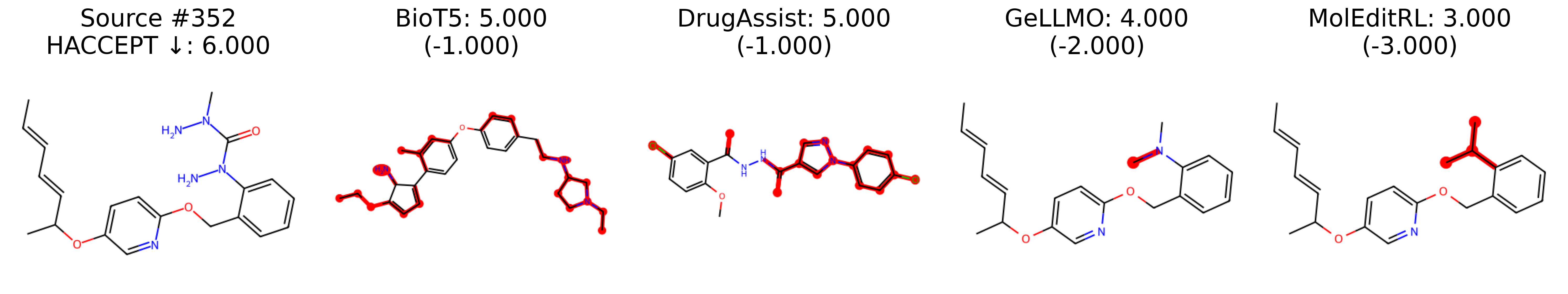}
  \end{subfigure}%

  \begin{subfigure}[b]{0.85\textwidth}
    \centering
    \includegraphics[width=\textwidth]{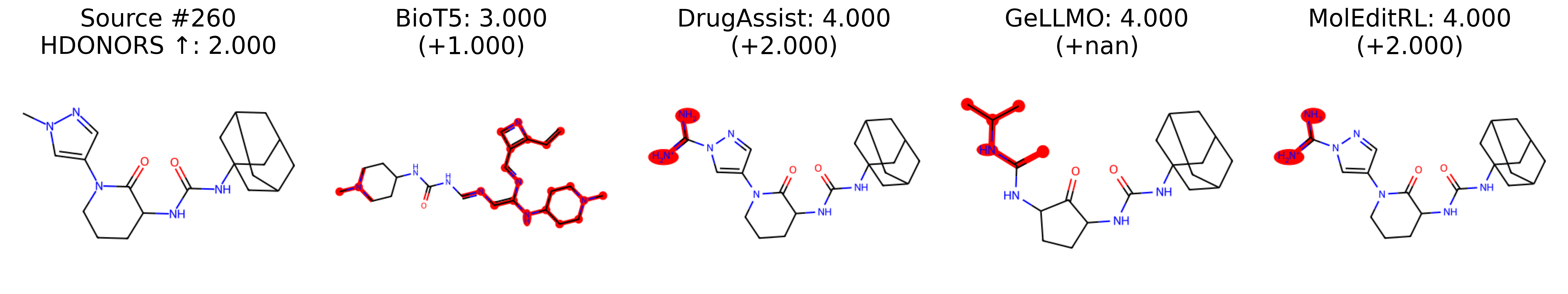}
  \end{subfigure}

  \begin{subfigure}[b]{0.85\textwidth}
    \centering
    \includegraphics[width=\textwidth]{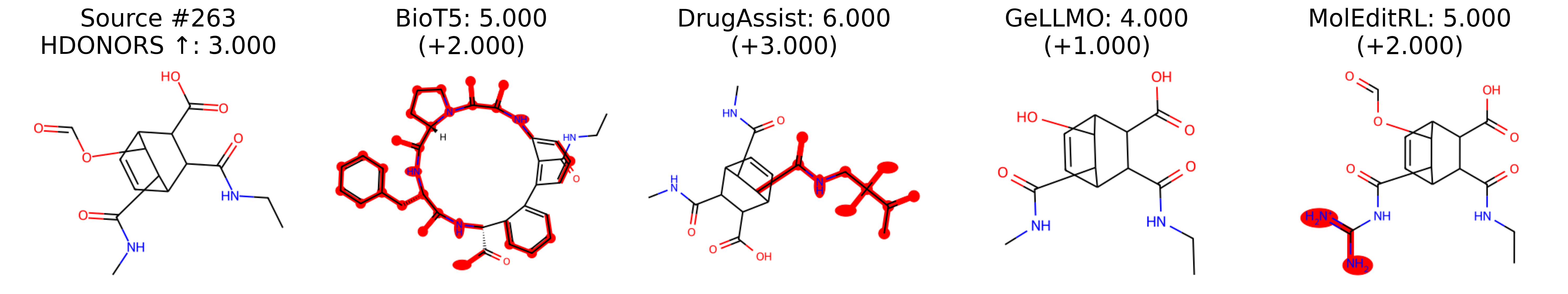}
  \end{subfigure}

  \begin{subfigure}[b]{0.85\textwidth}
    \centering
    \includegraphics[width=\textwidth]{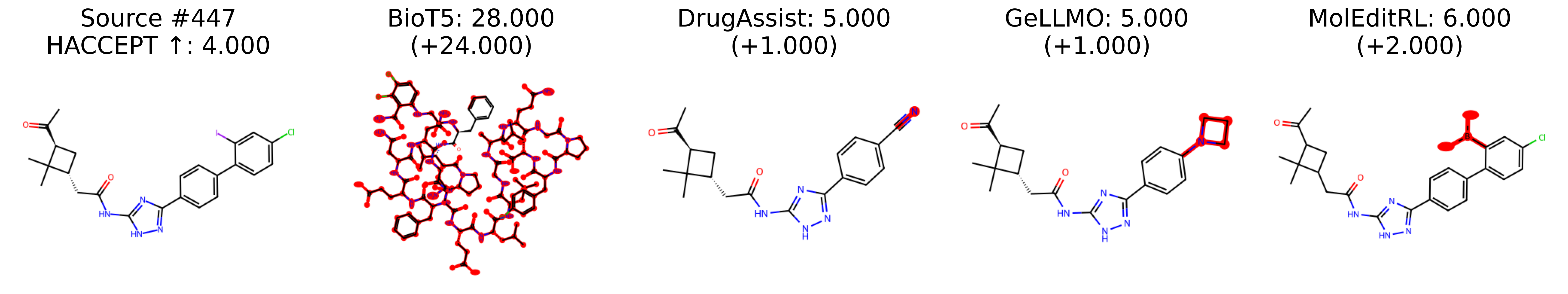}
  \end{subfigure}

  \begin{subfigure}[b]{0.85\textwidth}
    \centering
    \includegraphics[width=\textwidth]{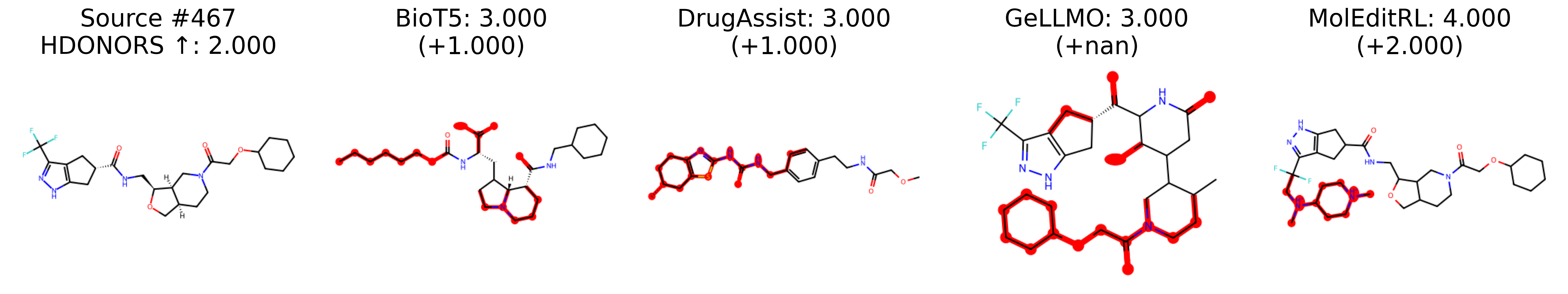}
  \end{subfigure}

  \caption{Qualitative comparison of molecular editing methods.}
  \label{fig:2}
\end{figure}

\begin{figure}[htbp]
  \centering
  \begin{subfigure}[b]{0.85\textwidth}
    \centering
    \includegraphics[width=\textwidth]{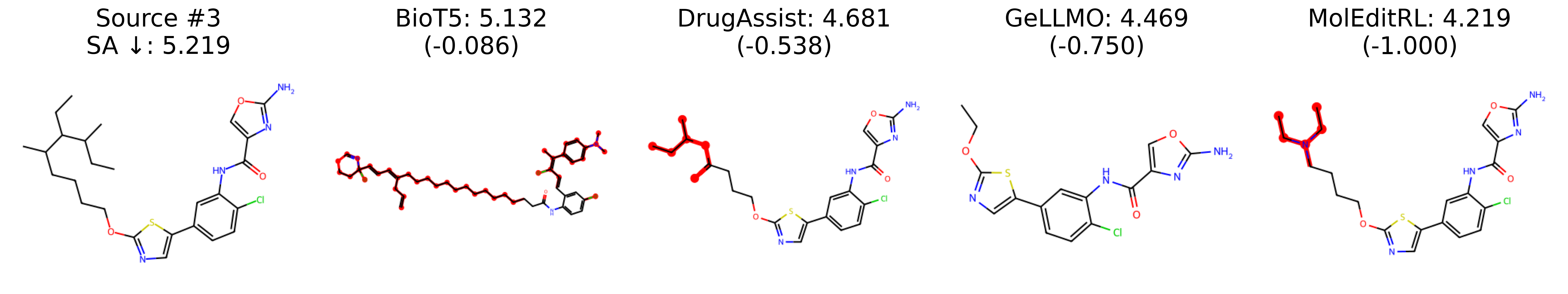}
  \end{subfigure}%

  \begin{subfigure}[b]{0.85\textwidth}
    \centering
    \includegraphics[width=\textwidth]{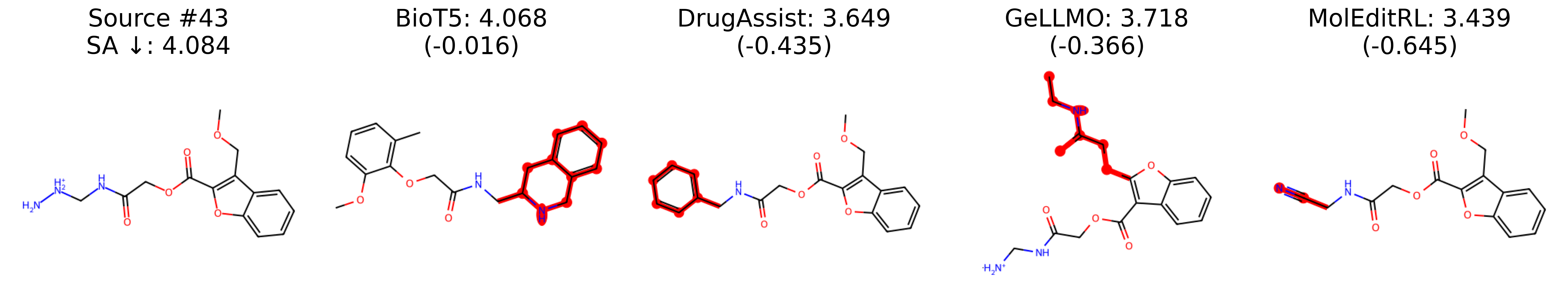}
  \end{subfigure}

  \begin{subfigure}[b]{0.85\textwidth}
    \centering
    \includegraphics[width=\textwidth]{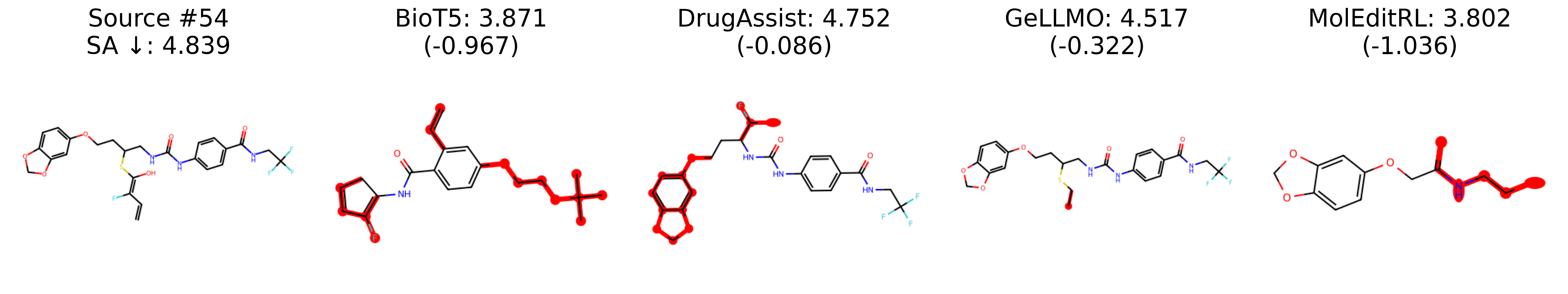}
  \end{subfigure}

  \begin{subfigure}[b]{0.85\textwidth}
    \centering
    \includegraphics[width=\textwidth]{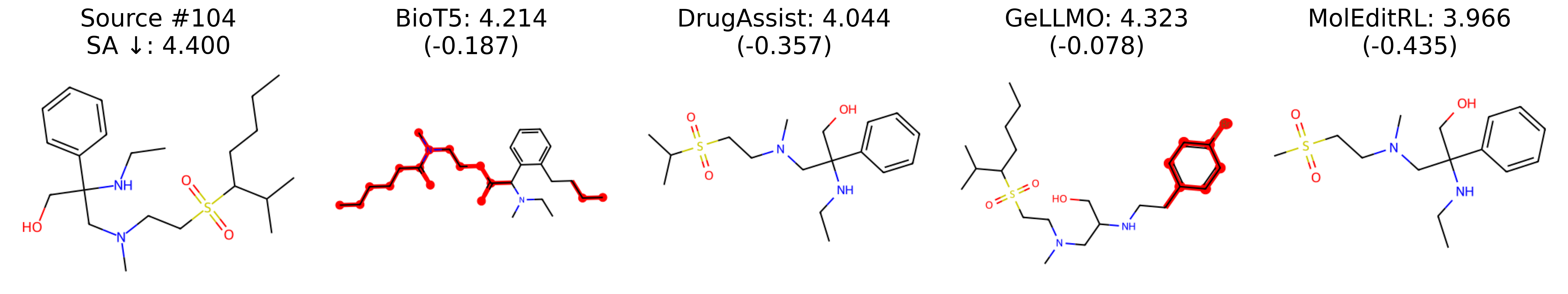}
  \end{subfigure}%

  \begin{subfigure}[b]{0.85\textwidth}
    \centering
    \includegraphics[width=\textwidth]{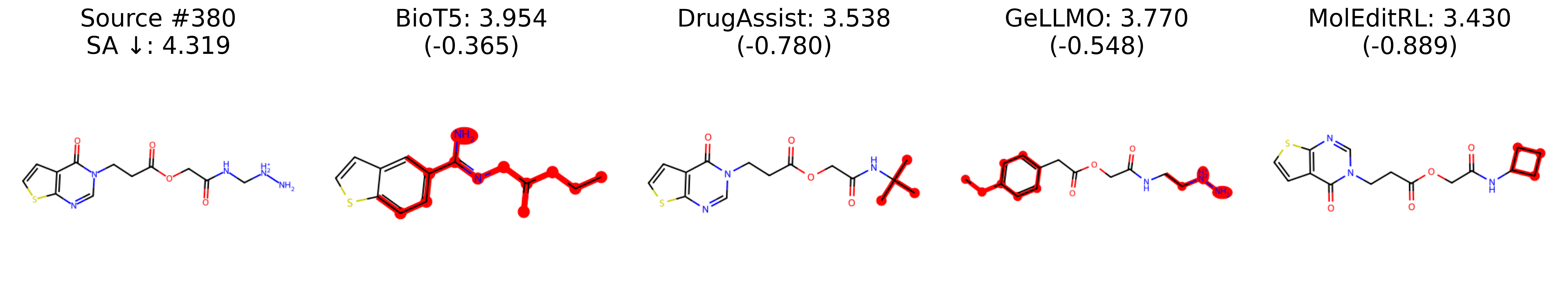}
  \end{subfigure}

  \begin{subfigure}[b]{0.85\textwidth}
    \centering
    \includegraphics[width=\textwidth]{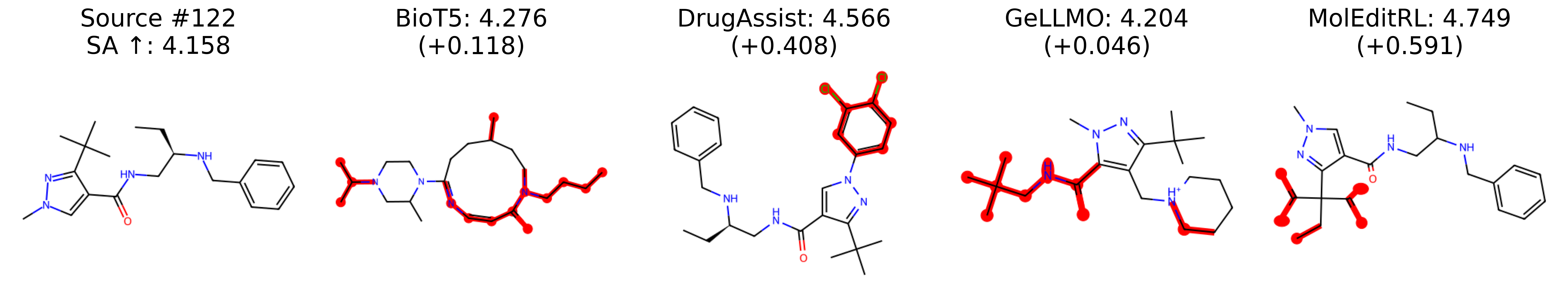}
  \end{subfigure}%

  \begin{subfigure}[b]{0.85\textwidth}
    \centering
    \includegraphics[width=\textwidth]{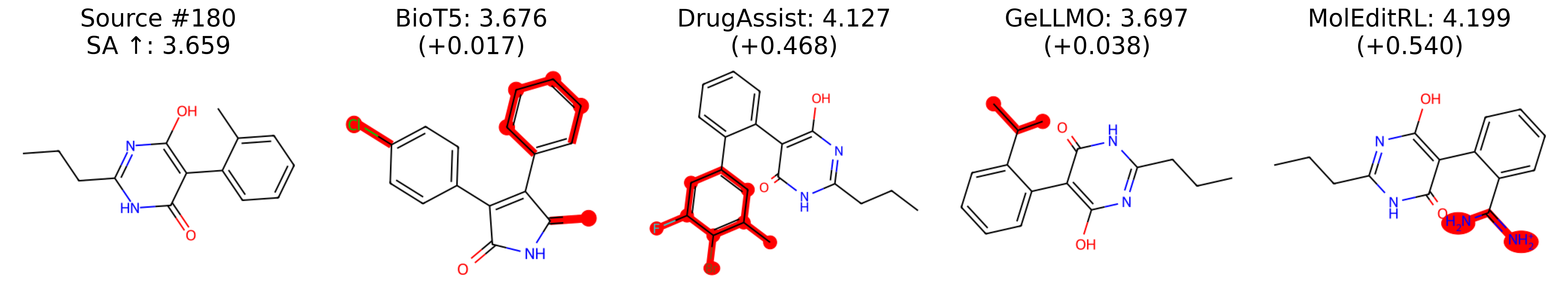}
  \end{subfigure}

  \begin{subfigure}[b]{0.85\textwidth}
    \centering
    \includegraphics[width=\textwidth]{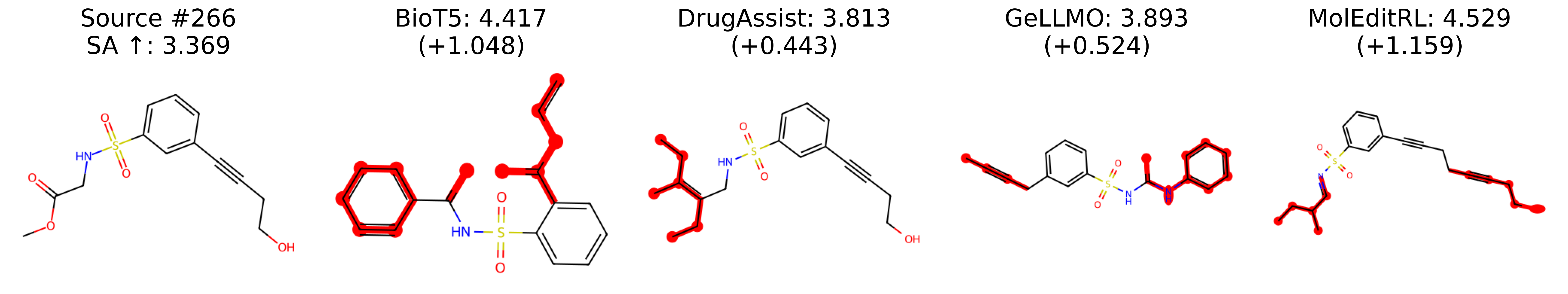}
  \end{subfigure}

  \begin{subfigure}[b]{0.85\textwidth}
    \centering
    \includegraphics[width=\textwidth]{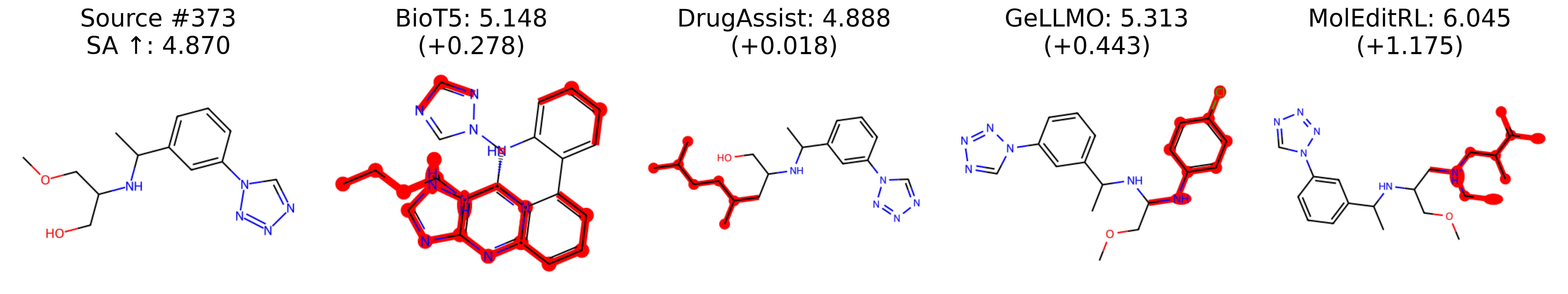}
  \end{subfigure}

  \begin{subfigure}[b]{0.85\textwidth}
    \centering
    \includegraphics[width=\textwidth]{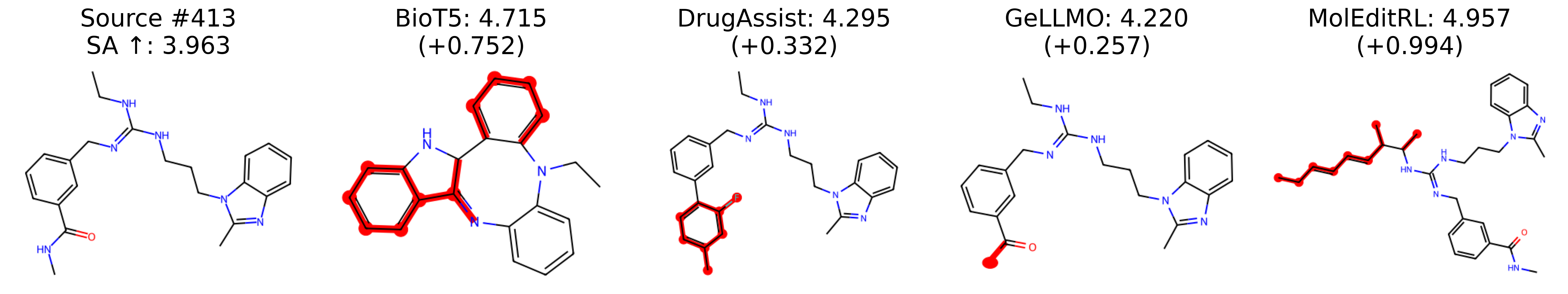}
  \end{subfigure}

  \caption{Qualitative comparison of molecular editing methods.}
  \label{fig:4}
\end{figure}

To further illustrate the editing behavior of different models, we present additional qualitative results in Figure~\ref{fig:visualize1}, Figure~\ref{fig:visualize2}, and Figure~\ref{fig:visualize3}. These figures show visualization of edits across 20 single-property tasks. For each task, subfigure (a) displays the source molecule, and subfigures (b–e) show successful edits produced by BioT5, DrugAssist, GeLLMO\_L, and MolEditRL, respectively. Red-colored substructures indicate regions that have been modified relative to the source molecule.
Across all tasks, MolEditRL consistently achieves the highest number of successful edits, as well as the best structural fidelity—preserving the core scaffold of the original molecule while precisely introducing the required modifications.
Additionally, Figure~\ref{fig:1}, Figure~\ref{fig:2}, and Figure~\ref{fig:4} highlight side-by-side visual comparisons of different models editing the same molecular structure for a single target property. These visualizations confirm that only MolEditRL can reliably perform property-aligned edits while preserving molecular similarity. Competing models often over-modify or disrupt key structural elements, leading to reduced similarity or invalid transformations.

\begin{figure}[htbp]
  \centering
  \begin{subfigure}[b]{0.85\textwidth}
    \centering
    \includegraphics[width=\textwidth]{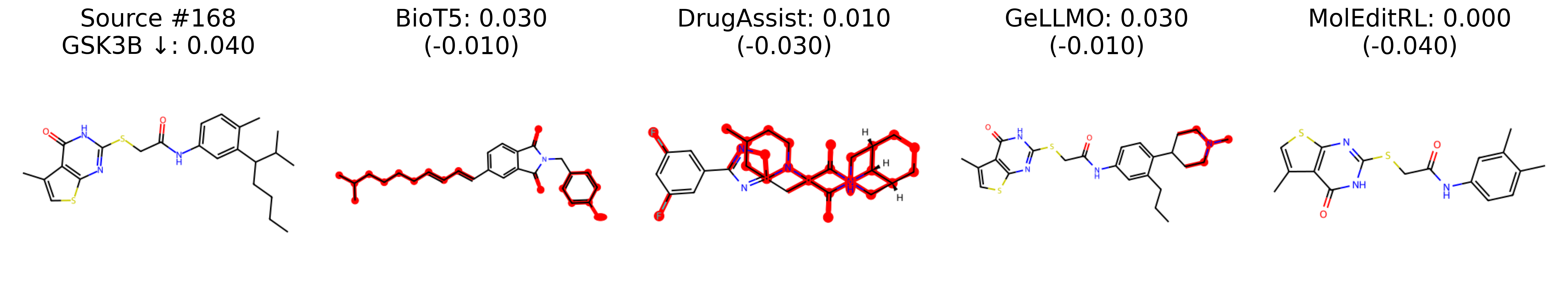}
  \end{subfigure}%

  \begin{subfigure}[b]{0.85\textwidth}
    \centering
    \includegraphics[width=\textwidth]{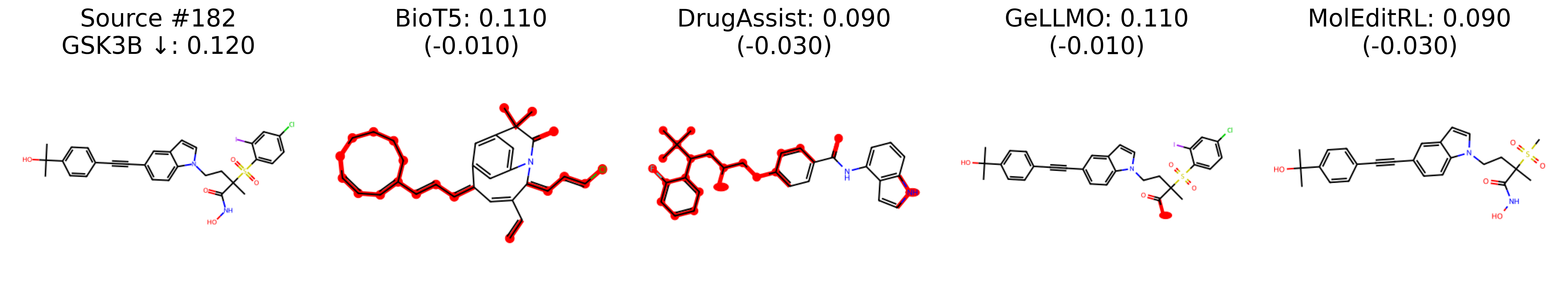}
  \end{subfigure}

  \begin{subfigure}[b]{0.85\textwidth}
    \centering
    \includegraphics[width=\textwidth]{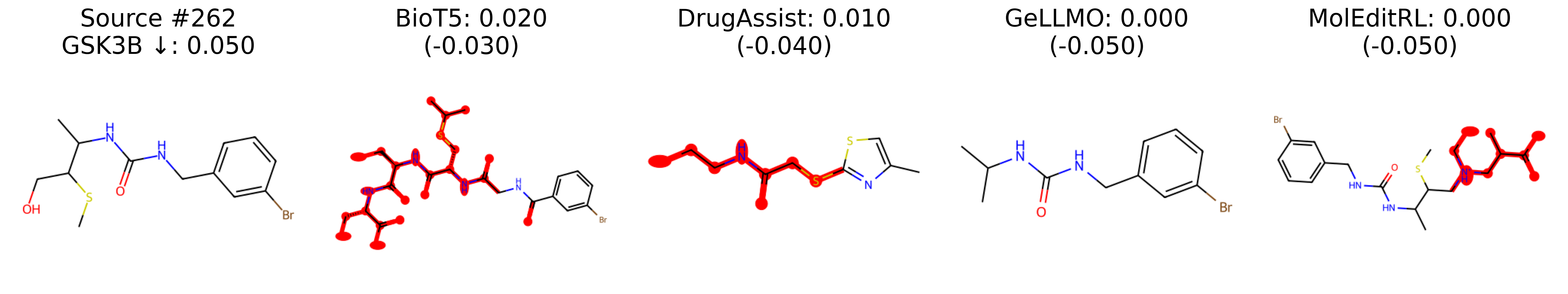}
  \end{subfigure}

  \begin{subfigure}[b]{0.85\textwidth}
    \centering
    \includegraphics[width=\textwidth]{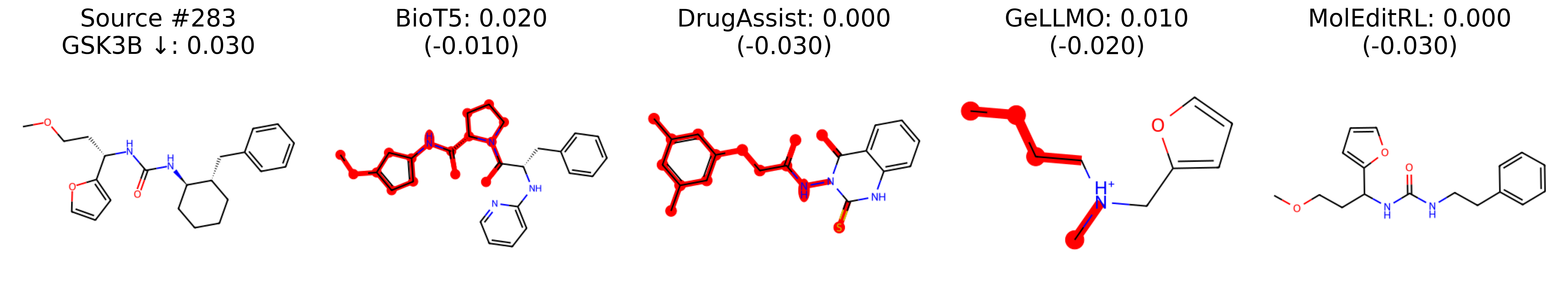}
  \end{subfigure}%

  \begin{subfigure}[b]{0.85\textwidth}
    \centering
    \includegraphics[width=\textwidth]{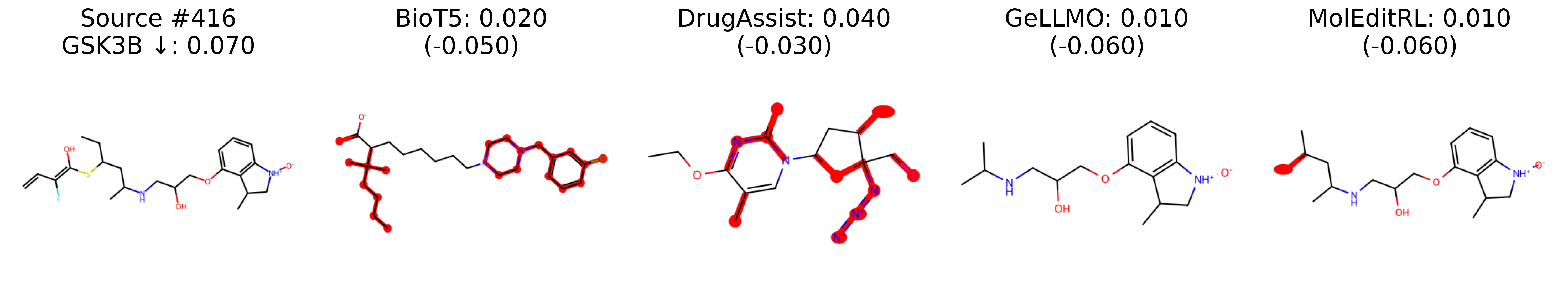}
  \end{subfigure}

  \begin{subfigure}[b]{0.85\textwidth}
    \centering
    \includegraphics[width=\textwidth]{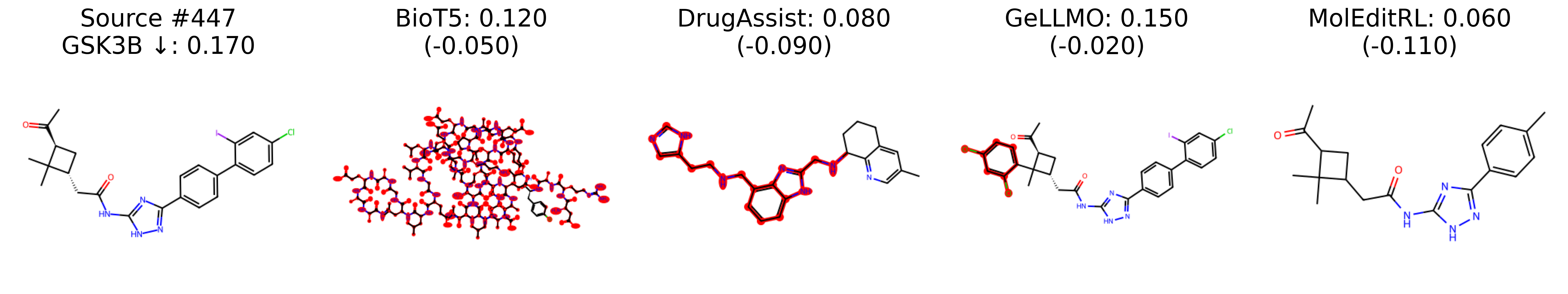}
  \end{subfigure}%

  \begin{subfigure}[b]{0.85\textwidth}
    \centering
    \includegraphics[width=\textwidth]{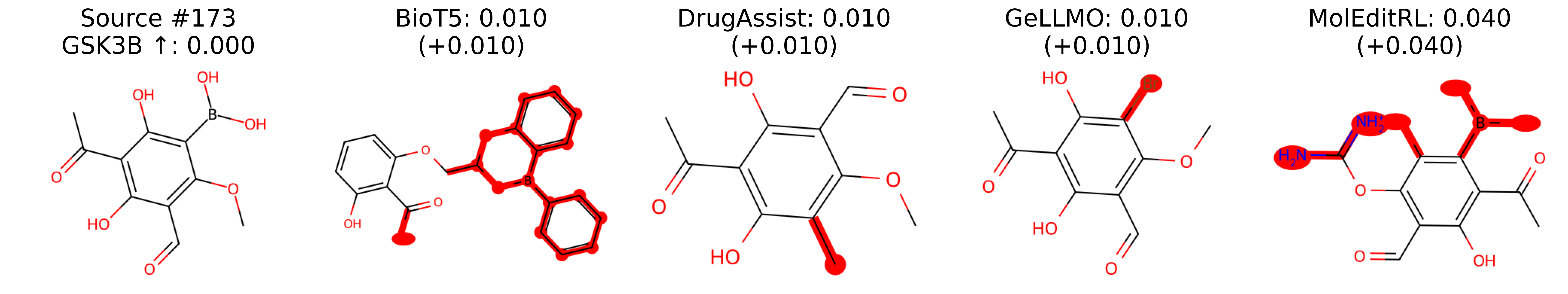}
  \end{subfigure}

  \begin{subfigure}[b]{0.85\textwidth}
    \centering
    \includegraphics[width=\textwidth]{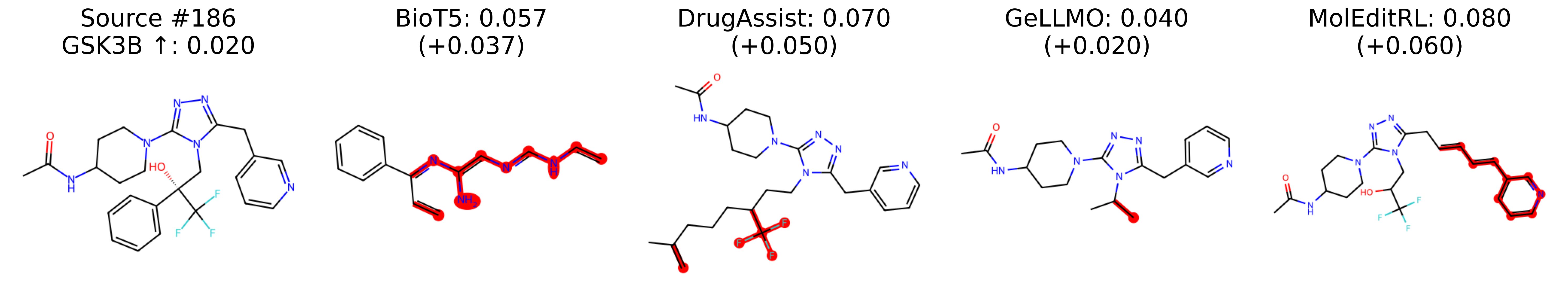}
  \end{subfigure}

  \begin{subfigure}[b]{0.85\textwidth}
    \centering
    \includegraphics[width=\textwidth]{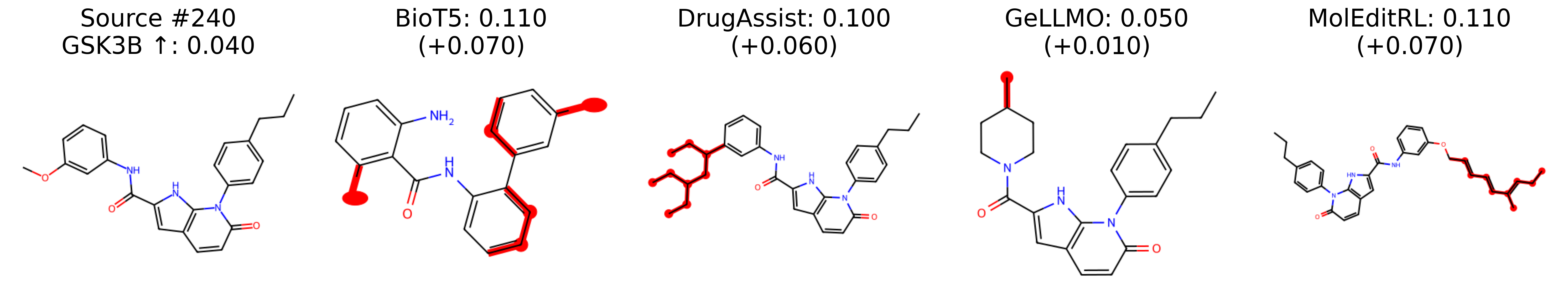}
  \end{subfigure}

  \begin{subfigure}[b]{0.85\textwidth}
    \centering
    \includegraphics[width=\textwidth]{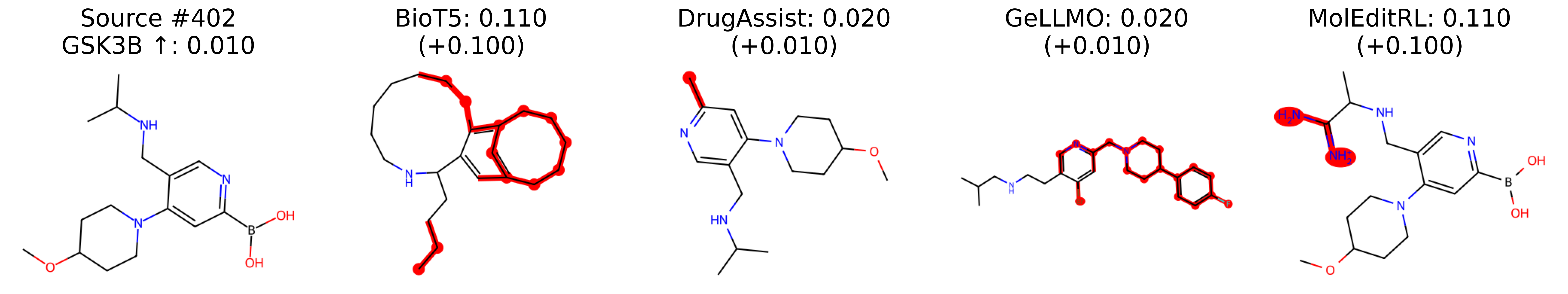}
  \end{subfigure}

  \caption{Qualitative comparison of molecular editing methods.}
  \label{fig:Ablation222}
\end{figure}

\end{document}